\documentclass[preprint,12pt]{elsarticle}
\usepackage{amssymb}
\usepackage{microtype}
\usepackage{graphicx}
\usepackage{subfigure}
\usepackage{booktabs} 
\usepackage{multirow}
\usepackage{algorithmic}
\usepackage{algorithm}
\usepackage{amsmath}
\usepackage{soul}
\usepackage{color}
\usepackage{wrapfig}
\usepackage{epstopdf}

\usepackage{hyperref}
\newcommand{\orcid}[1]{\href{https://orcid.org/#1}{\includegraphics[width=10pt]{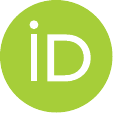}}}

\begin{document}

\begin{frontmatter}

\title{Jensen-Shannon Divergence Based Novel Loss Functions for Bayesian Neural Networks}

\author[inst1]{Ponkrshnan Thiagarajan \orcid{0000-0003-3946-3902}}
\affiliation[inst1]{organization={ Hopkins Extreme Materials Institute}, addressline={Johns Hopkins University}, state={MD}, country={USA}}
\affiliation[inst2]{organization={Department of Mechanical Engineering-Engineering Mechanics, Michigan Technological University}, city={Houghton},state={MI},country={USA}}
\author[inst2,inst3]{Susanta Ghosh\orcid{0000-0002-6262-4121}\corref{aaa}}

\cortext[aaa]{Corresponding Author: susantag@mtu.edu}

\affiliation[inst3]{organization={The  Center for Artificial Intelligence at the Institute of Computing and Cybersystems and \\The Center for Applied Mathematics and Statistics, \\Michigan Technological University},  city={Houghton}, state={MI},country={USA}}

\begin{abstract}
Bayesian neural networks (BNNs) are state-of-the-art machine learning methods that can naturally regularize and systematically quantify uncertainties using their stochastic parameters. Kullback-Leibler (KL) divergence-based variational inference used in BNNs suffer from unstable optimization and challenges in approximating light-tailed posteriors due to the unbounded nature of the KL divergence. To resolve these issues, we formulate a novel loss function for BNNs based on a new modification to the generalized Jensen-Shannon (JS) divergence, which is bounded. In addition,  we propose a Geometric JS divergence-based loss, which is computationally efficient since it can be evaluated analytically. We found that the JS divergence-based variational inference is intractable, and hence employed a constrained optimization framework to formulate these losses. Our theoretical analysis and empirical experiments on multiple regression and classification data sets suggest that the proposed losses perform better than the KL divergence-based loss, especially when the data sets are noisy or biased. Specifically, there are approximately 5\% and 8\% improvements in accuracy for a noise-added CIFAR-10 dataset and a regression dataset, respectively. There is about 13\% reduction in false negative predictions of a biased histopathology dataset. In addition, we quantify and compare the uncertainty metrics for the regression and classification tasks.  
\end{abstract}


\end{frontmatter}

\section{Introduction}\label{sec:introduction}

Despite the widespread success of deep neural networks (DNNs) and convolutional neural networks (CNNs) in numerous applications \cite{samarasinghe2016neural,li2021survey}, they suffer from \emph{overfitting} when the data set is small, noisy, or biased \cite{buda2018systematic, thiagarajan2021explanation}. Furthermore, due to deterministic parameters, CNNs cannot provide a robust measure of uncertainty. Without a measure of uncertainty in the predictions, erroneous predictions by these models may lead to catastrophic failures in applications that require high accuracy, such as autonomous driving and medical diagnosis. Several methods have been developed to provide prediction intervals as a measure of uncertainty in neural networks \cite{kabir2018neural}. 
Amongst these, Bayesian methods have gained eminence due to their rigorous mathematical foundation for uncertainty quantification through their stochastic parameters \cite{jospin2022hands, kabir2018neural}. In addition, Bayesian posteriors are robust to adversarial attacks \cite{10506195}. Due to these advantages, Bayesian Neural Networks have been widely adapted to perform uncertainty quantification in a broad range of fields including but not limited to cosmology \cite{lemos2023robust}, nuclear physics \cite{dong2023nuclear}, material science \cite{pathrudkar2023electronic}, structural health monitoring \cite{luo2024ultrasonic} etc. \\
A Bayesian neural network (BNN) has stochastic parameters whose posterior distribution is learned through the Bayes rule \cite{tishby1989consistent,denker1990transforming,Goan2020,gal2016uncertainty}. 
Since the posterior distribution of parameters is intractable, multiple approaches are proposed to approximate them \cite{magris2023bayesian}. The two most commonly used techniques to approximate the posterior distribution of parameters are variational inference (VI) \cite{hinton1993keeping,barber1998ensemble,graves2011practical,hernandez2015probabilistic} and Markov chain Monte Carlo methods (MCMC) \cite{neal2012bayesian,welling2011bayesian}. MCMC methods comprise a set of algorithms to sample from arbitrary and intractable probability distributions. Inference of posterior using MCMC algorithms can be very accurate but they are computationally demanding \cite{robert2018accelerating}. An additional limitation of MCMC algorithms is that they do not scale well with the model size. \\
The VI is a technique for approximating an intractable posterior distribution by a tractable distribution called the variational distribution. The variational distribution is learned by minimizing an objective function derived from its dissimilarity with respect to the true posterior \cite{blundell2015weight}. %
VI methods are efficient and scale well for larger networks and have gained significant popularity. Most of the VI techniques in the literature use the KL divergence as a measure of the aforementioned dissimilarity. However, the KL divergence is unbounded which may lead to failure during training, poor approximations, and difficulties in approximating light-tailed posteriors as reported in  \citet{hensman2014tilted, dieng2017variational, deasy2020constraining}. 
Therefore, it is imperative to explore alternative divergences that can alleviate these limitations.\\
In terms of exploring alternate divergences, Renyi's $\alpha$-divergences  have been introduced for VI in \citet{li2016renyi}. They proposed a family of variational methods that unified various existing approaches. 
A $\chi$-divergence-based VI has been proposed in \citet{dieng2017variational} that provides an upper bound of the model evidence. Additionally, their results have shown better estimates for the variance of the posterior. Along these lines, an f-Divergence based VI has been proposed in \citet{wan2020f} to use VI for all f-divergences that unified the Reyni divergence \cite{li2016renyi} and $\chi$ divergence \cite{dieng2017variational} based VIs. While these recent works \cite{li2016renyi,dieng2017variational,wan2020f} mainly focused on obtaining a generalized/unified VI framework, the present work specifically attempts to alleviate the limitations due to the unboundedness of the KL divergence-based VI through the Jensen-Shanon (JS) divergence. As a result, two novel loss functions are proposed, which outperform the KL loss in applications that require regularization. The choice of JS divergence is motivated by the fact that it is bounded and it provides an interpolation between forward and reverse KL divergences. In addition,  a modification to the skew-geometric Jensen-Shanon (JS) divergence has been proposed in \citet{deasy2020constraining} to introduce a new loss function for Variational Auto Encoders (VAEs),  which has shown a better reconstruction and generation as compared to existing VAEs. Due to these reasons, JS divergence is explored as an alternative to alleviate the limitations of KL divergence in this work.

\subsection{Key contributions}
In the present work, we propose two novel loss functions for BNNs based on: 
 1)  a novel modified generalized JS divergence (denoted as JS-A) that is bounded and
2)  the geometric JS divergence (denoted as JS-G), which can be analytically evaluated.
The primary contribution of this work is that it resolves the instability in optimization caused by the unboundedness of the KL divergence used in BNNs by formulating a JS-A divergence-based loss function. We found that the JS divergence-based variational inference
is intractable and therefore employ a constrained optimization framework to formulate these losses.

We show that these JS divergence-based loss functions are generalizations of the state-of-the-art KL divergence-based ELBO loss function.  
In addition, through numerical analysis, 
we explain why these loss functions should perform better. 
Further, we derive the conditions under which the proposed JS-G divergence-based loss function regularises better than that of the KL divergence. Furthermore, we perform experiments on multiple regression and classification datasets to show that the proposed loss functions perform better, especially for problems where the data set has noise or is biased towards a particular class. Such biased datasets are often encountered in biomedical image classification, for example, a histopathology dataset used in this work which is biased towards the negative class. Similarly, examples of datasets with noise are encountered in adversarial attacks.  By implementing the proposed method, the RMSE for a standard regression dataset is reduced by 8\%, in the best-case scenario, as compared to the KL loss. Similarly, there is a 12\% reduction in false negative predictions for a binary classification problem that is biased towards the positive class. For a multi-class classification dataset with added noise, the proposed method improves the accuracy by about 5\% as compared to the KL loss. These are significant improvements given that the KL loss is highly successful and widely used in Bayesian neural networks.  In our work, we provide both closed-form and MC-based algorithms for implementing the two JS divergences. The MC implementation can include priors of any family.

The present work is different from the existing work on JS divergence-based VI \citet{deasy2020constraining} for the following reasons: \textbf{(i)} The JS-G divergence proposed in the previous work is unbounded like KL. This issue is resolved by the \mbox{JS-A} divergence proposed in this work. \textbf{(ii)} \citet{deasy2020constraining} introduced the JS-G divergence-based loss for variational autoencoders (VAEs). In the present work, the distributions of parameters of BNNs are learned, which are numerous, as opposed to a small number of latent factors typically found in VAEs.  \textbf{(iii)} The previous work is restricted to Gaussian priors due to the closed-form implementation, which is not the case in this work due to our MC implementation.

\section{Mathematical Background}\label{sec:MathematicalBackground}
  
\subsection{Background: KL and JS divergences } \label{sec:kl&js}
The KL divergence between two random variables $\mathcal{P}$ and $\mathcal{Q}$ on a probability space $\Omega$  is defined as $\text{KL}[p\:||\: q] = \int_\Omega p(x) \log \left[ p(x)/q(x) \right] dx$,
where $p(x)$ and $q(x)$ are the probability distributions of $\mathcal{P}$ and $\mathcal{Q}$ respectively. 

The KL divergence is widely used in literature to represent the dissimilarity between two probability distributions for applications such as VI.  However, it has limitations such as the unboundedness property, i.e. the divergence is infinite when $q(x) = 0$ and $p(x) \neq 0$. This may lead to difficulty in approximating posteriors with a shorter tail than Gaussian distribution as reported in  \citet{hensman2014tilted}. In addition, KL divergence is asymmetric, i.e. $ \text{KL}[p\:||\: q] \neq  \text{KL}[q\:||\: p]$  and hence does not qualify as a metric.  
\\To overcome these limitations a symmetric JS divergence can be employed which is defined as $\text{JS} [p\:||\: q] =   \frac{1}{2} \text{KL} \left[p\left| \left|(p + q)/2\right.\right. \right] + \frac{1}{2} \text{KL} \left[q \left| \left| (p + q)/2 \right.\right. \right]$.

It can be further generalized as, 
\begin{align} \label{eq:js_wadd}
    \text{JS}^{A_\alpha} [p\:||\: q] = (1-\alpha) \text{KL} \left( p\:||\: A_\alpha \right) + \alpha \text{KL} \left( q\:||\: A_\alpha \right)
\end{align}
where, $A_\alpha$ is the weighted arithmetic mean of $p$ and $q$ defined as $A_\alpha = (1-\alpha)p + \alpha q$. Although this JS divergence is symmetric and bounded, unlike the KL divergence its analytical expression cannot be obtained even when $p$ and $q$ are Gaussians. To overcome this difficulty a generalization of the JS divergence using the geometric mean was proposed in \citet{nielsen2019jensen}. By using the weighted geometric mean  $G_\alpha(x,y) =  x^{1-\alpha}y^\alpha$, where $\alpha \in [0,1] $, for two real variables $x$ and $y$, they proposed the following family of skew geometric divergence
\begin{align} \label{eq:jsgalph}
    \text{JS}^{\text{G}_\alpha} \left[ p\:||\: q  \right] &= (1-\alpha) \text{KL}\left(p\:||\: \text{G}_\alpha(p,q)\right)  + \alpha \text{KL}\left(q\:||\: \text{G}_\alpha(p,q)\right) 
\end{align}
The parameter $\alpha$ called a skew parameter, controls the divergence skew between $p$ and $q$. However, the skew-geometric divergence in Eq.~\ref{eq:jsgalph} fails to capture the divergence between $p$ and $q$ and becomes zero for $\alpha = 0$ and $\alpha = 1$. To resolve this issue,  \citet{deasy2020constraining} used the reverse form of geometric mean $ G'_{\alpha}(x,y) =  x^{\alpha}y^{1-\alpha}$, with $ \alpha \in [0,1]$ for JS divergence to use in variational autoencoders. Henceforth, only this reverse form is used for the geometric mean. The JS divergence with this reverse of the geometric mean is given by 
\begin{align} \label{eq:jsgalphdash}
    \text{JS-G} \left[ p\:||\: q  \right] &= (1-\alpha) \text{KL}\left(p\:||\: \text{G}'_\alpha(p,q)\right) + \alpha \text{KL}\left(q\:||\: \text{G}'_\alpha(p,q)\right) 
\end{align}
This yields KL divergences in the limiting values of the skew parameter.
Note that for $\alpha \in [0,1]$, $\text{JS-G}( p || q ) |_\alpha =  \text{JS-G}(p || q)|_{1-\alpha}$ which is not symmetric. However, for $\alpha = 0.5$, the JS-G is symmetric with $ \text{JS-G}(p || q)|_{\alpha = 0.5} =  \text{JS-G}(p || q)|_{\alpha = 0.5}$.
The geometric JS divergences, $\text{JS}^{\text{G}_\alpha}$ and JS-G given in Eq.~\ref{eq:jsgalph} and Eq.~\ref{eq:jsgalphdash} respectively, have  analytical expressions when $p$ and $q$ are Gaussians. However, they are unbounded like the KL divergence.  Whereas, the generalized JS divergence JS$^{A_\alpha}$ in Eq.~\ref{eq:js_wadd} is both bounded and symmetric. 

\subsection{Background: Variational inference} \label{sec:vi}
Given a set of training data $\mathbb{D} = \{\mathbf{x}_i, \mathbf{y}_i\}_{i=1}^N$ and test input, $\mathbf{x} \in \mathbb{R}^{{p}}$, we learn a data-driven model (e.g a BNN) to predict the probability $P(\mathbf{y}|\mathbf{x},\mathbb{D})$ of output $\mathbf{y} \in \Upsilon $, where  $\Upsilon$ is the output space. The posterior probability distribution ($P(\mathbf{w}|\mathbb{D})$) of the parameters ($\mathbf{w}$) of BNN, can be obtained using the Bayes' rule: $P(\mathbf{w} | \mathbb{D})=P(\mathbb{D} | \mathbf{w}) P(\mathbf{w})/P(\mathbb{D})$.

Where $P(\mathbb{D}|\mathbf{w})$ and ${P}(\mathbf{w})$ are the likelihood term and the prior distribution respectively. 
The term $P(\mathbb{D})$, called the evidence, involves marginalization over the distribution of weights: 
$
P(\mathbb{D})=\int_{\Omega_\mathbf{w}} P(\mathbb{D} | \mathbf{w}) P(\mathbf{w}) d\mathbf{w}    
$.
Using the posterior distribution of weights, the predictive distribution of the output can be obtained by marginalizing the weights as $ P(\mathbf{y} | \mathbf{x},\mathbb{D}) =\int_{\Omega_\mathbf{w}} {P}(\mathbf{y} | \mathbf{x}, \mathbf{w}) P(\mathbf{w} | \mathbb{D}) d\mathbf{w}$.

The term $P(\mathbb{D})$ in the Bayes' rule is intractable due to marginalization over $ \mathbf{w}$, which in turn makes $P(\mathbf{w} | \mathbb{D})$ intractable. To alleviate this difficulty, the posterior is approximated using variational inference. 

In variational inference, the unknown intractable posterior $P(\mathbf{w} | \mathbb{D})$ is approximated by a known simpler distribution $q(\mathbf{w}|\boldsymbol{\theta})$ called the variational posterior having parameters $\boldsymbol{\theta}$. 

The set of parameters $\boldsymbol{\theta}$ for the model weights are learned by minimizing the divergence (e.g. KL divergence) between $P(\mathbf{w} | \mathbb{D})$ and  $q(\mathbf{w}|\boldsymbol{\theta})$ as shown in \citet{blundell2015weight}. 
\begin{align} \label{eq:kl_min}
  &\boldsymbol{\theta}^* = \arg \min_{\boldsymbol{\theta}} \text{KL}\left[ q(\mathbf{w}|\boldsymbol{\theta})\:||\: P(\mathbf{w}|\mathbb{D}) \right] \\
    &=  \arg \min_{\boldsymbol{\theta}} \int q(\mathbf{w}|\boldsymbol{\theta}) \left[\log \frac{q(\mathbf{w}|\boldsymbol{\theta})}{P(\mathbf{w})P(\mathbb{D}|\mathbf{w})} + \log P(\mathbb{D}) \right] d\mathbf{w}   
\end{align}

Note that the term $\log P(\mathbb{D})$ in Eq.~\ref{eq:kl_min} is independent of $\boldsymbol{\theta}$ and thus can be eliminated. The resulting loss function $\mathcal{F}(\mathbb{D},\boldsymbol{\theta})$, which is to be minimised to learn the optimal parameters $\boldsymbol{\theta}^*$ is expressed as:
\begin{equation}\label{eq:elbo}
    \mathcal{F}_{KL}(\mathbb{D},\boldsymbol{\theta}) = \text{KL}\left[ q(\mathbf{w}|\boldsymbol{\theta})\:||\: P(\mathbf{w}) \right] - \mathbb{E}_{q(\mathbf{w}|\boldsymbol{\theta})} [\log P(\mathbb{D}|\mathbf{w})]
\end{equation}
The negative of this loss is known as the evidence lower bound (ELBO) \cite{graves2011practical,blundell2015weight}.

\section{Methods}\label{Sec:ProposedWork}
In this section, we provide a modification to the generalized JS divergence, formulations of JS divergence-based loss functions for BNNs,  and insights into the advantages of the proposed loss. 
\subsection{Proposed modification to the generalized JS divergence}
The generalised JS divergence given in Eq.~\ref{eq:js_wadd} fails to capture the divergence between $p$ and $q$ in the limiting cases of $\alpha$ since,
\begin{align}
    &\left[\text{JS}^{\text{A}_\alpha} ( p\:||\: q) \right]_{\alpha=0} = 0;  & \left[ \text{JS}^{\text{A}_\alpha} (p\:||\: q)\right]_{\alpha=1} = 0
\end{align}
To overcome this limitation we propose to modify the weighted arithmetic mean as $A'_\alpha = \alpha p + (1-\alpha) q$, $\alpha \in [0,1]$ which modifies the generalized JS divergence as
\begin{equation} \label{eq:js_waddmod}
   \text{JS-A} [p\:||\: q] = (1-\alpha) \text{KL} \left( p\:||\: A'_\alpha \right) + \alpha \text{KL} \left( q\:||\: A'_\alpha \right)
\end{equation}
Hence, this yields KL divergences in the limiting cases of $\alpha$ as,
\begin{align} \label{eq:limit_case}
   &\left[\text{JS-A}(p\:||\: q) \right]_{\alpha=0} = \text{KL} (p\:||\: q)  \\& \left[\text{JS-A}(p\:||\: q)\right]_{\alpha=1} = \text{KL} (q\:||\: p)
\end{align}

Eq.~\ref{eq:js_waddmod} ensures that JS$(P||q) = 0$ if and only if $P=q, \; \forall \alpha \in [0,1]$. This is necessary since the divergence is a metric to represent statistical dissimilarity.\\ 

\begin{figure}[b!]
    \centering    \includegraphics[width=0.8\linewidth]{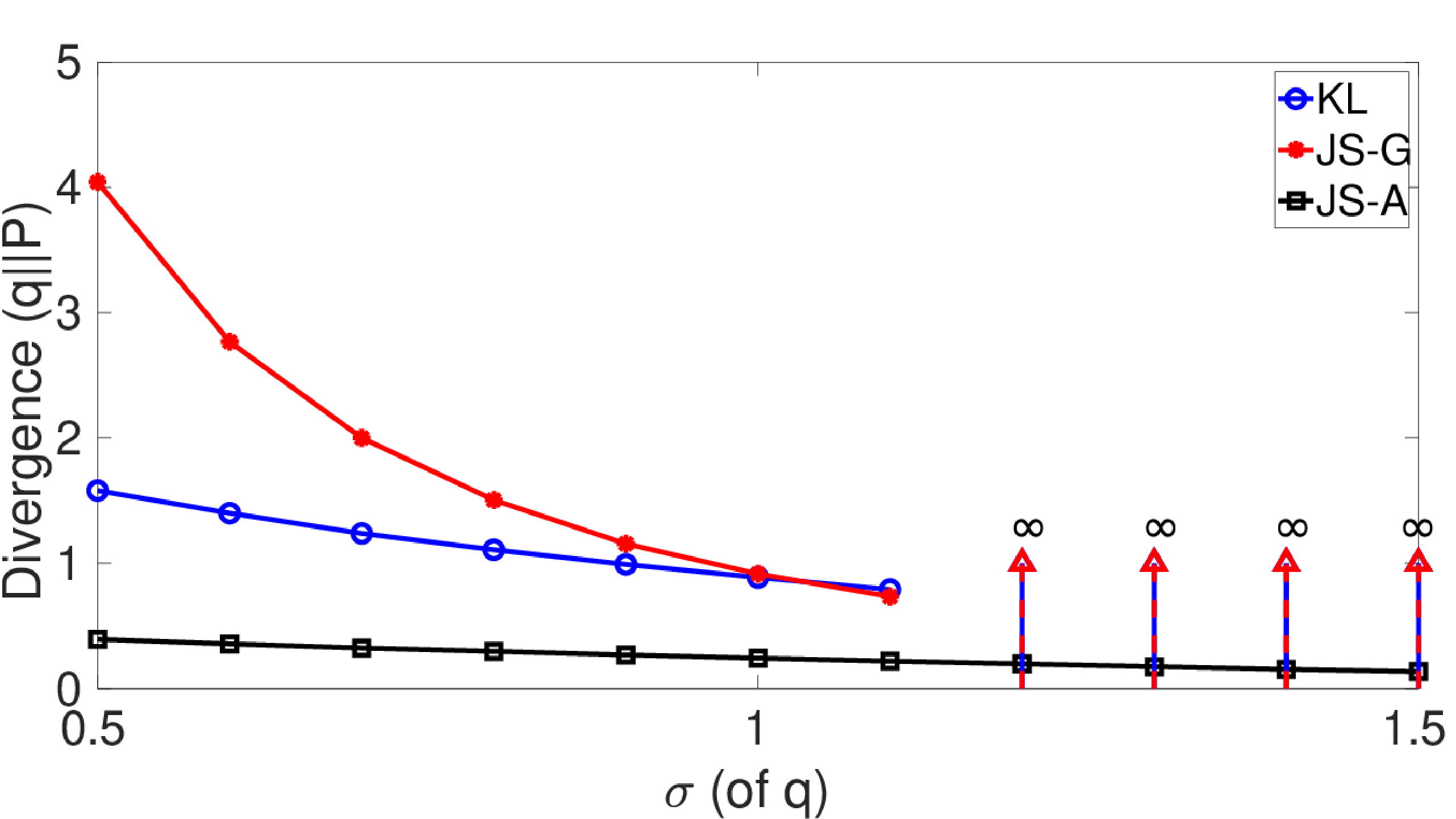}
    \caption{Depiction of the unboundedness (denoted by $\infty$) of the KL and JS-G divergence and the boundedness of the JS-A divergence. The distributions $q$ and $P$ are assumed Gaussian and Uniform respectively with $q = \mathcal{N}(0,\sigma)$ and $P = \mathcal{U}(-5,5)$.} 
    \label{fig:jsa_bound}
\end{figure} 
\textbf{Theorem 1:} Boundedness of the modified generalized JS divergence\\
\textit{For any two distributions $P_1(t)$ and $P_2(t)$, $t \in \Omega$,  the value of the JS-A divergence is bounded such that,}
\begin{align}
    \text{JS-A}(P_1(t) || P_2(t)) \leq -(1- \alpha) \log \alpha  &-  \alpha  \log (1-\alpha), \;\;\text{for } \alpha \in (0,1) 
\end{align}

The proof of Theorem 1 is presented in App.~\ref{sec:theorem1_proof}. Due to this boundedness property of the JS-A divergence, the ensuing loss functions overcome the instability in optimization that is encountered in the KL divergence-based loss.

The unboundedness of the KL and JS-G divergence and the boundedness of JS-A divergence are depicted in Fig~\ref{fig:jsa_bound}. Note that, when both the distributions $P$ and $q$ are heavy-tailed, such as the Gaussian distribution, KL divergence does not become singular.  However, this restricts the choice of the prior significantly.\\
We provide a comparison of various divergences used in this work in  App.~\ref{app:div} and \ref{sec:theorem1_proof}. 
\subsection{Intractability of the JS divergence-based loss functions formulated through the variational inference approach}
In this subsection, we demonstrate that the JS divergence-based variational inference is intractable. 
If the JS-G divergence is used instead of the KL divergence in the VI setting (see Eq.~\ref{eq:kl_min}), the optimization problem becomes,  
\begin{equation} \label{eq:js_min}
    \boldsymbol{\theta}^* = \arg \min_{\boldsymbol{\theta}} \text{JS-G}\left[ q(\mathbf{w}|\boldsymbol{\theta})\:||\: P(\mathbf{w}|\mathbb{D}) \right]\\
\end{equation}
The loss function can then be written as, 
\begin{align}
\label{eq:elbo_js}
\mathcal{F}_{JSG}(\mathbb{D},\boldsymbol{\theta}) &= \text{JS-G}\left[ q(\mathbf{w}|\boldsymbol{\theta})\:||\: P(\mathbf{w}|\mathbb{D}) \right]
    \\&=  (1-\alpha) \text{KL}\left(q\:||\: \text{G}'_\alpha(q,P)\right) + \alpha \text{KL}\left(P\:||\: \text{G}'_\alpha(q,P)\right) \nonumber
\end{align}
Where, $G'_\alpha (q,P) = q(\textbf{w}|\boldsymbol{\theta})^\alpha P(\textbf{w}|\mathbb{D})^{(1-\alpha)} $.
Rewriting the first and the second term in Eq.~\ref{eq:elbo_js} as, 
\begin{align}
    &T_1 = (1-\alpha)^2 \int q(\textbf{w}|\boldsymbol{\theta})\log\left[ \frac{q(\textbf{w}|\boldsymbol{\theta})}{P(\textbf{w}|\mathbb{D})}\right]d\textbf{w}; \\& T_2 = \alpha^2 \int {P(\textbf{w}|\mathbb{D})} \log \left[  \frac{P(\textbf{w}|\mathbb{D})}{q(\textbf{w}|\boldsymbol{\theta})}\right] d\textbf{w}
\end{align}
A detailed derivation of terms $T_1$ and $T_2$ is given in App.~\ref{sec:intract_vi}. Term $T_1$ is equivalent to the loss function in Eq.~\ref{eq:elbo} multiplied by a constant $(1-\alpha)^2$. 

The term $P(\textbf{w}|\mathbb{D})$ in $T_2$ is intractable as explained in section~\ref{sec:vi}. Therefore the JS-G divergence-based loss function given in Eq.~\ref{eq:elbo_js} cannot be used to find the optimum parameter $\boldsymbol{\theta}^*$ which contrasts the KL divergence-based loss function in Eq.~\ref{eq:elbo}. Similarly, the JS-A divergence-based loss function obtained through VI is also intractable. We address this issue of intractability in the following subsection.

\subsection{Proposed JS divergence-based loss functions formulated through a constrained optimization approach}\label{sec:loss_fn} 
To overcome the intractability of the variational inference, we propose to use a constrained optimization framework, following \citet{higgins2016beta, deasy2020constraining}, to derive JS divergence-based loss functions for BNNs. We also show that such a loss function is a generalization of the loss function obtained through the variational inference.

Given a set of training data $\mathbb{D}$, we are interested in learning the probability distribution $q(\mathbf{w}|\boldsymbol{\theta})$ of network parameters such that, the likelihood of observing the data given the parameters is maximized. Thus, the optimization problem can be written as
\begin{equation}
    \max_{\boldsymbol{\theta}} \mathbb{E}_{q(\mathbf{w}|\boldsymbol{\theta})}\left[\log P(\mathbb{D}|\mathbf{w})\right]
\end{equation}
Where $\boldsymbol{\theta}$ is a set of parameters of the probability distribution $q(\mathbf{w}|\boldsymbol{\theta})$. This optimization is constrained to make $q(\mathbf{w}|\boldsymbol{\theta})$ similar to a prior $P(\mathbf{w})$. This leads to a constrained optimization problem as given below:
 
\begin{align}
    \label{eq:const_obj}
    \max_{\boldsymbol{\theta}} \mathbb{E}_{q(\mathbf{w}|\boldsymbol{\theta})}&\left[\log P(\mathbb{D}|\mathbf{w})\right] \;\;\;\text{subject to } D(q(\mathbf{w}|\boldsymbol{\theta}) \:||\:  P(\mathbf{w})) < \epsilon 
\end{align}
where $\epsilon$ is a real number that determines the strength of the applied constraint and D is a divergence measure. Following the  KKT approach, the Lagrangian function corresponding to the constrained optimization problem can be written as
\begin{equation}
    \mathcal{L} = \mathbb{E}_{q(\mathbf{w}|\boldsymbol{\theta})}\left[\log P(\mathbb{D}|\mathbf{w})\right] - \lambda (D(q(\mathbf{w}|\boldsymbol{\theta}) \:||\:  P(\mathbf{w})) - \epsilon)
\end{equation}
Since $\epsilon$ is a constant it can be removed from the optimization. Also changing the sign of the above equations leads to the following loss function that needs to be minimized. \footnote{~The constrained optimization approach-based loss functions are marked by an overhead tilde.}.
\begin{equation}
   \label{eq:cost_fn}
    \widetilde{\mathcal{F}}_D =  \lambda D(q(\mathbf{w}|\boldsymbol{\theta}) \:||\:  P(\mathbf{w})) - \mathbb{E}_{q(\mathbf{w}|\boldsymbol{\theta})}\left[\log P(\mathbb{D}|\mathbf{w})\right]
\end{equation}
This loss function reproduces the ELBO loss  \cite{blundell2015weight} when KL divergence is used and $\lambda$ is taken as 1.

In the following, we obtain loss functions based on the JS-A and JS-G divergences.

\subsubsection{Geometric JS divergence}
Using the modified geometric JS divergence (JS-G) as the measure of divergence in Eq.~\ref{eq:cost_fn} leads to the following loss function: 
\begin{subequations}
   \begin{align}\label{eq:cost_fn_js}
    &\widetilde{\mathcal{F}}_{JSG} = \lambda \text{ JS-G}(q(\mathbf{w}|\boldsymbol{\theta}) \:||\:  P(\mathbf{w})) - \mathbb{E}_{q(\mathbf{w}|\boldsymbol{\theta})}\left[\log P(\mathbb{D}|\mathbf{w})\right] \\
    &=  \lambda (1-\alpha) \text{ KL}(q ||  G'_\alpha (q,P_w)) + \lambda \alpha \text{ KL}(P_w ||  G'_\alpha (q,P_w)) \nonumber  - \mathbb{E}_{q(\mathbf{w}|\boldsymbol{\theta})}\left[\log P(\mathbb{D}|\mathbf{w})\right] 
    \end{align}
\end{subequations}
Note,
 \begin{align*} 
         \text{ KL}(q \:||\:  G'_\alpha (q,P_w)) = \int q&(\mathbf{w}|\boldsymbol{\theta}) \log \frac{q(\mathbf{w}|\boldsymbol{\theta})}{q(\mathbf{w}|\boldsymbol{\theta}) ^\alpha P(\mathbf{w})^{1-\alpha}} d\mathbf{w} \\= (1-&\alpha) \int q(\mathbf{w}|\boldsymbol{\theta}) \log \frac{q(\mathbf{w}|\boldsymbol{\theta})}{ P(\mathbf{w})} d\mathbf{w} \\        
      \text{ KL}(P_w \:||\:  G'_\alpha (q,P_w)) = \int& P(\mathbf{w}) \log \frac{P(\mathbf{w})}{q(\mathbf{w}|\boldsymbol{\theta}) ^\alpha P(\mathbf{w})^{1-\alpha}} d\mathbf{w} \\= \alpha & \int P(\mathbf{w}) \log \frac{ P(\mathbf{w}) }{q(\mathbf{w}|\boldsymbol{\theta})} d\mathbf{w}
 \end{align*}
Hence, the loss function can be written as,
\begin{align} \label{eq:gmex}  
    \begin{split}
     \widetilde{\mathcal{F}}_{JSG} &= \lambda  (1-\alpha)^2 \mathbb{E}_{q(\mathbf{w}|\boldsymbol{\theta})} \left[ \log \frac{q(\mathbf{w}|\boldsymbol{\theta})}{P(\mathbf{w})} \right] \\& \;\;\;\;+  \lambda \alpha^2 \mathbb{E}_{P(\mathbf{w})} \left[ \log \frac{P(\mathbf{w})}{q(\mathbf{w}|\boldsymbol{\theta})} \right] - \mathbb{E}_{q(\mathbf{w}|\boldsymbol{\theta})}\left[\log P(\mathbb{D}|\mathbf{w})\right]      
    \end{split}
\end{align}

In Eq.~\ref{eq:gmex}, the first term is the the \textit{mode seeking} reverse KL divergence  $\text{KL}(q(\mathbf{w}|\boldsymbol{\theta})||P(\mathbf{w}))$  and the second term is the \textit{mean seeking} forward KL divergence KL($P(\mathbf{w})||q(\mathbf{w}|\boldsymbol{\theta})$). Therefore, the proposed loss function offers a weighted sum of the forward and reverse KL divergences in contrast to only the reverse KL divergence in ELBO. Whereas the likelihood part remains identical. The relative weighting between the forward and the reverse KL divergences can be controlled by the parameter $\alpha$. The proposed loss function would ensure better regularisation by imposing stricter penalization if the posterior is away from the prior distribution which will be demonstrated in detail in Sec.~\ref{sec:insight_reg}. The parameters $\lambda$ and $\alpha$ can be used to control the amount of regularisation \footnote{$\lambda$ is taken as 1 for $ \widetilde{\mathcal{F}}_{JSG}$ in this work unless otherwise stated.}.

\subsubsection{Modified generalized JS divergence}
Using the modified Generalised JS divergence (JS-A) as the measure of divergence  in Eq.~\ref{eq:cost_fn} leads to the following loss function:
\begin{align}        
    \begin{split}\label{eq:cost_fn_jsa}
    &\widetilde{\mathcal{F}}_{JSA} = \lambda \text{ JS-A}(q(\mathbf{w}|\boldsymbol{\theta}) \:||\:  P(\mathbf{w})) - \mathbb{E}_{q(\mathbf{w}|\boldsymbol{\theta})}\left[\log P(\mathbb{D}|\mathbf{w})\right]  \\
    &=  \lambda (1-\alpha) \text{ KL}(q \:||\:  A'_\alpha (q,P_w)) + \lambda \alpha \text{ KL}(P_w \:||\:  A'_\alpha (q,P_w)) - \mathbb{E}_{q(\mathbf{w}|\boldsymbol{\theta})}\left[\log P(\mathbb{D}|\mathbf{w})\right]
    \end{split}
\end{align}
Where, $A'_\alpha (q,P_w) = \alpha q + (1-\alpha)P_w$.
The above equation, Eq.~\ref{eq:cost_fn_jsa}, can be expanded as,
\begin{align}     
    \begin{split}    \label{eq:cost_fnex_jsa}    \widetilde{\mathcal{F}}_{JSA} &=  \lambda (1-\alpha) \mathbb{E}_{q(\mathbf{w}|\boldsymbol{\theta})}  \left[ \log \frac{q(\mathbf{w}|\boldsymbol{\theta})}{A'_\alpha (q,P_w)} \right] \\&+  \lambda \alpha  \mathbb{E}_{P(\mathbf{w})} \left[ \log \frac{P(\mathbf{w})}{A'_\alpha (q,P_w)} \right] - \mathbb{E}_{q(\mathbf{w}|\boldsymbol{\theta})}\left[\log P(\mathbb{D}|\mathbf{w})\right]    
     \end{split}
\end{align}   

Note that the proposed loss functions in Eq.~\ref{eq:gmex} and Eq.~\ref{eq:cost_fnex_jsa} yield the ELBO loss for $\alpha = 0$ and $\lambda = 1$. The minimization algorithms for the loss functions Eq.~\ref{eq:gmex} and Eq.~\ref{eq:cost_fnex_jsa} are given in the following section.

\section{Minimisation of the proposed loss functions} \label{sec:cf_imp}

Note that when training using mini-batches, the divergence part of the loss function is normalized by the size of the minibatch ($M$). Therefore, loss for the minibatch $i = 1,2,3,...,M$ can be written as,
\begin{align}
    \widetilde{\mathcal{F}}_i &=  \frac{\lambda}{M} \text{ D}(q(\mathbf{w}|\boldsymbol{\theta}) \:||\:  P(\mathbf{w})) - \mathbb{E}_{q(\mathbf{w}|\boldsymbol{\theta})}\left[\log P(\mathbb{D}_i|\mathbf{w})\right]
\end{align}
\subsection{Evaluation of the JS-G divergence in a closed-form} 
\begin{algorithm}[h]
\caption{Minimization of the JS-G loss function: Closed-form evaluation of the divergence}\label{alg:alg3}
\begin{algorithmic}
\STATE
\STATE \textbf{Initialize} $\boldsymbol{\mu}, \boldsymbol{\rho}$
\STATE \textbf{Evaluate} JS-G term of Eq.~\ref{eq:cost_fn_js} analytically using Eq.~\ref{eq:cfjs_mod}
\STATE \textbf{Evaluate} $\mathbb{E}_{q(\mathbf{w}|\boldsymbol{\theta})}\left[\log P(\mathbb{D}|\mathbf{w})\right]$ term of Eq.~\ref{eq:cost_fn_js} 
\STATE \hspace{0.5cm}Sample ${\boldsymbol{\varepsilon}_i \sim \mathcal{N}}({0}, 1); i=1,..., \text{No. of samples}$
\STATE \hspace{0.5cm}$\mathbf{w}_i \gets \boldsymbol{\mu}+{\log} ({1}+{\exp} (\boldsymbol{\rho})) \circ {\boldsymbol{\varepsilon}_i}.$
\STATE \hspace{0.5cm}$f_{1} \gets \sum_i \log P(\mathbb{D}|\mathbf{w}_i) $
\STATE \textbf{Loss: }
\begin{flalign*}
\;\;\;\;\;\; F &\gets \lambda \: \text{JS-G} - f_{1}&
\end{flalign*}
\STATE \textbf{Gradients: }
\begin{flalign*}
 \;\;\;\;\;\;\frac{\partial F }{\partial \boldsymbol{\mu}} &\gets \sum_i \frac{\partial F}{\partial \mathbf{w}_i}+\frac{\partial F}{\partial \boldsymbol{\mu}}&\\
 \;\;\;\;\;\;\frac{\partial F}{\partial \boldsymbol{\rho}} &\gets \sum_i \frac{\partial F}{\partial \mathbf{w}_i} \frac{\varepsilon_i}{1+\exp (-\boldsymbol{\rho})}+\frac{\partial F}{\partial \boldsymbol{\rho}}
\end{flalign*}
\STATE \textbf{Update: }
\begin{flalign*}
    \;\;\;\;\;\;\boldsymbol{\mu} &\gets \boldsymbol{\mu}-{\beta}\frac{\partial F }{\partial \boldsymbol{\mu}};
\;\;\;\;\;\;\boldsymbol{\rho} \gets \boldsymbol{\rho}-{\beta} { \frac{\partial F }{\partial \boldsymbol{\rho}}}&
\end{flalign*}
\end{algorithmic}
\label{alg2}
\end{algorithm}
In this subsection, we describe the minimization of the JS-G divergence-based loss function by evaluating the divergence in closed form for Gaussian priors.  Assuming the prior and the likelihood are Gaussians, the posterior will also be a Gaussian. Let the prior and posterior be diagonal multivariate Gaussian distributions denoted by $P_N (\mathbf{w}|\boldsymbol{\theta}) = \mathcal{N}(\boldsymbol{\mu_2}, \boldsymbol{\Sigma_2}^2)$ and $q_N (\mathbf{w}) = \mathcal{N}(\boldsymbol{\mu_1}, \boldsymbol{\Sigma_1}^2)$ respectively \footnote{~Where the subscript $()_N$ indicates Gaussian distribution. $\boldsymbol{\mu}_1$ and $\boldsymbol{\mu}_2$ are n-dimensional vectors and $\boldsymbol{\Sigma_1}^2$, $\boldsymbol{\Sigma_2}^2$ are assumed to be diagonal matrices such that  $\boldsymbol{\mu}_1 = [\mu_{11},\mu_{12},...\mu_{1n}]^T $ and $\boldsymbol{\Sigma}_1^2 = \text{diag}( \sigma_{11}^2,\sigma_{12}^2,...\sigma_{1n}^2 )$ (similarly for $\boldsymbol{\mu}_2$ and $\boldsymbol{\Sigma}_2^2$). }.
The closed-form expression of the JS-G divergence between  $q_N (\mathbf{w})$ and $P_N (\mathbf{w}|\boldsymbol{\theta})$  can be written as,
\begin{equation} \label{eq:cfjs_mod}
\begin{split} 
    \text{JS-G}(q_N || P_N) &= \frac{1}{2} \sum_{i=1}^n \left[ \frac{(1-\alpha)\sigma_{1i}^2 + \alpha \sigma_{2i}^2}{\sigma_{\alpha i}^2}   + \log { \frac{(\sigma_{\alpha i}')^2}{\sigma_{1i}^{2(1-\alpha)}\sigma_{2i}^{2\alpha}} } \right.  \\& \;\;\;+(1-\alpha)\frac{(\mu'_{\alpha i}-\mu_{1i})^2}{(\sigma_{\alpha i}')^2} +  \frac{\alpha(\mu'_{\alpha i}-\mu_{2i})^2}{(\sigma_{\alpha i}')^2} - 1 \bigg]
\end{split}
\end{equation}
where,\\ $ (\sigma_{\alpha i}')^2 = \frac{\sigma_{1i}^2\sigma_{2i}^2}{(1-\alpha) \sigma_{1i}^2 + \alpha\sigma_{2i}^2}$;\;\;\; $\mu'_{\alpha i}  = (\sigma'_{\alpha i})^2 \left[ \frac{\alpha\mu_{1 i}}{\sigma_{1i}^2}  + \frac{(1-\alpha) \mu_{2i}}{ \sigma_{2i}^2} \right]$\\

Therefore, the divergence term of the proposed loss function, the first term in Eq.~\ref{eq:cost_fn_js}, can be evaluated by this closed-form expression given in Eq.~\ref{eq:cfjs_mod}. The expectation of the log-likelihood, the second term in  Eq.~\ref{eq:cost_fn_js}, can be approximated by a Monte-Carlo sampling \footnote{The approximations to the loss functions are denoted by F}. The details of the minimization process are given in Algorithm~\ref{alg:alg3}.
Note, for sampling $w_i$ the reparametrization trick is used to separate the deterministic and the stochastic variables. 
\subsection{Evaluation of divergences via a Monte Carlo sampling}
\begin{algorithm}[hp]
\caption{Minimization of the JS-G and JS-A loss functions: Monte Carlo approximation of the divergence}\label{alg:alg2}
\begin{algorithmic}
\STATE 
\STATE \textbf{Initialize} $\boldsymbol{\mu}, \boldsymbol{\rho}$
\STATE \textbf{Approximate} $\mathbb{E}_{q(\mathbf{w}|\boldsymbol{\theta})}$ terms of Eq.~\ref{eq:gmex} or Eq.~\ref{eq:cost_fnex_jsa}
\STATE \hspace{0.5cm}Sample ${\boldsymbol{\varepsilon}_i^q \sim \mathcal{N}}({0}, 1); i=1,..., \text{No. of samples}$
\STATE \hspace{0.5cm}$\mathbf{w}_i^q \gets \boldsymbol{\mu}+{\log} ({1}+{\exp} (\boldsymbol{\rho})) \circ {\boldsymbol{\varepsilon}_i^q}.$\\
\STATE 
\STATE \hspace{0.5cm} {Evaluate first and third terms of Eq.~\ref{eq:gmex}:}
\STATE \hspace{0.5cm}$f_{1} \gets   \sum_i c_1 \log q(\mathbf{w}_i^q|\boldsymbol{\theta})- c_1\log {P(\mathbf{w}_i^q)}- \log P(\mathbb{D}|\mathbf{w}_i^q) $\\
\hspace{0.5cm} where, $c_1 = \lambda  (1-\alpha)^2$\\
\hspace{4cm} (or)
\STATE \hspace{0.5cm} {Evaluate first and third terms of Eq.~\ref{eq:cost_fnex_jsa}:}
\STATE \hspace{0.5cm}$f_{1} \gets   \sum_i c_1 \log q(\mathbf{w}_i^q|\boldsymbol{\theta})- c_1\log {A_\alpha(\mathbf{w}_i^q)}- \log P(\mathbb{D}|\mathbf{w}_i^q) $\\
\hspace{0.5cm} where, $c_1 = \lambda  (1-\alpha)$\\
\STATE 
\STATE  \textbf{Approximate} $\mathbb{E}_{P(\mathbf{w})}$ terms of Eq.~\ref{eq:gmex} or Eq.~\ref{eq:cost_fnex_jsa}
\STATE \hspace{0.5cm}Sample $\mathbf{w}_j^p \sim P(\mathbf{w}); j=1,..., \text{No. of samples}$
\STATE 
\STATE \hspace{0.5cm} {Evaluate second term of  Eq.~\ref{eq:gmex}:}
\STATE \hspace{0.5cm}$ f_2 \gets \sum_j c_2 \log P(\mathbf{w}_j^p) - c_2 \log q(\mathbf{w}_j^p|\boldsymbol{\theta})$\\
\hspace{0.5cm} where, $c_2 = \lambda  \alpha^2$\\
\hspace{4cm} (or)
\STATE \hspace{0.5cm} {Evaluate second term of Eq.~\ref{eq:cost_fnex_jsa}:}
\STATE \hspace{0.5cm}$ f_2 \gets \sum_j c_2 \log A'_\alpha(\mathbf{w}_j^p) - c_2 \log q(\mathbf{w}_j^p|\boldsymbol{\theta})$\\
\hspace{0.5cm} where, $c_2 = \lambda  \alpha^2$
\STATE \textbf{Loss: }
\begin{flalign*}
\;\;\;\;\;\;F &\gets  f_1 + f_2&
\end{flalign*}
\STATE \textbf{Gradients: } 

\begin{flalign*}
 \;\;\;\;\;\;\frac{\partial F }{\partial \boldsymbol{\mu}} &\gets \sum_i \frac{\partial F}{\partial \mathbf{w}_i^q}+ \frac{\partial F}{\partial \boldsymbol{\mu}}&\\
 \;\;\;\;\;\;\frac{\partial F}{\partial \boldsymbol{\rho}} &\gets \sum_i \frac{\partial F}{\partial \mathbf{w}_i^q} \frac{\varepsilon_i}{1+\exp (-\boldsymbol{\rho})}+\frac{\partial F}{\partial \boldsymbol{\rho}}
\end{flalign*}
\STATE \textbf{Update: }

\begin{flalign*}    \;\;\;\;\;\;\boldsymbol{\mu} &\gets \boldsymbol{\mu}-{\beta}\frac{\partial F }{\partial \boldsymbol{\mu}};
\;\;\;\;\;\;\boldsymbol{\rho} \gets \boldsymbol{\rho}-{\beta} { \frac{\partial F }{\partial \boldsymbol{\rho}}}&
\end{flalign*}
\end{algorithmic}
\label{alg3}
\end{algorithm}
In this subsection, we describe the minimization of the JS divergence-based loss functions by evaluating the divergences using the Monte Carlo sampling technique. The algorithm provided in this subsection is more general as it applies to both the JS-G and the JS-A divergences with no restrictions on the priors. 
The loss functions given in Eq.~\ref{eq:gmex} and Eq.~\ref{eq:cost_fnex_jsa} can be approximated using Monte Carlo samples from the corresponding distributions as shown in Algorithm \ref{alg:alg2}.

\section{Insights into the proposed JS divergence-based loss functions}
To better understand the proposed JS divergence-based loss functions, we use a contrived example to compare them against the conventional KL divergence-based loss function. In the following, we explore the regularization ability of the proposed loss functions. Further insights on Monte Carlo estimates are given in App.~\ref{app:mc}

\subsubsection{Regularisation performance of JS divergences} \label{sec:insight_reg}
\begin{figure}[h]
    \centering
    \subfigure[]{\includegraphics[width=0.4\linewidth]{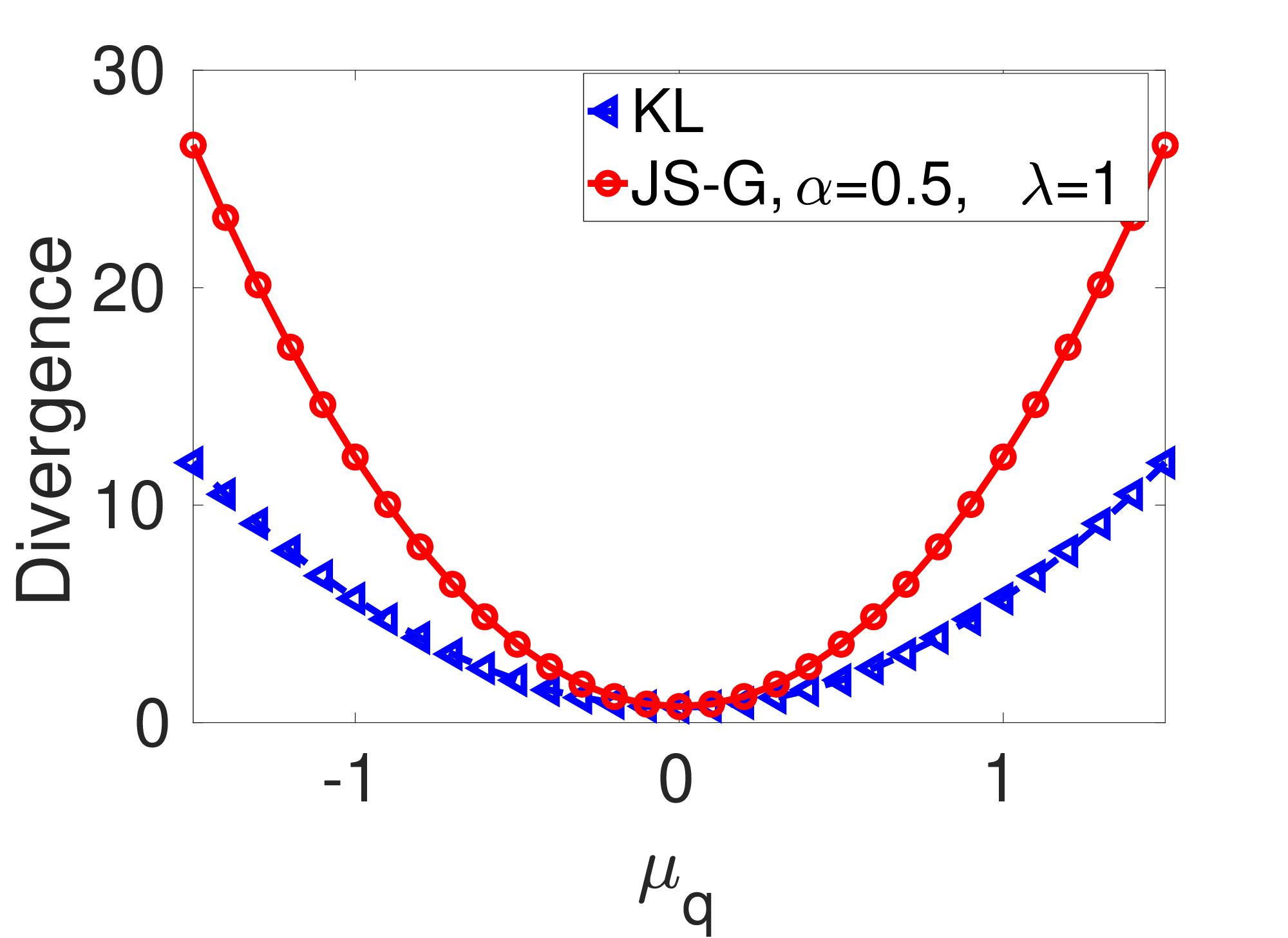}\label{fig:mu_div}}
    \subfigure[]{\includegraphics[width=0.4\linewidth]{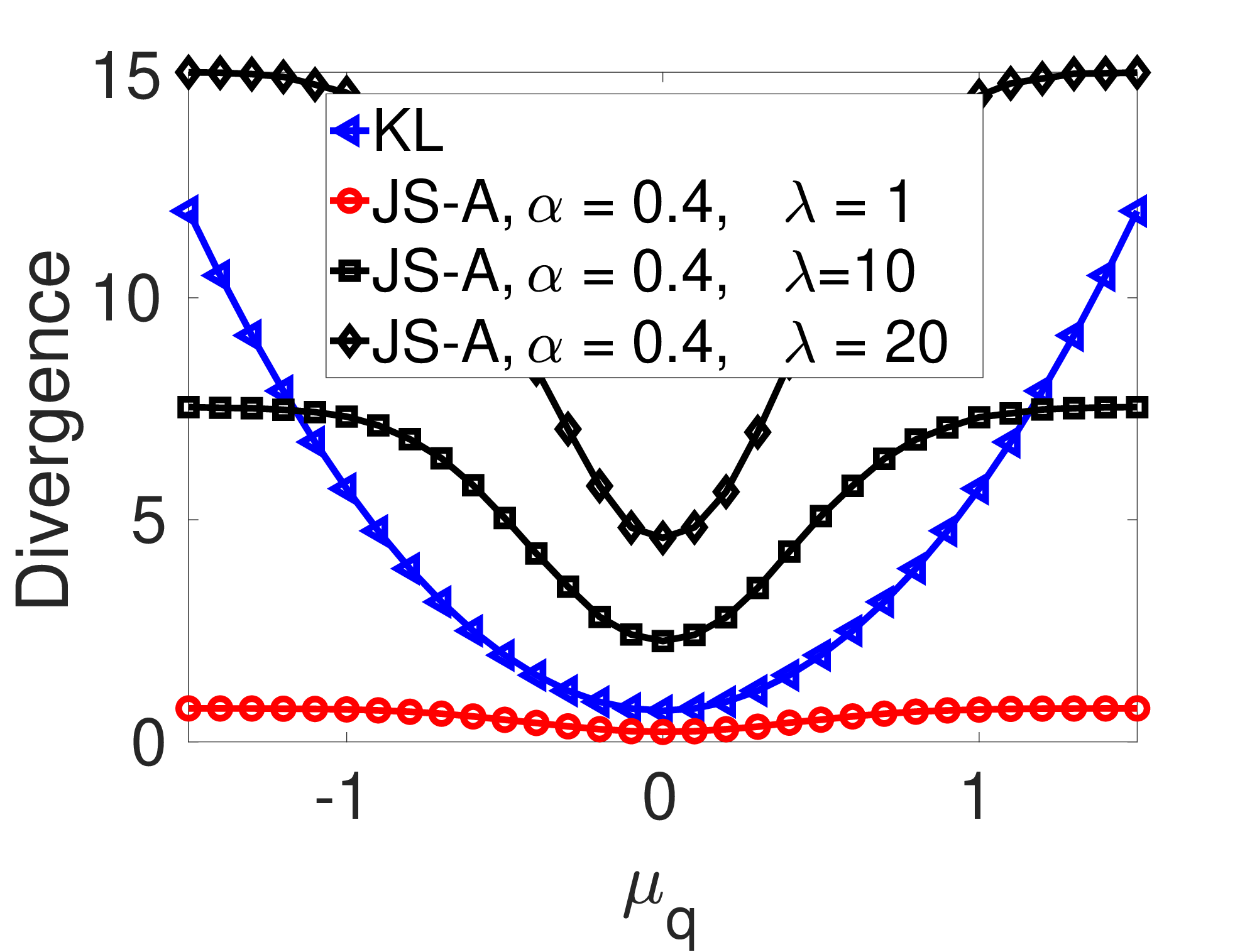}\label{fig:mu_div_jsa}}
    \subfigure[]{\includegraphics[width=0.4\linewidth]{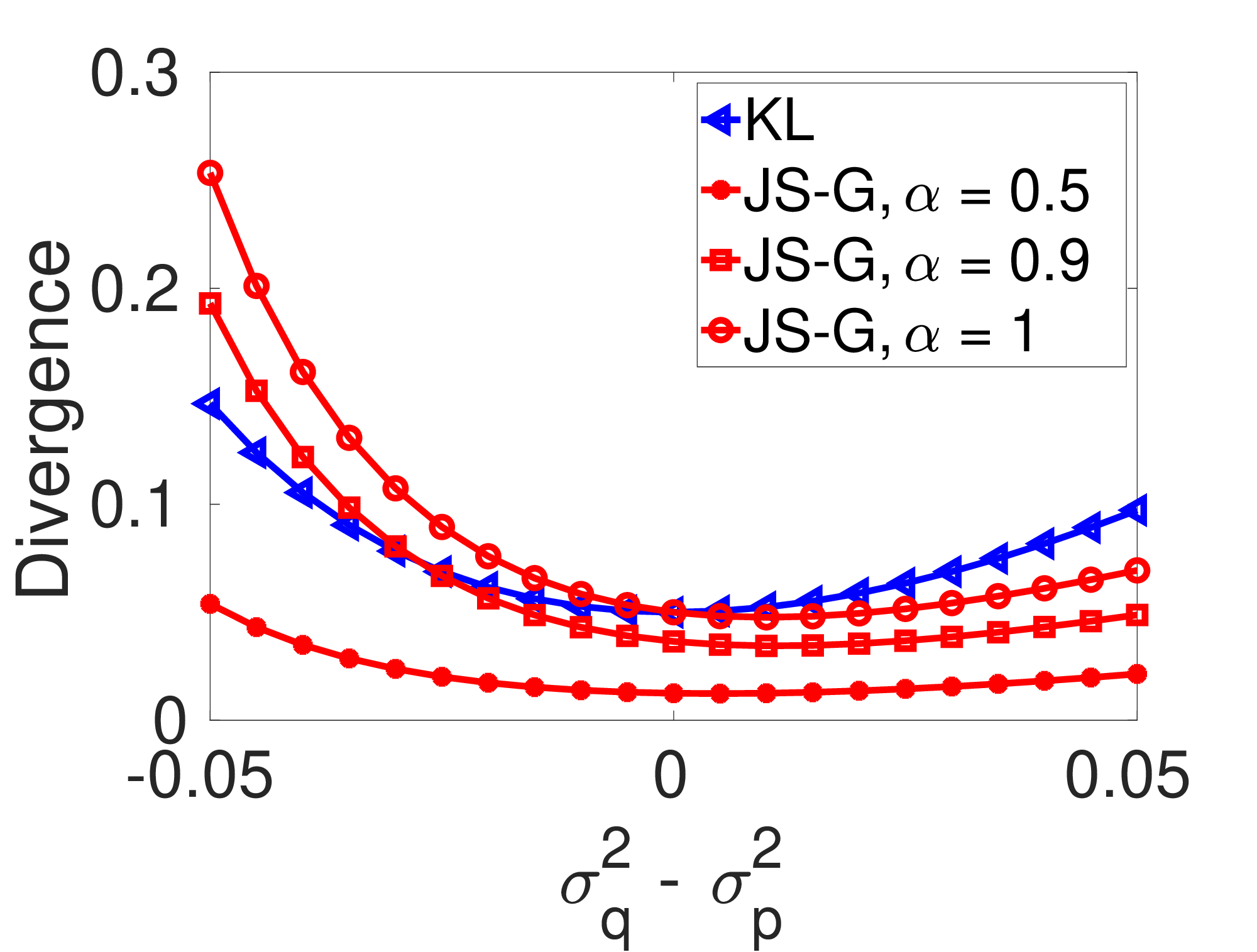}\label{fig:sig_div_kl}}
    \subfigure[]{\includegraphics[width=0.4\linewidth]{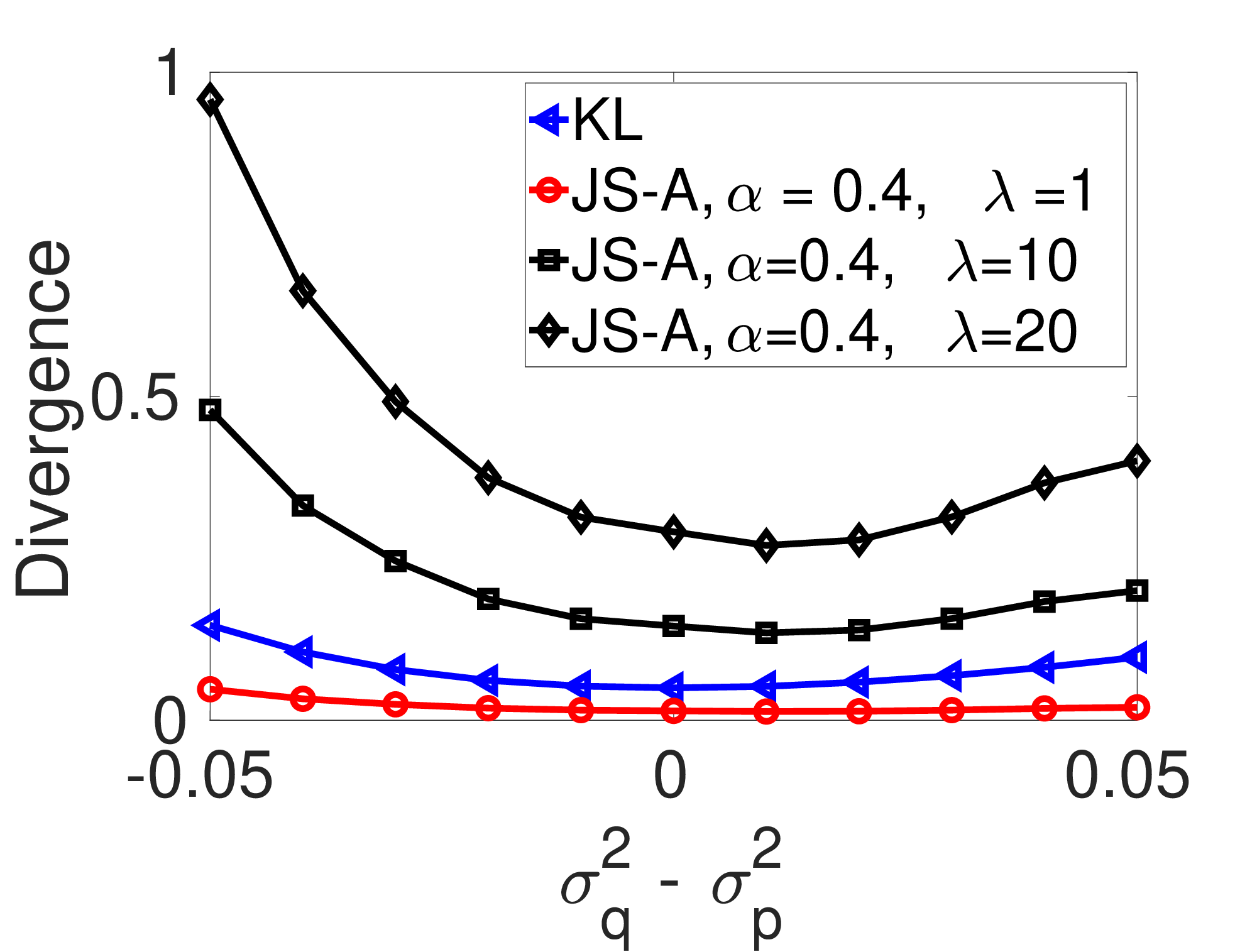}\label{fig:sig_div_jsa}}
    \caption{Comparison of the KL and the JS divergences of distributions P and q. (a) and (b) $\sigma_q^2, \mu_p, \sigma_p^2 $ are fixed and $\mu_q$ is varied. (c) and (d)  $\mu_q, \mu_p, \sigma_p^2 $ are fixed and $\sigma_q^2$ is varied. The fixed values of the parameters are  $\mu_q = 0.1 ,\sigma_q^2 = 0.01, \mu_p=0, \sigma_p^2 = 0.1$}
    \label{fig:div_comp}
\end{figure}
Let two Gaussian distributions $q = \mathcal{N}(\mu_q,\sigma_q^2)$ and $P = \mathcal{N}(\mu_p,\sigma_p^2)$ represent the posterior and the prior distribution of a parameter in a BNN. The KL, JS-A, and JS-G divergences are evaluated by varying the mean and variance of the distribution $q$. This emulates the learning of the network parameter during training. 
Fig.~\ref{fig:div_comp} shows that as the posterior distribution ($q$) moves away from the prior distribution ($P$), the JS divergences increase more rapidly than the KL divergence. In the case of the JS-A divergence in Fig.\ref{fig:mu_div_jsa} and \ref{fig:sig_div_jsa}, this is achieved by a higher value of $\lambda$. This implies that a greater penalization is offered by JS divergences than the KL divergence as the posterior deviates away from the prior (also seen in App.~\ref{sec:app_evol}). Thus, by assuming small values for the means of prior distributions we can regularize better by the proposed JS divergences. In practice, zero mean Gaussian priors are widely accepted for BNNs. For such priors, higher penalization of the loss function implies pushing the parameters' mean closer to zero while learning the complexity of the data. In doing this, we can achieve better regularization. This regularization process requires finding optimal values of $\alpha$ and $\lambda$ through hyperparameter optimization. In the following subsection, we theoretically analyze the regularization performance of the JS-G divergence.

\subsubsection{Condition for better regularisation of $\widetilde{\mathcal{F}}_{JSG} $} \label{app:reg}
The above example shows that the JS-G divergence is greater than the KL for the given Gaussian distributions. To generalize it further, we propose the following theorems that hold for any two arbitrary distributions. \\

\textbf{Theorem 2. }\textit{For any two arbitrary distributions $P$ and $q$ such that $P \neq q$, $ \widetilde{\mathcal{F}}_{JSG} > {\mathcal{F}}_{KL} $ if and only if $\alpha > \frac{2 \text{ KL} (q||P)}{\text{KL}(q||P) + \text{KL} (P||q)} \in (0,\infty)$}\\

\textbf{Proof:}
Assuming,     $\widetilde{\mathcal{F}}_{JSG} - {\mathcal{F}}_{KL} > 0$ and from Eq.~\ref{eq:elbo} and Eq.~\ref{eq:gmex} we have,
\begin{align*}
     (1-\alpha)^2 \text{KL}(q||P) + \alpha^2 \text{KL} (P||q) - \text{KL}(q||P) > 0&\\
     (\alpha^2 - 2\alpha) \text{KL}(q||P) + \alpha^2 \text{KL} (P||q) > 0&\\
     \text{This leads to, \;\;\;\;\;}
     \alpha > \frac{2 \text{ KL} (q||P)}{\text{KL}(q||P) + \text{KL} (P||q)}& 
\end{align*}
This proves that if $ \widetilde{\mathcal{F}}_{JSG} > {\mathcal{F}}_{KL} $  then $\alpha > \frac{2 \text{ KL} (q||P)}{\text{KL}(q||P) + \text{KL} (P||q)} $ .  
The converse can be proved similarly. A detailed proof is shown in the App.~\ref{sec:theorem2_proof}.\\

\textbf{Theorem 3.} \textit{If $P = \mathcal{N}(\mu_p,\sigma_p^2)\,$ and $\,q = \mathcal{N}(\mu_q,\sigma_q^2)\,$ are Gaussian distributions and $\,P \neq q\,$, then $\frac{2 \text{ KL} (q||P)}{\text{KL}(q||P) + \text{KL} (P||q)}<1 \,$ if and only if $\, \sigma_p^2 > \sigma_q^2$ }.\\

\textbf{Proof: }
Assuming   $\frac{2 \text{ KL} (q||P)}{\text{KL}(q||P) + \text{KL} (P||q)} <1$, we get 
\begin{align}
    \text{KL} (P||q)&>\text{KL}(q||P) \label{eq:ineqkl}
\end{align}
Since $P = \mathcal{N}(\mu_p,\sigma_p^2)$ and $q = \mathcal{N}(\mu_q,\sigma_q^2)$,  Eq.~\eqref{eq:ineqkl} can be written as, 
\begin{align*}
     \ln \frac{\sigma_q^2}{\sigma_p^2} + \frac{\sigma_p^2 + (\mu_q - \mu_p)^2}{\sigma_q^2} &-1 >  \ln \frac{\sigma_p^2}{\sigma_q^2} + \frac{\sigma_q^2+(\mu_p - \mu_q)^2}{\sigma_p^2} -1 
     \intertext{Denoting $\gamma = \frac{\sigma_p^2}{\sigma_q^2} $, we get,}
     \gamma - \frac{1}{\gamma} + \ln{\frac{1}{\gamma}} - \ln{\gamma} +&  \frac{(\mu_q - \mu_p)^2}{\sigma_q^2} - \frac{(\mu_p - \mu_q)^2}{\gamma \sigma_q^2} > 0
\end{align*}
\begin{align}
\begin{split}
    \text{or,} \;\;\; \ln\left[\frac{1}{\gamma^2}\exp\left({\gamma - \frac{1}{\gamma}}\right)\right] &+  \frac{(\mu_q - \mu_p)^2}{\sigma_q^2}\left(1 - \frac{1}{\gamma}\right) > 0 \label{eq:fianl_cond}
\end{split}
\end{align}
This condition Eq.~\ref{eq:fianl_cond} is satisfied only when $\gamma > 1$, which implies $\sigma_p^2 > \sigma_q^2$.
Thus if $\frac{2 \text{ KL} (q||P)}{\text{KL}(q||P) + \text{KL} (P||q)} <1$ then $\sigma_p^2 > \sigma_q^2\,$. This result is also observed in Fig.~\ref{fig:sig_div_kl}. The converse can be proved similarly as shown in App.~\ref{sec:theorem3_proof}.\\

\textbf{Corollary: }From Theorem 2 and 3: $ \widetilde{\mathcal{F}}_{JSG} > {\mathcal{F}}_{KL} $ if  $\sigma_p^2 > \sigma_q^2$ and $\forall \, \alpha \in (0,1]$ such that $ \alpha > \frac{2 \text{ KL} (q||P)}{\text{KL}(q||P) + \text{KL} (P||q)}$. Where, $P$ and $q$ are Gaussians and $P \neq q$.

\section{Experiments} \label{sec:Experiments}
In order to demonstrate the advantages of the proposed losses in comparison to the KL loss, we performed experiments. 
We have implemented the divergence part of the JS-G loss and the JS-A loss via a closed-form expression and a Monte-Carlo method respectively, in these experiments.

\subsection{Data sets} \label{sec:dataset}
The experiments were performed on multiple regression and classification datasets. Regression experiments were performed on six data sets (Airfoil, Aquatic, Building, Concrete, Real Estate, and Wine) from the UCI Machine Learning Repository \cite{uci}. 

For classification two data sets are considered: the CIFAR-10 data set  \cite{krizhevsky2009learning}  and a histopathology data set \cite{janowczyk2016deep,cruz2014automatic,mooney_b_2017}. 
To demonstrate the effectiveness of regularisation, varying levels of Gaussian noise were added to the normalized CIFAR-10 data set for training, validation, and testing.
We also used a histopathology data set which is highly biased towards one class.\\
Further details on these data sets and the pre-processing steps used here are provided in \ref{app:dataset}.

\subsection{Hyperparameter optimisation and network architecture} \label{sec:hyp}
Regression experiments on multiple datasets are conducted following the framework of \citep{wan2020f,li2016renyi}. The network architecture is a single layer with 50 hidden units and a ReLU activation function. We assume the priors to be Gaussians $p \sim \mathcal{N}(0, I)$ and 100 MC approximations are used to calculate the NLL part of the loss function for the KL and JS-G divergence-based losses, and 10 MC samples are used to evaluate the JS-A divergence based loss. The likelihood function is Gaussian with noise parameter $\sigma$ which is also learned along with the parameters of the network. Six datasets (Airfoil, Aquatic, Building, Concrete, Real Estate, and Wine)  are evaluated and they are split into 90\% and 10\% for training and testing respectively.

\par Hyperparameters of the networks used for classification are chosen through hyperparameter optimization. It is performed using a Tree-structured Parzen Estimator (TPE) algorithm \cite{bergstra2011algorithms}  to maximize the validation accuracy for different hyperparameter settings of the network. Further details and the results of the hyperparameter optimization are given in \ref{app:hyperres}. 
 The architecture of the networks used for classification follows the ResNet-18 V1 model \cite{he2016deep} without the batch normalization layers. The network parameters are initialized with the weights of ResNet-18 trained on the Imagenet data set \cite{krizhevsky2012imagenet}.

\section{Classification results}\label{sec:Results}
This section presents the results of the classification experiments and the performance comparison between the KL loss and the proposed JS losses. 

\subsection{Training and validation}

Three Bayesian CNNs were trained by minimizing the KL and proposed JS losses. Training of the networks is done until the loss converges or the validation accuracy starts to decrease.
\begin{figure}[htpb]
    \centering
    \subfigure[][]{\includegraphics[width=0.8\linewidth]{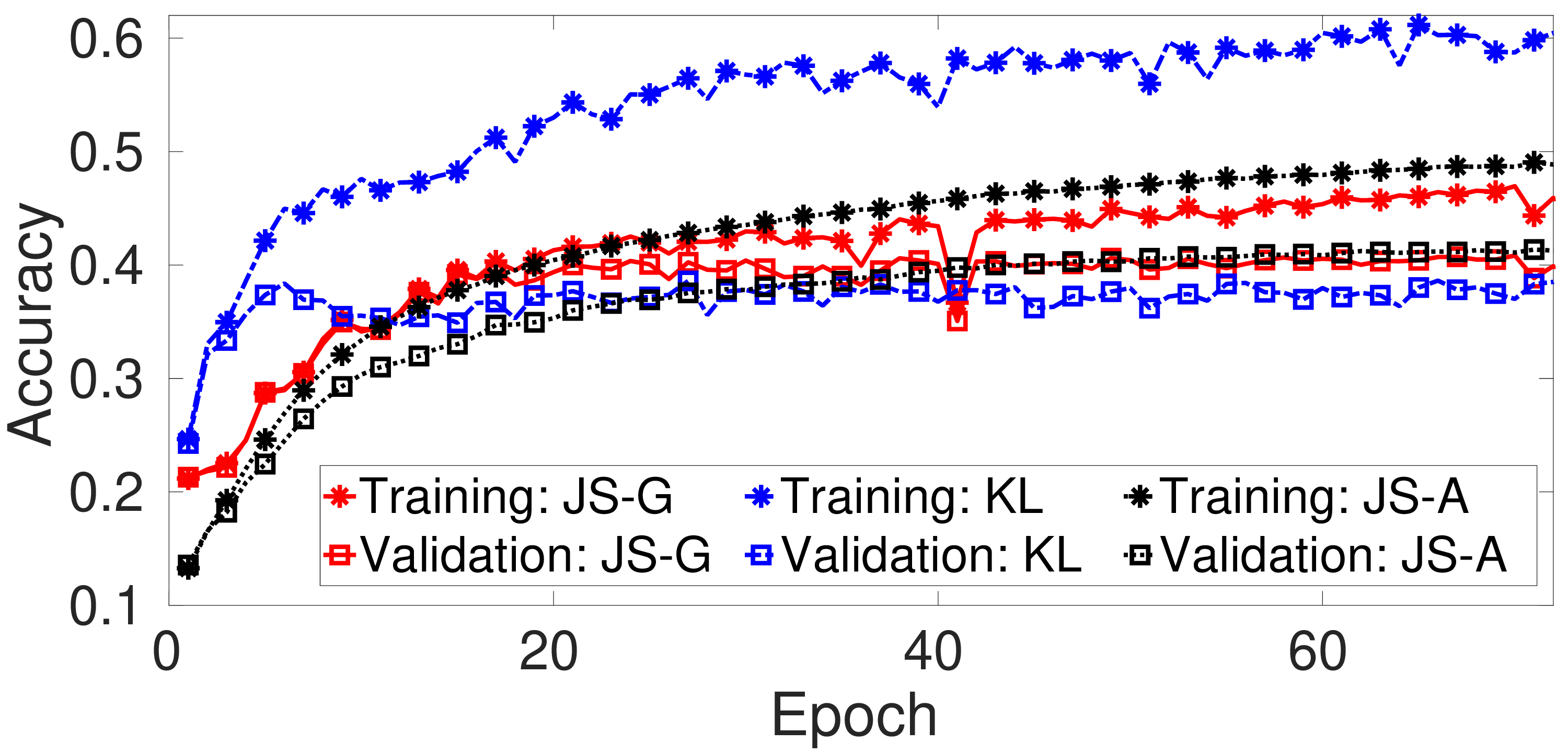}\label{fig:training_Cifar}}
    \subfigure[][]{\includegraphics[width=0.8\linewidth]{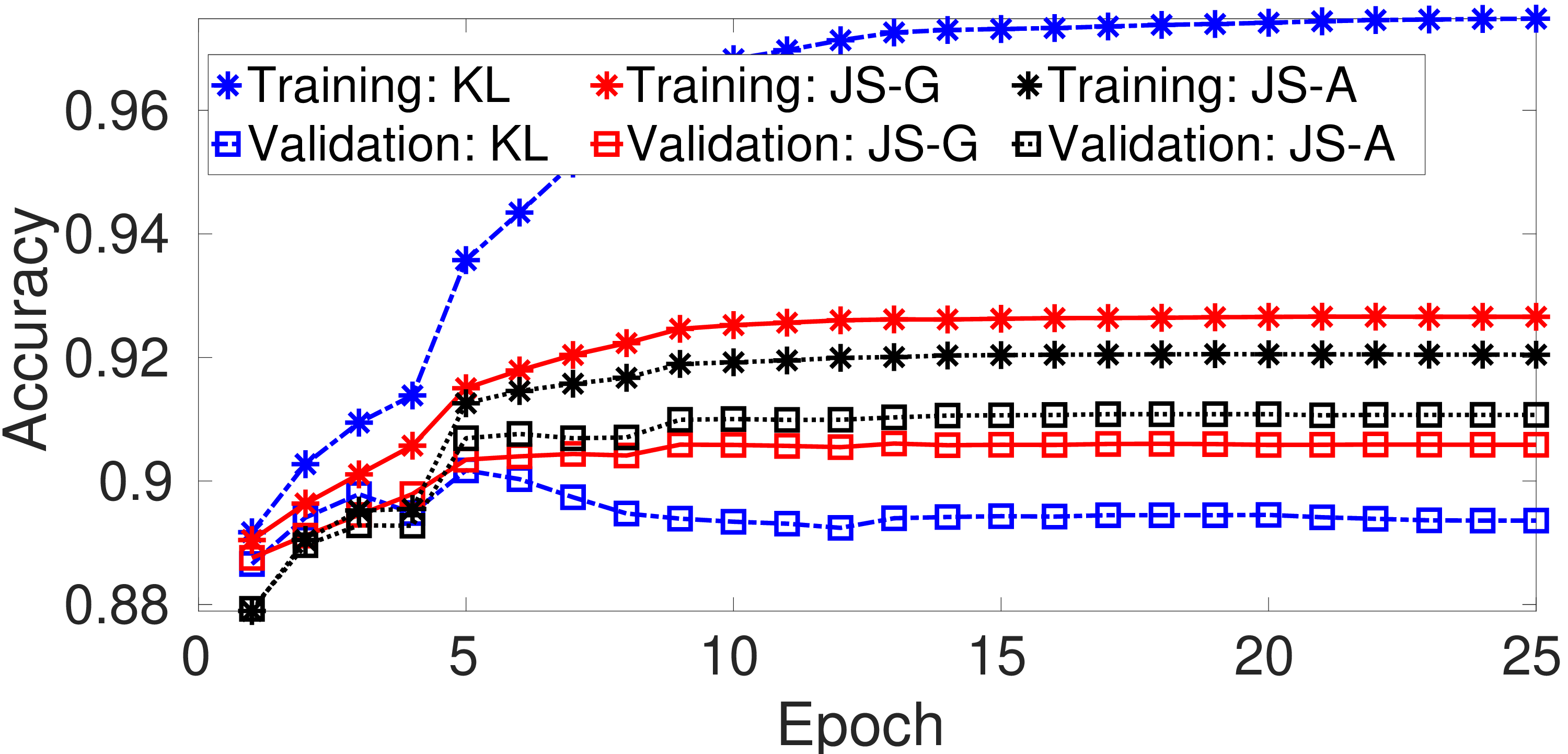}\label{fig:training_histo}}
    \caption{Training and validation of (a) CIFAR-10 with added Gaussian noise (b) histopathology data set with bias.}
\end{figure}
Training of the CIFAR-10 data set is performed with varying levels of noise intensity. Accuracy of training and validation sets for noise $\mathcal{N}(\mu = 0,\sigma = 0.9)$ is presented for both KL loss and the proposed JS  losses in Fig.~\ref{fig:training_Cifar}. For the histopathology data set, a learning rate scheduler is used during training in which the learning rate is multiplied by a factor of 0.1 in the 4th, 8th, 12th, and 20th epochs.  Fig.~\ref{fig:training_histo} shows the accuracy of training and validation of the histopathology set for the KL loss and the proposed JS losses. It is evident that the KL loss learns the training data too well and fails to generalize for the unseen validation set on both data sets. Whereas, the proposed JS losses regularise better and provide more accurate results for the validation set.
\subsection{Testing}
Results obtained on the test sets of the CIFAR-10 data set and the histopathology data set are presented in this section. The test results correspond to the epoch in which the validation accuracy was maximum.
\begin{figure}[htpb]
    \centering
    \subfigure[][]{\includegraphics[width=0.7\linewidth]{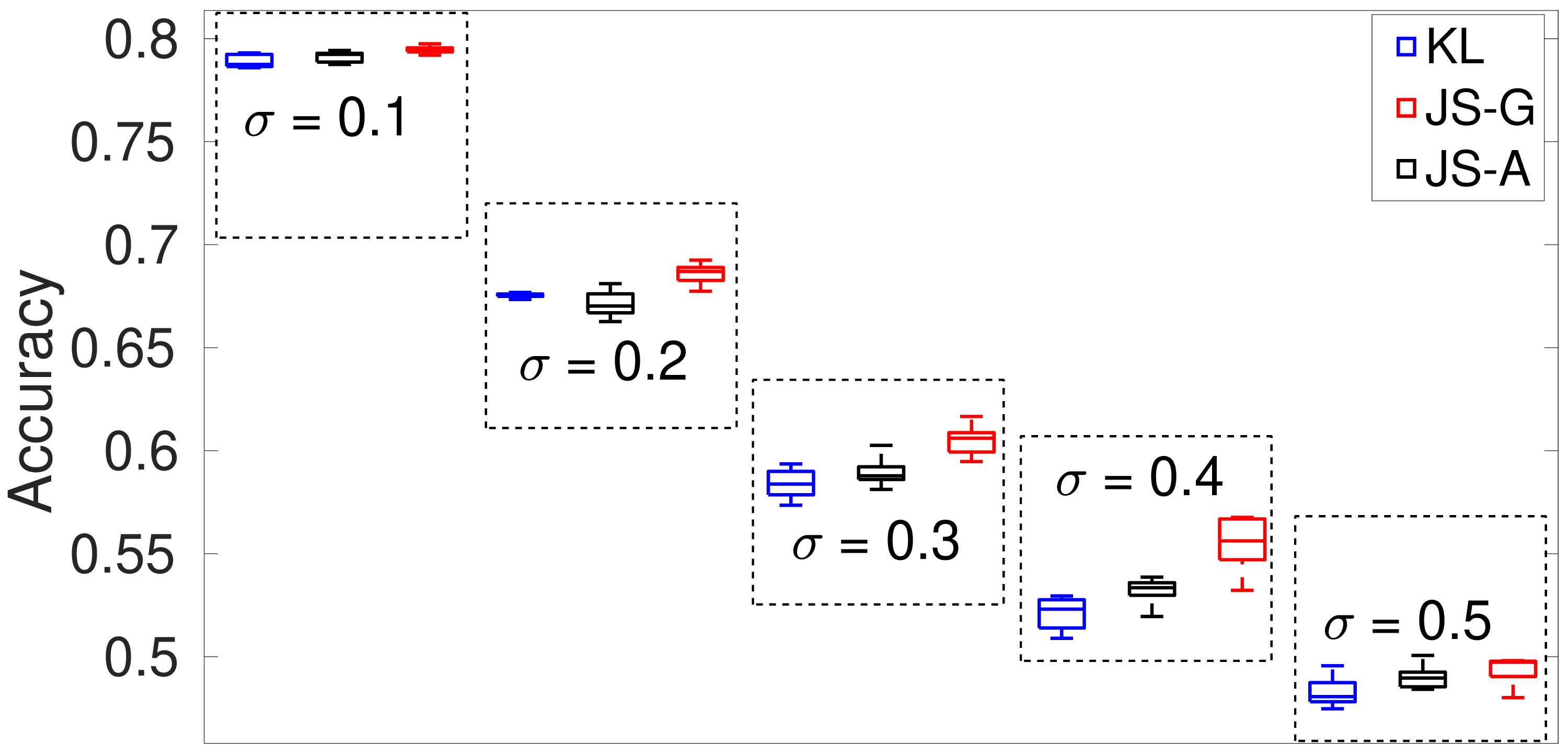}\label{fig:pt1_pt5}} \hfill
    \subfigure[][]{\includegraphics[width=0.7\linewidth]{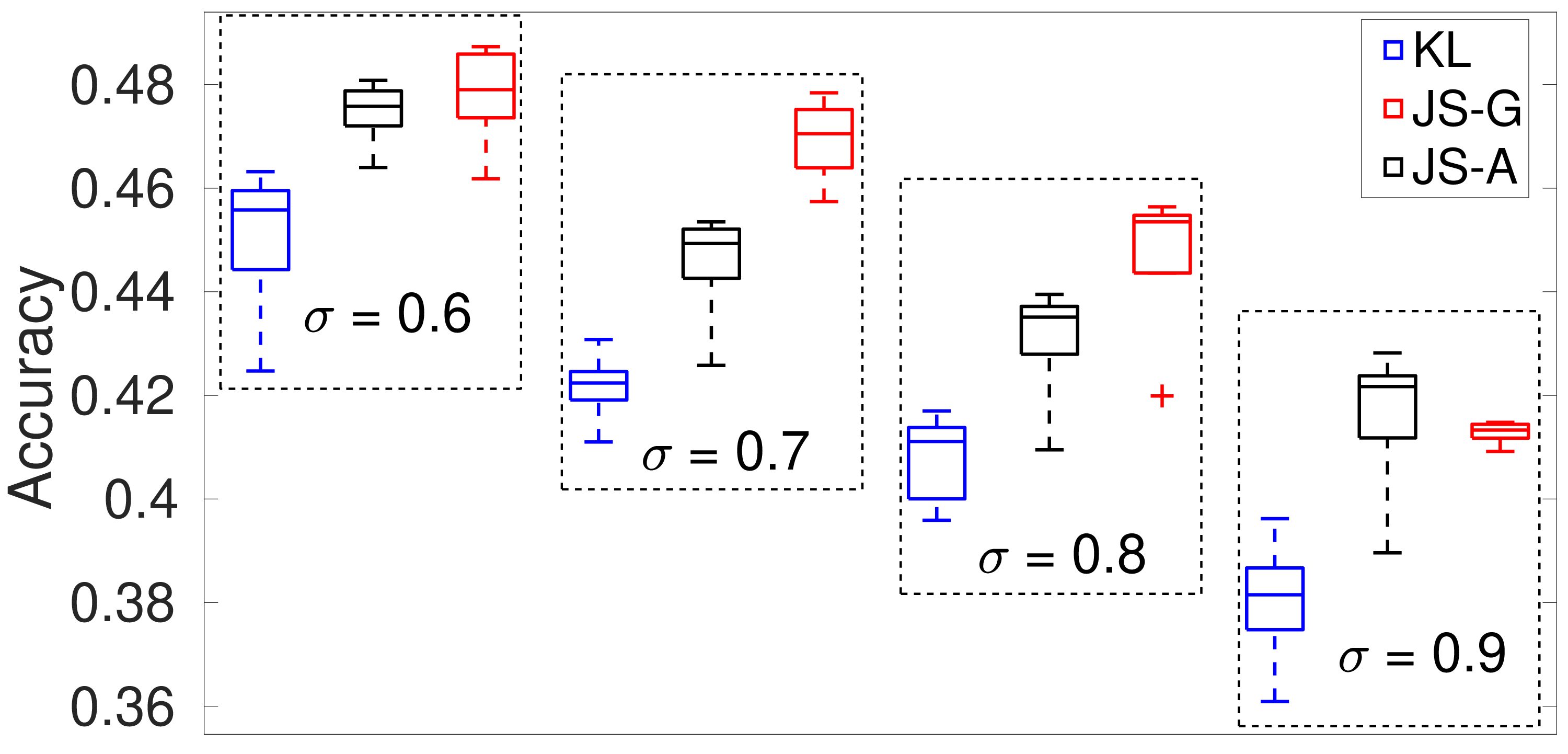}\label{fig:pt6_pt9}}
    \subfigure[][]{\includegraphics[width=0.45\linewidth]{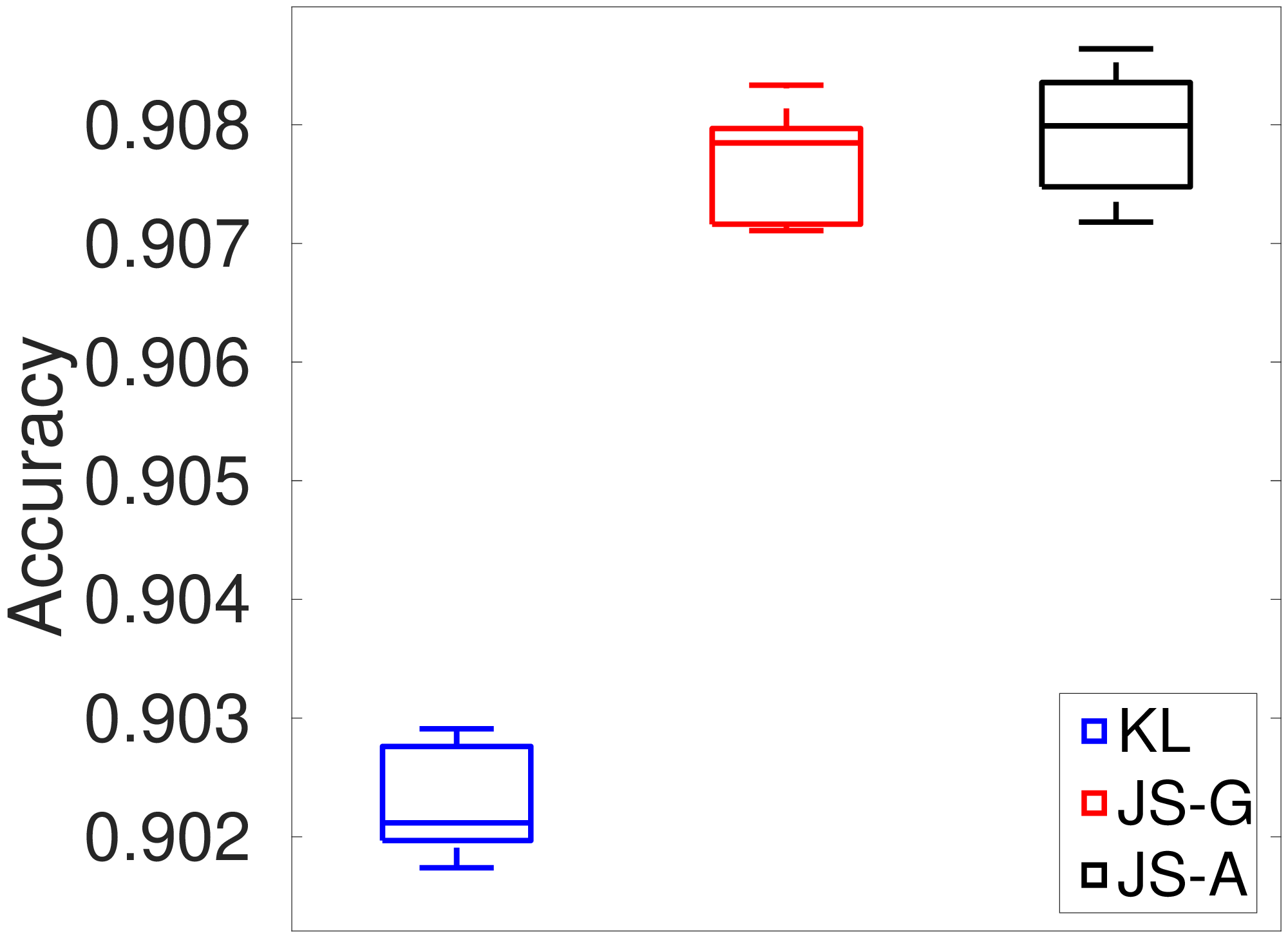}
    \label{fig:acc_histo}}
    \subfigure[][]{\includegraphics[width=0.45\linewidth]{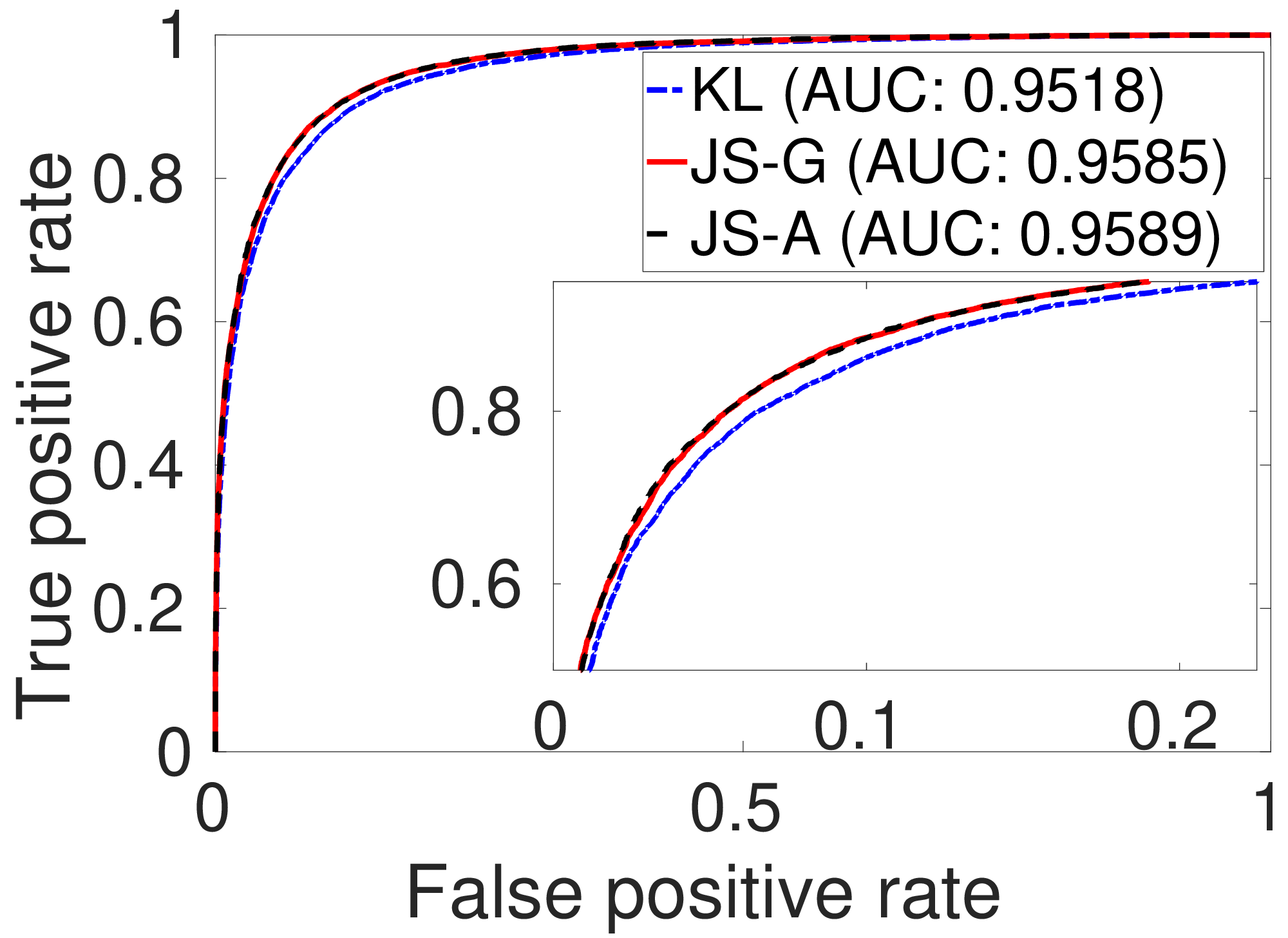}\label{fig:roc}}     \hfil    
    \caption{Accuracy on (a-b) the CIFAR-10 test data at different noise levels (c) histopathology test data. Each box chart displays the median as the center line, the lower and upper quartiles as the box edges, and the minimum and maximum values as whiskers.  (d) ROC curves, (e)-(g) Confusion matrices for histopathology data.}
\end{figure}
\begin{figure}
    \centering
    \subfigure[][KL]{\includegraphics[width=0.2\linewidth]{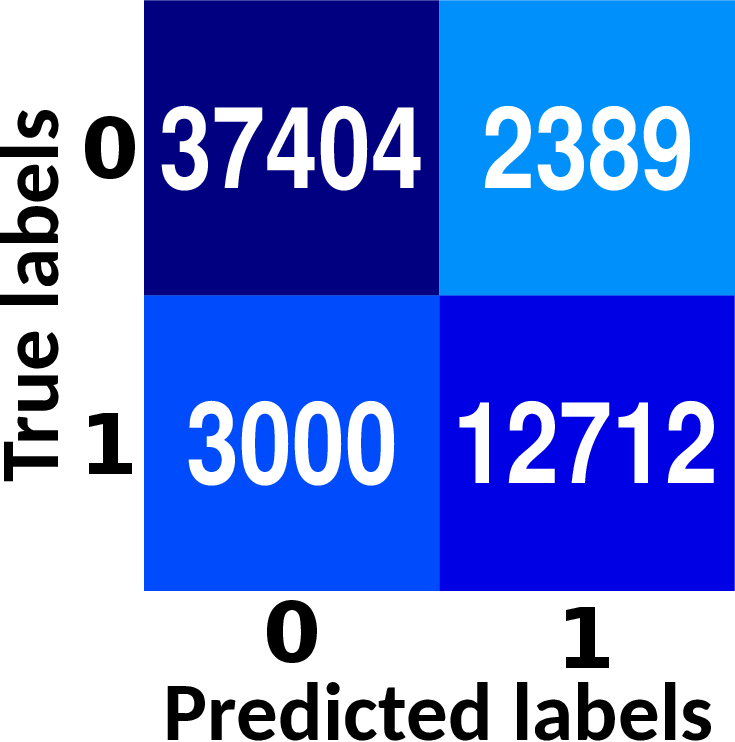}\label{fig:confa_hist}} \hfill
    \subfigure[][JS-G]{\includegraphics[width=0.2\linewidth]{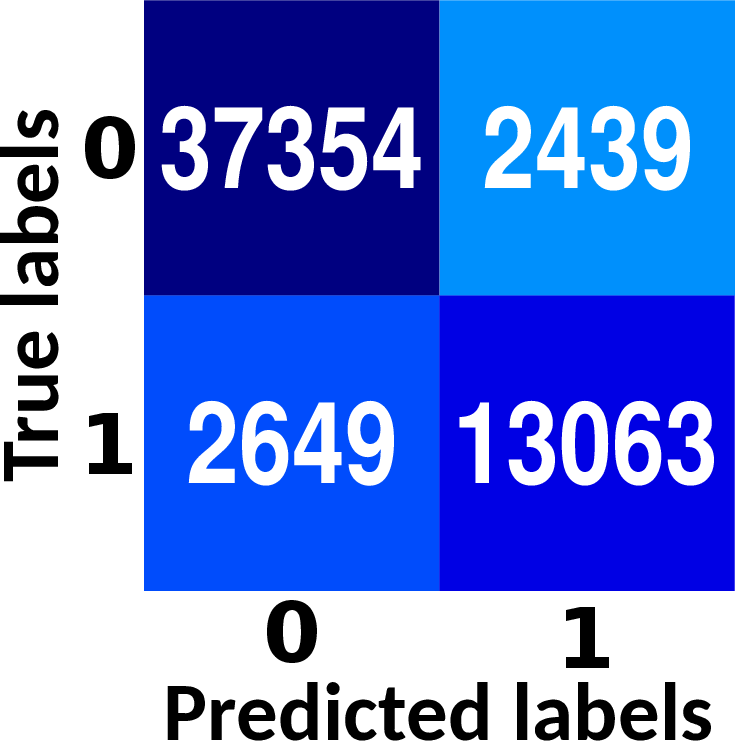}\label{fig:confb_hist}} \hfill
    \subfigure[][JS-A]{\includegraphics[width=0.2\linewidth]{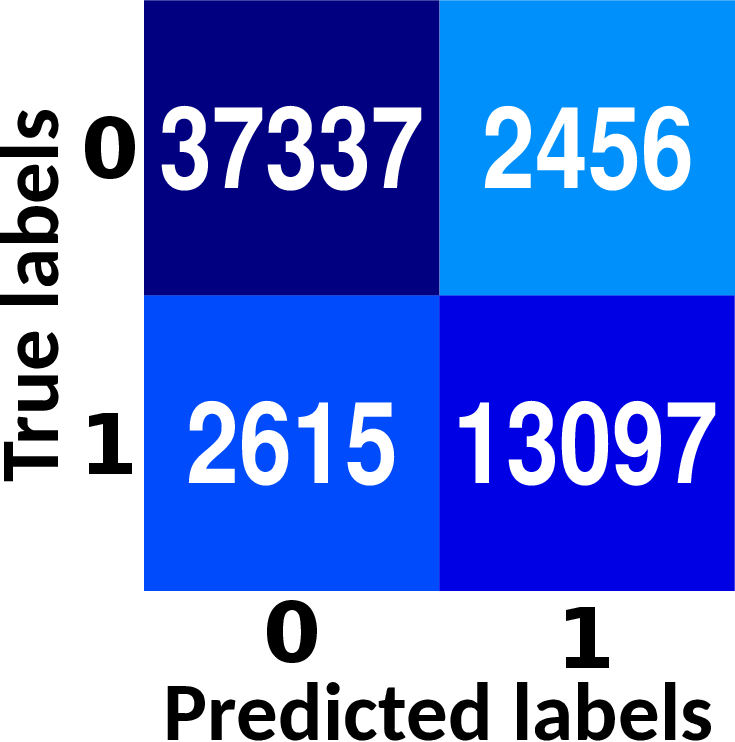}\label{fig:confc_hist}} 
    \caption{Confusion matrices for the histopathology dataset}
\end{figure}
Five runs were performed with different mutually exclusive training and validation tests to compare the results of the KL loss and the proposed JS losses.
The accuracy of the noisy CIFAR-10 test data set at varying noise levels is presented in Fig.~\ref{fig:pt1_pt5} and Fig.~\ref{fig:pt6_pt9}.
It is evident that the accuracy of both the proposed JS losses is better than KL for all the noise level cases. Further, the difference in accuracy between KL loss and the JS losses shows an increasing trend with increasing noise levels. This demonstrates the regularising capability of the proposed JS losses.
The results of the five runs of the KL loss and the proposed JS losses on the biased histopathology data set are compared in Fig.~\ref{fig:acc_histo}. It is evident that both the proposed JS losses perform better than the KL loss in all five runs with different training and validation sets. Since this data set is biased toward the negative class, the performance improvement shown by the proposed JS losses is attributed to better regularisation and generalization capabilities of the loss functions.

The receiver operating characteristic (ROC) curve is plotted in Fig.~\ref{fig:roc} for the classification of the histopathology data set. The proposed JS losses perform better than the KL loss in terms of the area under the curve. 
The confusion matrices in Fig.\ref{fig:confa_hist}-\ref{fig:confc_hist} show that in addition to improving the accuracy of predictions, the proposed JS-G and the JS-A losses reduce the number of false negative predictions by 11.7\% and 12.8\% respectively, as compared to the KL loss. Given that the data set is biased towards the negative class, this is a significant achievement.

\subsection{Comparison against baselines} \label{app:exp}
We performed additional experiments to evaluate the performance of the proposed loss functions against various baselines. The results of these experiments are presented here.
\subsubsection{Deterministic networks}
\begin{figure}[h]
    \centering
    \includegraphics[width = 0.9\linewidth]{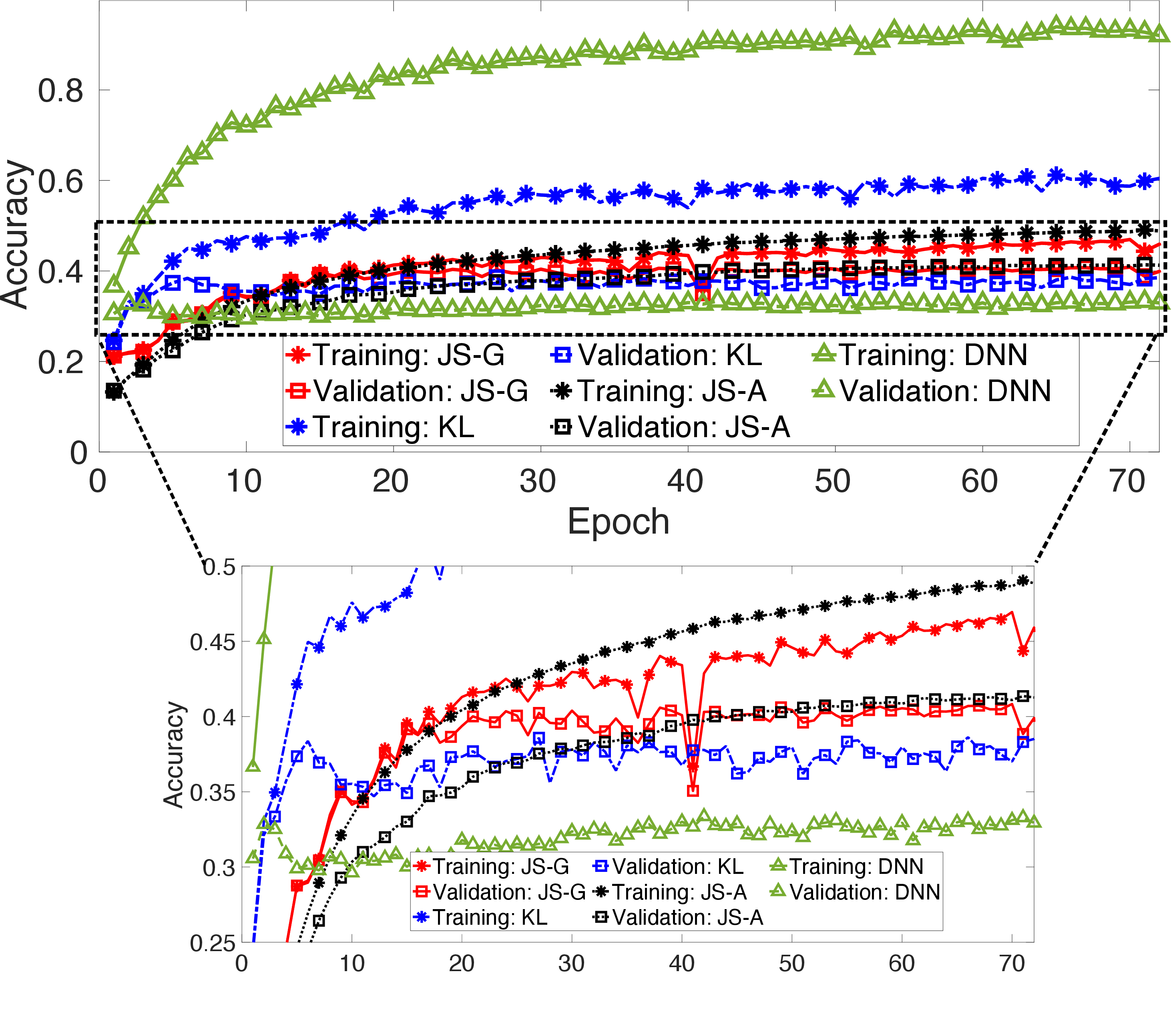}
    \caption{Performance of a deterministic CNN on CIFAR-10 dataset with added Gaussian noise $\mathcal{N}(\mu = 0,\sigma = 0.9)$ and its magnified view. }
    \label{fig:dnn_Cifar}
\end{figure}

Bayesian CNNs with the proposed loss functions are compared with a deterministic CNN of the same architecture and the results are presented in Fig.~\ref{fig:dnn_Cifar}. The deterministic CNN significantly overfits the training data and hence generalizes poorly for the validation dataset. Whereas, the Bayesian CNNs with their stochastic parameters can regularize much better than their deterministic counterparts. The validation accuracy of the deterministic CNN is about 8\% lower than the proposed BNNs and 5\% lower than the KL-based BNN. The difference between the training and validation accuracies are 63\% and 86\% less in KL- and JS-based BNNs respectively compared to the DNN, affirming greater generalization by KL- and JS-based BNNs.

\subsubsection{$\lambda$ KL baseline}
\begin{figure}[h]
    \centering
    \includegraphics[width = 0.9\linewidth]{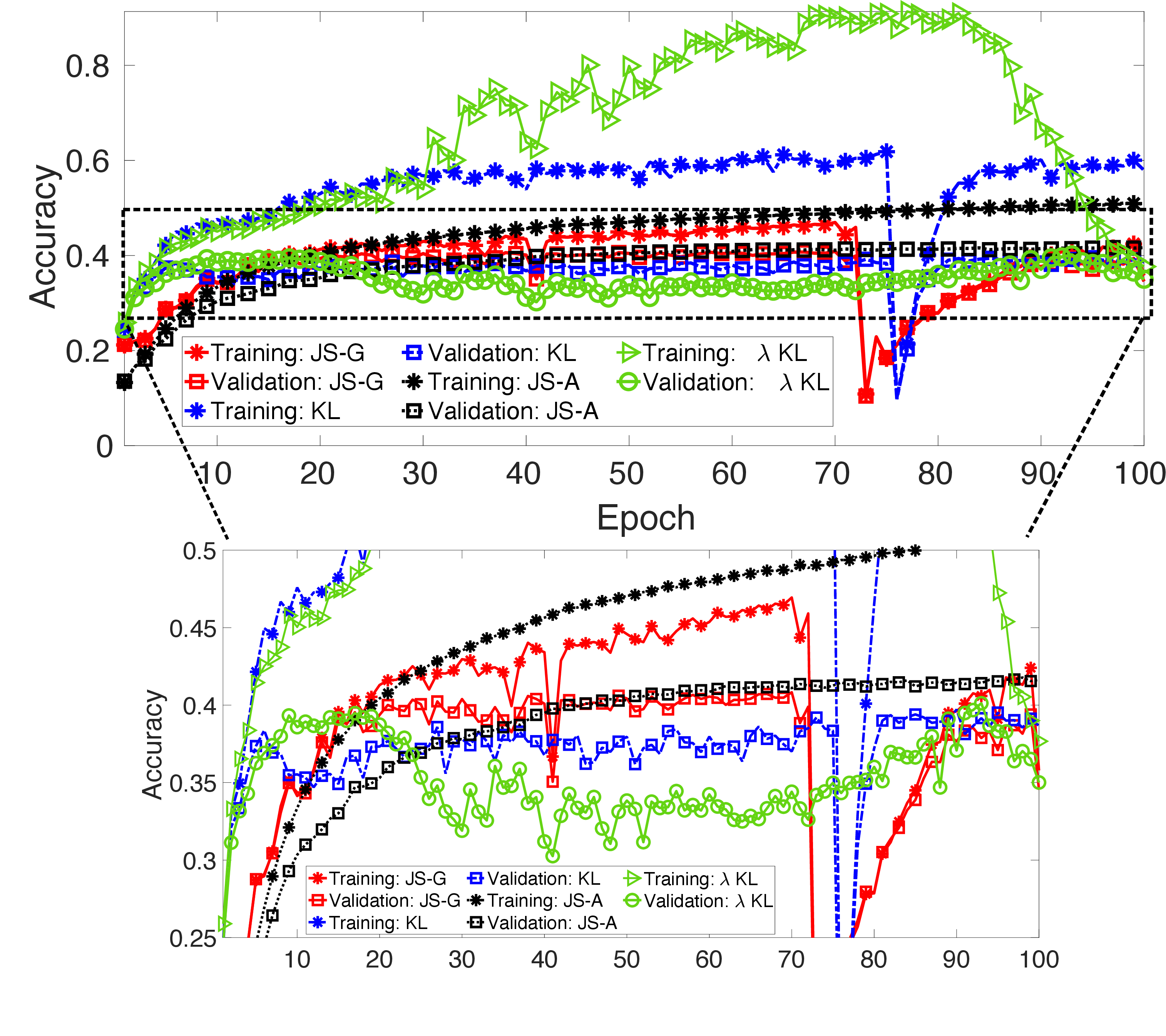}
    \caption{Performance of the $\lambda$ KL baseline for $\lambda = 68.9$ with a magnified view.}
    \label{fig:train_lamdakl}
\end{figure}

Implementing KL divergence in the constrained optimization framework (Eq.~\ref{eq:cost_fn}) yields a loss function which is henceforth called the $\lambda$KL baseline. We performed hyperparameter optimization on this baseline to tune the hyperparameters. The results of the $\lambda$KL baseline with the best-performing hyperparameters are presented in Fig.~\ref{fig:train_lamdakl}. We found that the maximum validation accuracy is 40.12\% for this baseline on the CIFAR-10 images with sigma=0.9. This validation accuracy is higher than the ELBO loss, which is 39.72 \%, which corresponds to lambda=1. However, it is still less than the proposed JS loss, which has a maximum validation accuracy of 41.78\%. In addition, for the lambda KL baseline, the regularization performance improves with lambda=68.9 although there are broad regions (between 20 to 90 epochs) where the network overfits. 
\\The improved performance of JS divergences is a result of optimal penalization when the posterior is away from the prior. The functional form of the regularization term is now changed since the JS divergence is a weighted average of forward and reverse KL divergences. Through the alpha term of JS divergence, we can choose the optimal weights between the forward and reverse KL which is not present in the standard KL divergence. This combination of forward and reverse KL divergence allows us to alter the shape of the multi-dimensional regularization term by adapting to the data, which is not possible to achieve by scalar multiplication of the regularization term.  

\subsubsection{Unmodified JS divergences}
\begin{figure}[h]
    \centering
    \includegraphics[width = 0.9\linewidth]{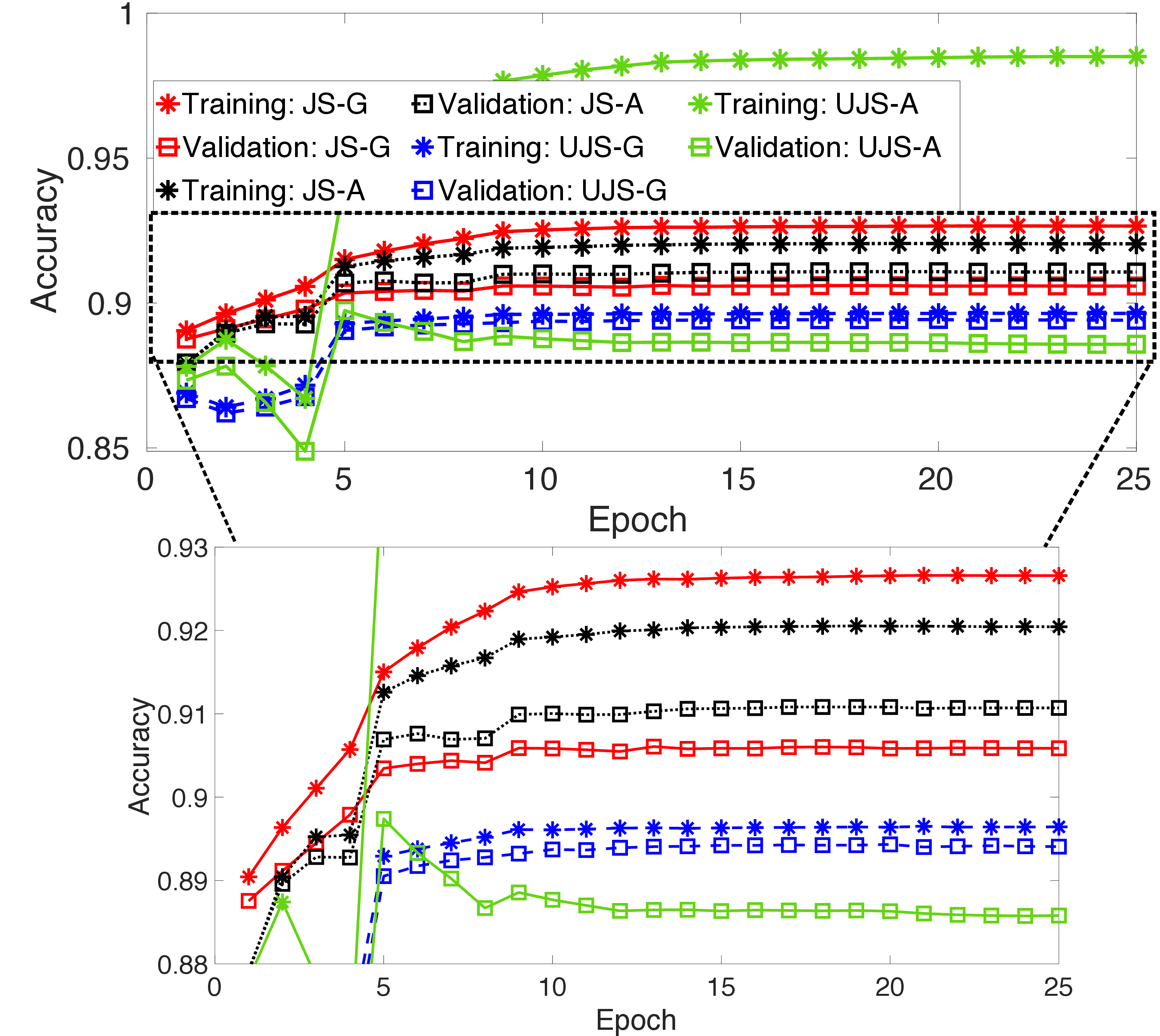}
    \caption{Performance of unmodified divergences (UJS) on histopathology with a magnified view.}
    \label{fig:train_unmod}
\end{figure}

The unmodified JS divergences in Eq.~\ref{eq:js_wadd} and Eq.~\ref{eq:jsgalph} fail to capture the dissimilarity between two distributions in the limiting cases of $\alpha$ as explained in the main text. However, we implemented these unmodified JS divergences in the loss function in Eq.~\ref{eq:cost_fn} and compared their performance with the modified versions for completeness. From Fig.~\ref{fig:train_unmod} it is evident that the modified JS divergences-based losses outperform the unmodified JS divergences-based losses on the histopathology dataset. The validation accuracy is less by about 3\% and 2\% for the unmodified JS divergences in Eq.~\ref{eq:js_wadd} and Eq.~\ref{eq:jsgalph} as compared to the modified ones respectively.

\subsection{CIFAR-100 dataset}
The CIFAR-100 data set \cite{krizhevsky2009learning} consists of 60,000 images of size $32 \times 32 \times 3$ belonging to 100 mutually exclusive classes. Experiments on this dataset were carried out to evaluate the performance of the loss functions. The images were normalized using the min-max normalization technique and the training data set was split into 80\% 20\% for training and validation respectively. For this dataset, a ResNet18-V1 type architecture without the batch norm layers was used where the first two layers ( convolution and max pooling layer) are replaced with a single convolution layer with 3×3 kernel and 1×1 stride.

\begin{figure}[h]
    \centering
     \subfigure[][]{\includegraphics[width=0.8\linewidth]{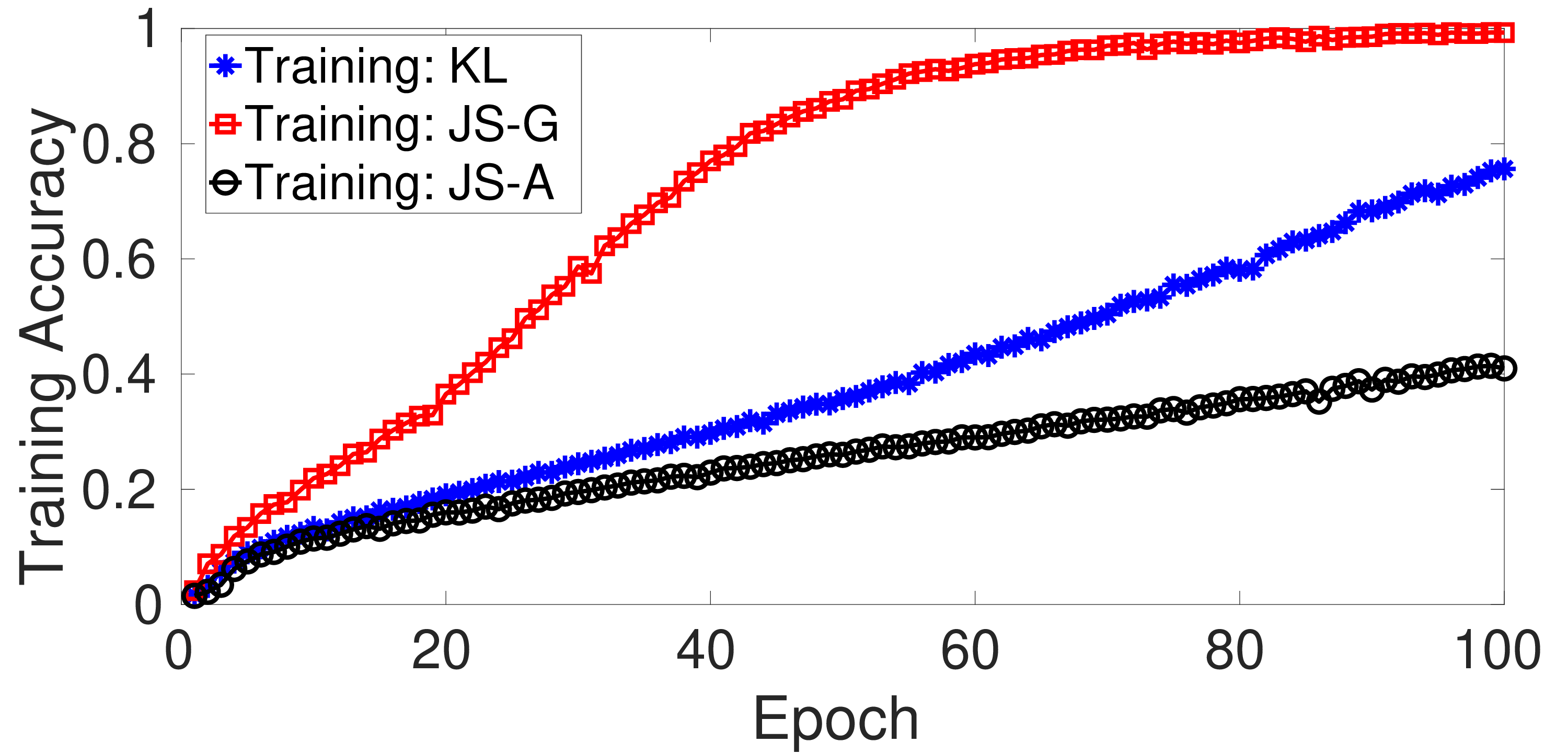}\label{fig:Cifar100_train}}
    \hfil
    \subfigure[][]{\includegraphics[width=0.8\linewidth]{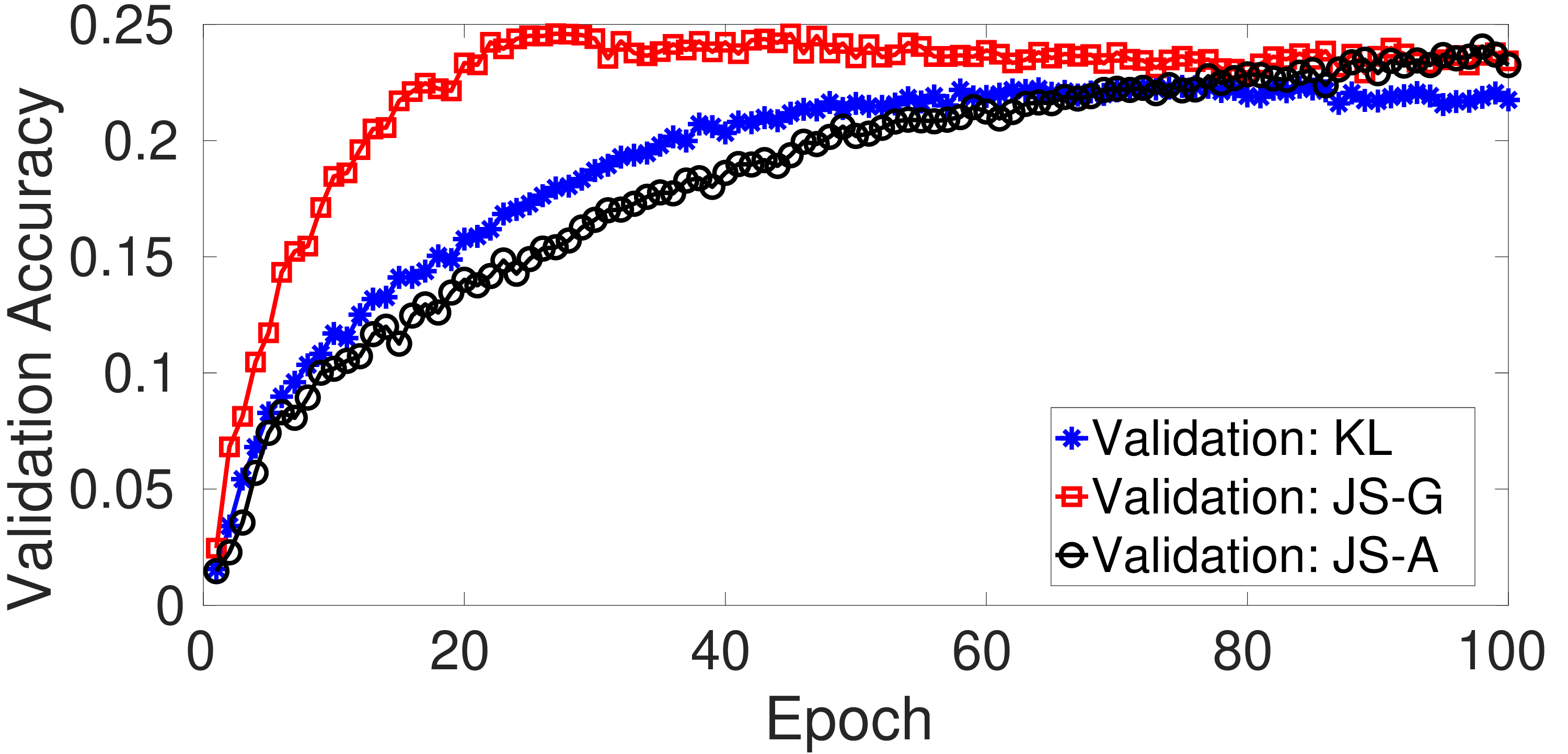}\label{fig:Cifar100_vld}}
    \caption{Performance comparison of the KL, JS-G, and JS-A divergence-based loss functions with CIFAR-100 dataset with added Gaussian noise $\mathcal{N}(\mu = 0,\sigma = 0.5)$. }
    \label{fig:Cifar100}
\end{figure}

The results of the experiment are presented in Fig.~\ref{fig:Cifar100}. The test accuracies of KL, JS-G, and JS-A divergence-based losses were 22.81 \%, 24.51 \%, and 24.02 \% respectively. Both the proposed JS divergence-based losses perform better than the KL loss in terms of test and validation accuracies. The regularization performance of the JS-A divergence was better than KL for this noise level. However, in terms of regularization the KL divergence performs better than the JS-G divergence for this dataset at the given noise level.  

\subsection{Computational cost} \label{app:cost}
All the convolutional neural networks presented in the paper were built on Python 3 using Apache MXNet with CUDA 10.2. Training, validation, and testing of all the networks were performed using the Nvidia Tesla V100 32GB GPUs.\\
A comparison of the computational time per epoch during training of the histopathology dataset is provided for the three loss functions in Table.~\ref{Tab:comp1}. The number of MC samples for the KL divergence and JS-A divergence-based loss functions was taken as 1. Thus, the computational effort is almost equal for the MC sampled KL divergence and the closed-form evaluated JS-G divergence. Whereas, to evaluate the JS-A divergence both the prior and the posterior distributions of the parameters need to be sampled. Due to this, there is an increased computational effort to evaluate the JS-A divergence-based loss function which is reflected in the increased computational time in Table.~\ref{Tab:comp1}
\begin{table}[h]
    \centering
    \caption{A comparison of computational time for the three loss functions.}
    \begin{tabular}{cc }
    \toprule
        \textbf{Divergence} & \textbf{Training Time per epoch (s)} \\ \midrule
        
        KL  &  1140 \\ \midrule 
        JS-G &  1168 \\ \midrule   
        JS-A & 1856  \\ 
        \bottomrule
    \end{tabular}
    \label{Tab:comp1}
\end{table}

\subsection{Comparison of numerical and theoretical results} \label{sec:app_evol}
\begin{figure}[h!]
    \centering
    \subfigure[][NLL]{\includegraphics[width=0.49\linewidth]{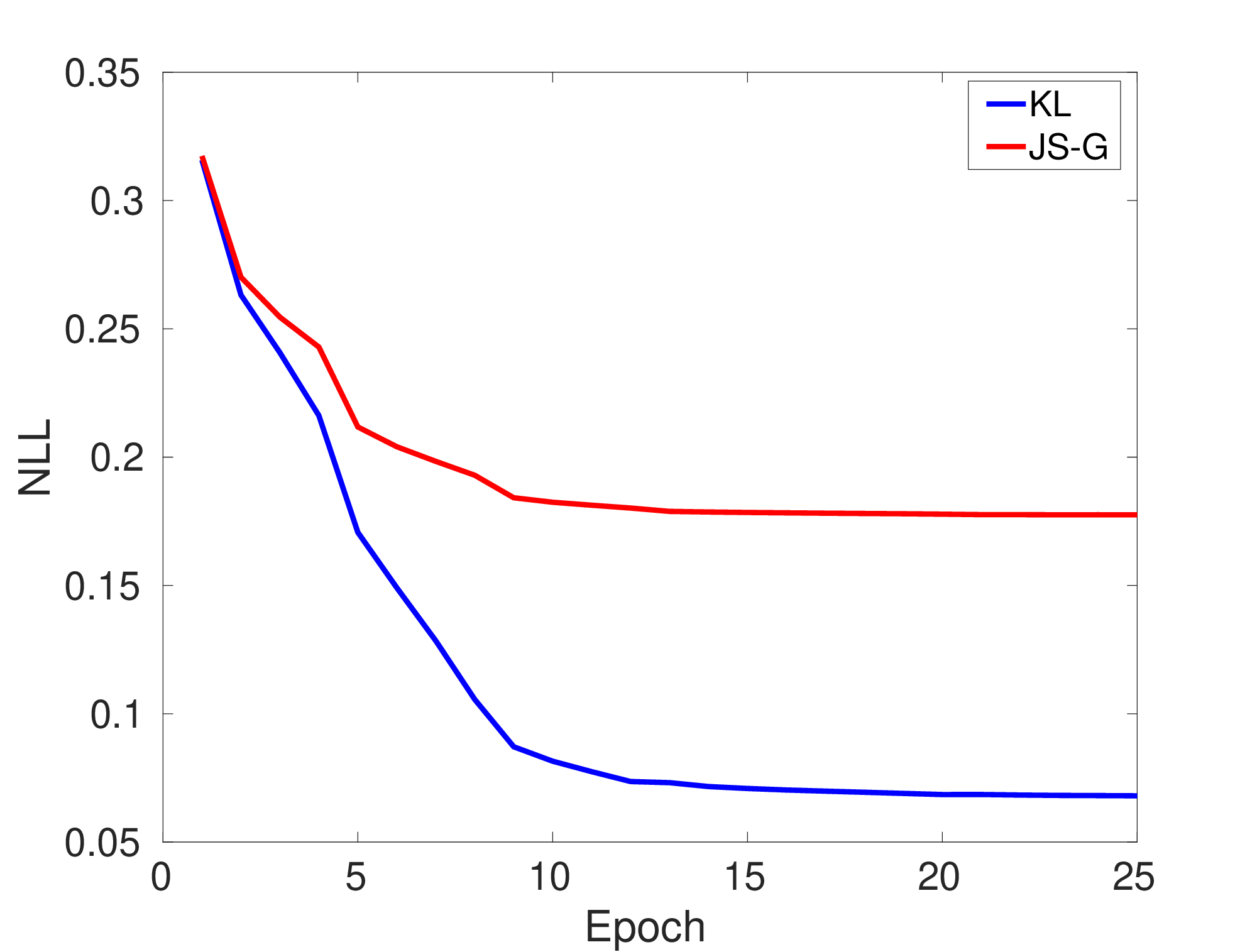}\label{fig:nll}}
    \hfil
    \subfigure[][Divergence Value]{\includegraphics[width=0.49\linewidth]{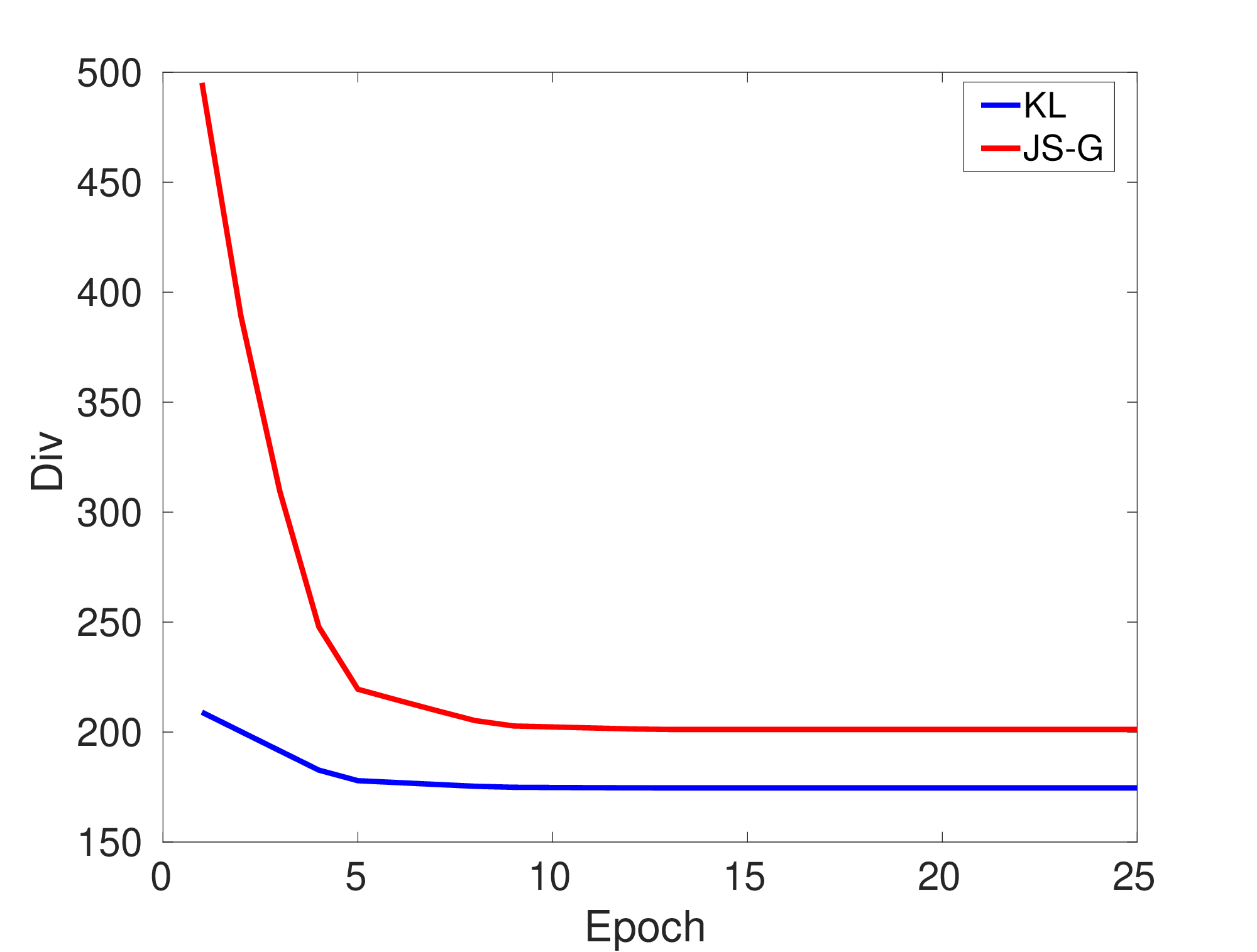}}
    \caption{Comparison of the evolution of training loss for the histopathology dataset}
    \label{fig:loss}
\end{figure}

The evolution of the divergence and negative log-likelihood part of the loss function during optimization is shown in Fig.~\ref{fig:loss} for the histopathology dataset. It is to be noted that the values are normalized by the number of test samples. The results are aligned with the theoretical results that the JS-G divergence penalizes higher than the KL divergence when the distribution q is farther from p. It is also seen that the negative log-likelihood is higher for the JS-G divergence as compared to the KL divergence in training. However, for the test dataset, the negative log-likelihood is lower for the JS divergences than the KL divergence which is desirable. 
\subsection{Comparing uncertainty metrics}
In this section, we compare the Test NLL, Test loss, and expected calibration error for the three losses for the histopathology dataset. From Table.~\ref{Tab:comp2} It is observed that the proposed JS divergences perform better in terms of both ECE and negative log-likelihood than the KL divergence for the histopathology dataset. Note that the test loss and test NLL are normalized by the number of test samples. The JS-A loss is significantly higher because of the value of $\lambda = 100$.
\begin{table}[h!]
    \centering
    \caption{Performance comparison for the three loss functions.}
    \begin{tabular}{cccc}
    \toprule
        \textbf{Divergence} & \textbf{Test NLL} & \textbf{Test loss} & \textbf{ECE} \\ \midrule
        
        KL  &  0.0810 & 177.6 & 0.0323\\ \midrule 
        JS-G &  0.0706 & 201.2 & 0.0158\\ \midrule   
        JS-A & 0.0689 &  15151.3 & 0.0091 \\ 
        \bottomrule
    \end{tabular}
    \label{Tab:comp2}
\end{table}

\section{Results on regression}
The setup for the regression experiments follows \citep{wan2020f,li2016renyi} in general. 20 trials are performed for each of the six regression datasets and the average test root mean squared error and average negative log-likelihood for the JS divergences are compared in Table~\ref{Tab:rmse} and \ref{Tab:nll} with various other divergence proposed for variational inference of BNNs in the literature \citep{wan2020f}. Since these datasets and the architecture do not demand regularization, the proposed JS divergence-based loss functions perform as good as the state-of-the-art methods or in some cases better. Competitive performances on these datasets demonstrate the versatility of the loss proposed loss functions. From these experiments, it is evident that the proposed loss functions can perform better than the state-of-the-start in applications requiring regularization and can perform as good as the state-of-the-art in cases where the need for regularization is absent.
\begin{table}[htpb]
    \centering
    \caption{Average root mean squared error. Except JS-G and JS-A all other values are taken from \citep{wan2020f}}
    \begin{tabular}{ccccccccc}
    \toprule
        \textbf{Data}& JS-G& JS-A & {KL-VI} & {$\chi$-VI} \\ \midrule
        Air &2.22$\pm$.25 & 2.32$\pm$.19  & \textbf{2.16$\pm$.07} & 2.36$\pm$.14 \\
        Aq & 1.63$\pm$.19 & 1.13$\pm$.13 & \textbf{1.12$\pm$.06} & 1.20$\pm$.06  \\
        Bos &3.34$\pm$.88 & 2.91$\pm$0.73 & \textbf{2.76$\pm$.36} & 2.99$\pm$.37  \\
        Con & 5.10$\pm$.67 &4.88$\pm$.63 & 5.40$\pm$.24 & \textbf{3.32$\pm$.34}  \\
        R.E & \textbf{6.77$\pm$1.1} & 7.37$\pm$2.3 & 7.48$\pm$1.4 & 7.51$\pm$1.4 \\
        Ya & 0.82$\pm$.35 & \textbf{0.69$\pm$.25} & 0.78$\pm$.12 & 1.18$\pm$.18 \\   \midrule
        \textbf{Data}& {$\alpha$-Vi} & TV-VI & $f_{c1}$-VI & $f_{c2}$-VI \\ \midrule
        Air  & 2.30$\pm$.08 & 2.47$\pm$.15 & 2.34$\pm$.09 & 2.16$\pm$.09\\
        Aq  & 1.14$\pm$.07 & 1.23$\pm$.10 & 1.14$\pm$.06 & 1.14$\pm$.06 \\
        Bos  & 2.86$\pm$.36 & 2.96$\pm$.36 & 2.87$\pm$.36 & 2.89$\pm$.38 \\
        Con  & 5.32$\pm$.27 & 5.27$\pm$.24 & 5.26$\pm$.21 & 5.32$\pm$.24 \\
        R.E & 7.46$\pm$1.4 & 8.02$\pm$1.6 & 7.52$\pm$1.4 & 7.99$\pm$1.6 \\
        Ya &  0.99$\pm$.12 & 1.03$\pm$.14 & 1.00$\pm$.18 & 0.82$\pm$.16 \\ 
        \bottomrule
    \end{tabular}
    \label{Tab:rmse}
\end{table}
\begin{table}[htpb]
    \centering
    \caption{Average negative log-likelihood. Except JS-G and JS-A all other values are taken from \citep{wan2020f}}
    \begin{tabular}{ccccccccc}
    \toprule
        \textbf{Data}& JS-G& JS-A & {KL-VI} & {$\chi$-VI}  \\ \midrule
        Air & 2.22$\pm$.09 & 2.72$\pm$.10 & \textbf{2.17$\pm$.03} & 2.27$\pm$.03  \\
        Aq & 1.94$\pm$.13& 1.78$\pm$.23& \textbf{1.54$\pm$.04} & 1.60$\pm$.08  \\
        Bos &2.69$\pm$.31 &3.3$\pm$1.12 & 2.49$\pm$.08 & 2.54$\pm$.18  \\
        Con & 3.1$\pm$.17 & 3.11$\pm$.20 & 3.10$\pm$.04 & \textbf{2.61$\pm$.18} \\
        R.E & \textbf{3.49$\pm$.06} & 3.56$\pm$.20 & 3.60$\pm$.30 & 3.70$\pm$.45 \\
        Ya & 1.51$\pm$.15 & \textbf{1.43$\pm$0.09} & 1.70$\pm$.02 & 1.79$\pm$.03 \\ \midrule
        \textbf{Data}& {$\alpha$-Vi} & TV-VI & $f_{c1}$-VI & $f_{c2}$-VI \\ \midrule
        Air & 2.26$\pm$.02 & 2.28$\pm$.04 & 2.29$\pm$.02 & 2.18$\pm$.03 \\
        Aq  & 1.54$\pm$.07 & 1.56$\pm$.07 & 1.54$\pm$.06 & 1.55$\pm$.04 \\
        Bos & \textbf{2.48$\pm$.13} & 2.51$\pm$.18 & 2.49$\pm$.13 & 2.51$\pm$.10 \\
        Con &  3.09$\pm$.04 & 3.10$\pm$.05 & 3.09$\pm$.03 & 3.10$\pm$.04\\
        R.E &  3.59$\pm$.32 & 3.86$\pm$.52 & 3.62$\pm$.33 & 3.74$\pm$.37\\
        Ya &  1.82$\pm$.01 & 1.78$\pm$.02 & 2.05$\pm$.01 & 1.86$\pm$.02\\
        \bottomrule
    \end{tabular}
    \label{Tab:nll}
\end{table}

\section{Limitations and Future work}
The proposed losses have two additional hyperparameters ($\alpha$ and $\lambda$) that need to be optimized to realize their full potential in regularization performance, which increases computational expenses. This might limit the application of the proposed methods to tasks involving excessive computational cost, either due to large datasets or extensive models. However, when such expenses can not be afforded,  the parameters can be set to the fixed values $\alpha = 0$ and $\lambda = 1$ to recover the KL loss. 
As a future direction, the proposed loss functions can be used in an application involving adversarial training to see the effectiveness of the method where a high degree of noise is involved. Further, the uncertainty quantification aspects of the proposed loss functions can be extensively analyzed to compare them against the existing uncertainty estimates of machine learning models' predictions. 

\section{Conclusions}\label{sec:Conclusions}
We summarize the main findings of this work in the following. \emph{Firstly}, 
the proposed loss function using a novel bounded JS-A divergence resolves the issue of unstable optimization associated with KL divergence-based loss.
The proposed loss encompasses the KL divergence-based loss and extends it to a wider class of bounded divergences. 
\emph{Secondly}, we also introduced a loss function based on the JS-G divergence that can be evaluated analytically and hence is computationally efficient. 
\emph{Thirdly}, better regularization performance by the two proposed loss functions compared to the state-of-the-art is established analytically and/or numerically. \emph{Fourthly}, empirical experiments on the histopathology data sets having bias show about a 13\% reduction in false negative predictions, and the CIFAR-10 dataset with various degrees of added noise shows a 5\% improvement in accuracy over the state-of-the-art with the proposed losses. Similarly, experiments on standard regression datasets show an 8\% reduction in the root mean square error, in the best-case scenario, when using the proposed losses as compared to the KL loss.

\section*{Acknowledgement} This work was supported by grant DE-SC0023432 funded by the U.S. Department of Energy, Office of Science. This research used resources of the National Energy Research Scientific Computing Center, a DOE Office of Science User Facility supported by the Office of Science of the U.S. Department of Energy under Contract No.~DE-AC02-05CH11231, using NERSC awards BES-ERCAP0025205 and BES-ERCAP0025168. 
The authors thank the Superior HPC facility at MTU, and the Applied Computing and MRI GPU clusters at MTU for making available some of the computing resources used in this work. PT acknowledges the Doctoral Finishing Fellowship awarded by the Graduate School at Michigan Technological University.

\appendix

\section{Divergences} \label{app:div}
\begin{figure}[h]
    \centering
     \subfigure[][Distributions q and P]{\includegraphics[width=0.5\linewidth]{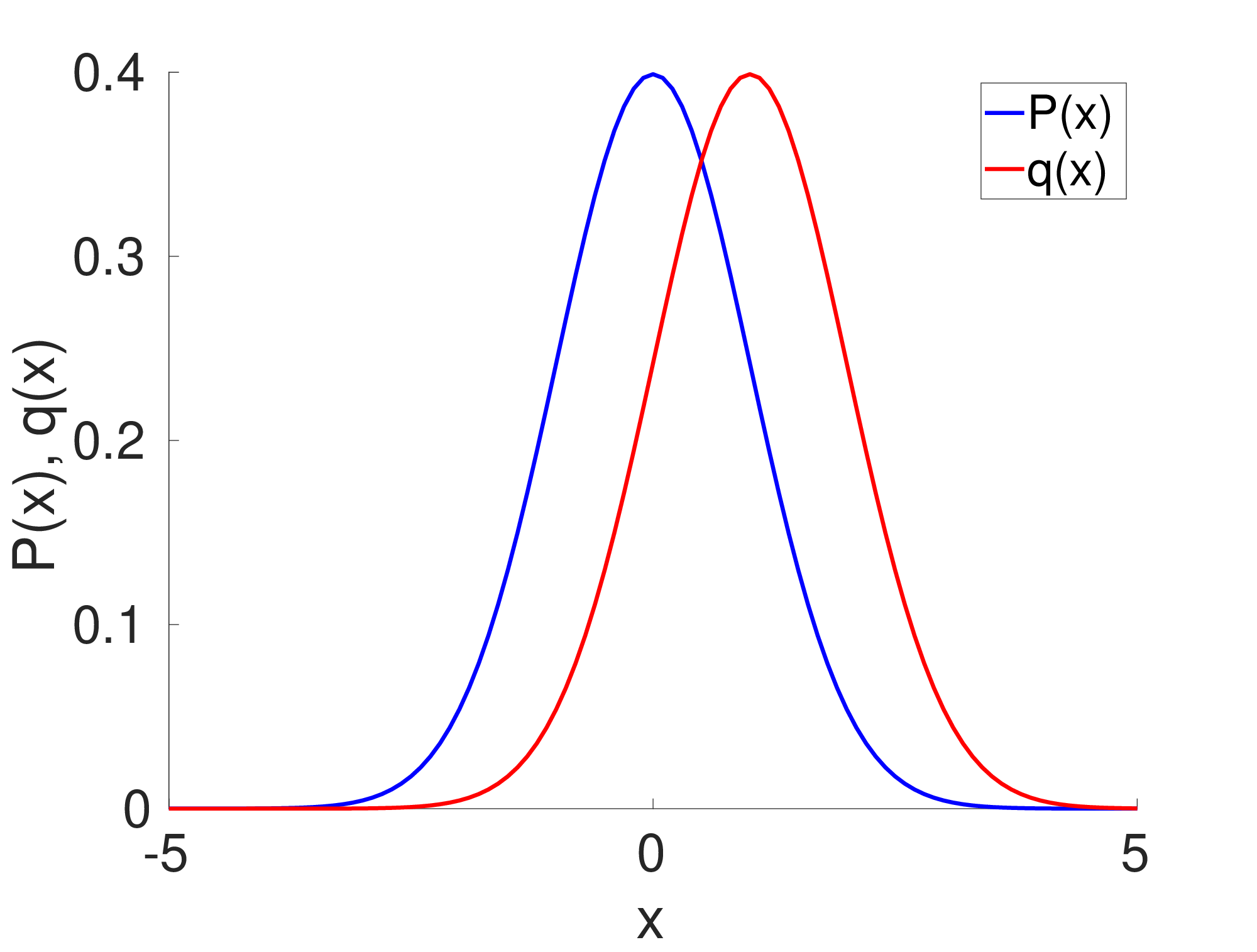}\label{fig:pandq_dist}}
    \hfil
    \subfigure[][Divergence shown as shaded regions.]{\includegraphics[width=0.5\linewidth]{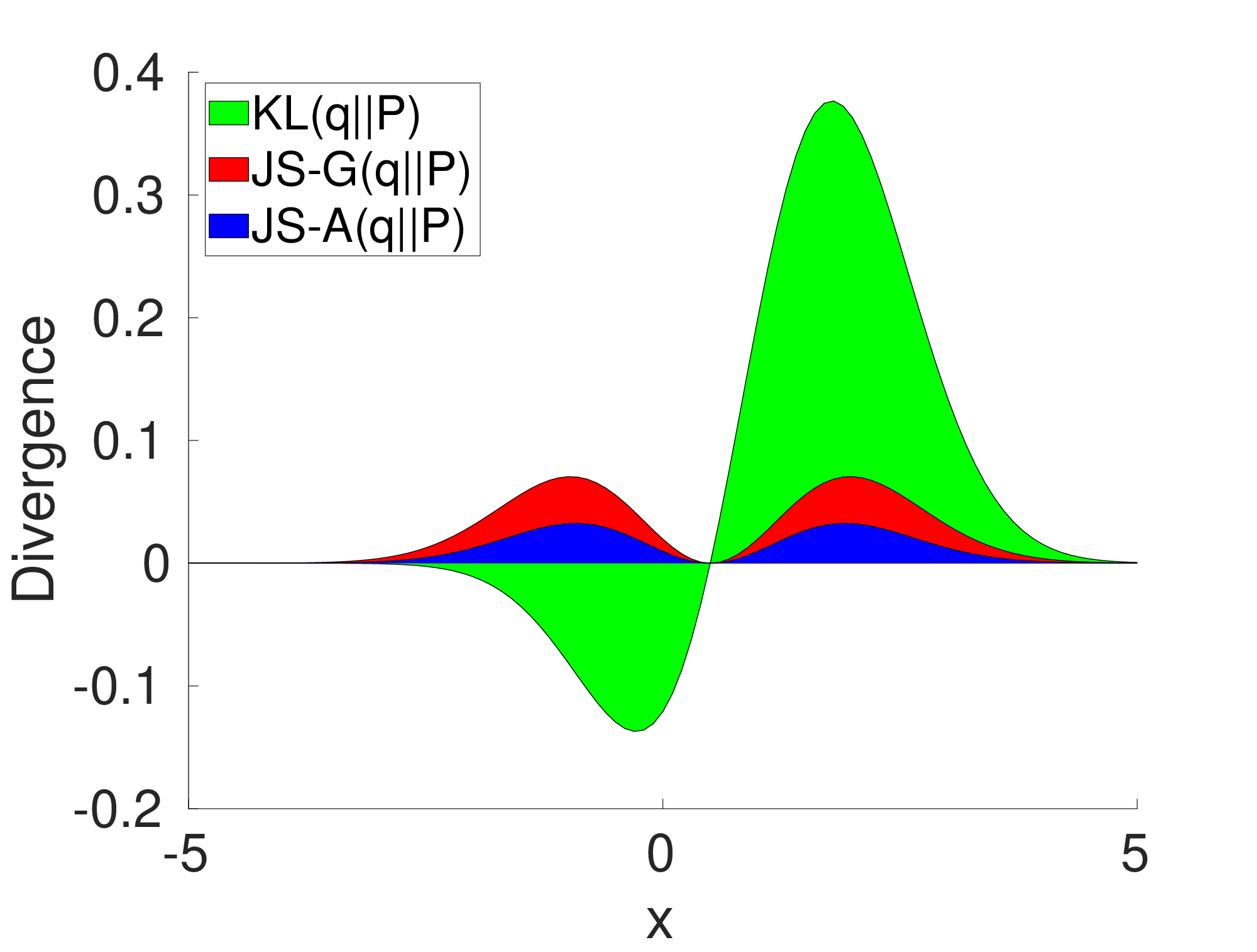}\label{fig:div}}
    \caption{Depiction of KL, JS-G, and JS-A divergences for two Gaussian distributions. The area under the curves is the value of the divergences.}
    \label{fig:divergences}
\end{figure}

Two distributions, $P(x) = \mathcal{N}(0, 1)$ and $q(x) = \mathcal{N}(1, 1)$ are shown in Fig.~\ref{fig:divergences} along with the divergence as a function of $x$. The area under the curve shown as a shaded region is the area to be integrated to obtain the divergence. 
We provide a comparison of symmetry (at $\alpha = 0.5$) and boundedness for divergences used in this work in Table \ref{tab:prop}
\begin{table}[htpb]
    \centering
    \caption{Properties of various divergences}
    \begin{tabular}{cccc} 
        \toprule
         \textbf{Div.} & \textbf{Bounded} & \textbf{Symmetric} & \textbf{Analytical exp.}\\ \midrule         
         KL & $\times$ & $\times$ & \checkmark \\ 
         JS-A & \checkmark & \checkmark & $\times$ \\
         JS-G &  $\times$& \checkmark & \checkmark \\
         \bottomrule
    \end{tabular}
    \label{tab:prop}
\end{table}

\section{Proof of Theorem 1} \label{sec:theorem1_proof}
\textbf{Theorem 1:} Boundedness of the modified generalized JS divergence\\
\textit{For any two distributions $P_1(t)$ and $P_2(t)$, $t \in \Omega$,  the value of the divergence $\text{JS-A}$ is bounded such that,}
\begin{align*}
&\text{JS-A}(P_1(t) || P_2(t))\leq -(1- \alpha) \log \alpha  -  \alpha  \log (1-\alpha), &\text{for } \alpha \in (0,1)
\end{align*}
\textbf{Proof: }

\begin{align*}
    &\text{JS-A} (P_1 || P_2) = (1-\alpha)  \int_\Omega P_1 \log \frac{P_1 }{A_\alpha' }dt + \alpha \int_\Omega P_2 \log \frac{P_2 }{A_\alpha' }dt \\
    &= \int_\Omega A_\alpha'  \left[ (1-\alpha)\frac{P_1}{A_\alpha' } \log \frac{ P_1 }{A_\alpha' } + \alpha \frac{P_2}{A_\alpha' } \log \frac{P_2 }{A_\alpha' }  \right]dt\\
    &= \int_\Omega A_\alpha'  \left[ \frac{(1-\alpha)}{\alpha}\frac{\alpha P_1}{ A_\alpha' } \left( \log \frac{\alpha P_1 }{A_\alpha' } - \log \alpha \right) \right. \\& + \left.  \frac{\alpha}{(1- \alpha)} \frac{ (1- \alpha) P_2}{A_\alpha' } \left( \log \frac{(1- \alpha)P_2 }{A_\alpha' } - \log (1- \alpha) \right)  \right] \: dt\\
    &= \int_\Omega -(1- \alpha) P_1 \log \alpha \: dt - \int_\Omega \alpha P_2 \log (1-\alpha) \: dt \: - \\&
     \int_\Omega A_\alpha' \left[\frac{(1-\alpha)}{\alpha}H\left(\frac{\alpha P_1}{ A_\alpha' }\right) + \frac{\alpha}{(1- \alpha)} H\left(\frac{ (1- \alpha) P_2}{A_\alpha' }\right)\right] dt \\
    &=  -(1- \alpha) \log \alpha  -  \alpha  \log (1-\alpha) - \mathcal{H}  \\
    &\leq  -(1- \alpha) \log \alpha  -  \alpha  \log (1-\alpha)
\end{align*}
Where, $H(f(t)) = - f(t) \log f(t)$ and \\
 $\mathcal{H} =  \int_\Omega A_\alpha' \left[\frac{(1-\alpha)}{\alpha}H\left(\frac{\alpha P_1}{ A_\alpha' }\right)+ \frac{\alpha}{(1- \alpha)} H\left(\frac{ (1- \alpha) P_2}{A_\alpha' }\right)\right] dt $\vspace{5pt} \\ 
Note, $ H(f(t)) \ge 0; \;\;$ $\forall f(t) \in [0,1]$. 
Therefore, $\mathcal{H} \ge 0; \;\;$ $\forall t \in \Omega$

\begin{figure}
    \centering
     \subfigure[][Distributions q and P]{\includegraphics[width=0.9\linewidth]{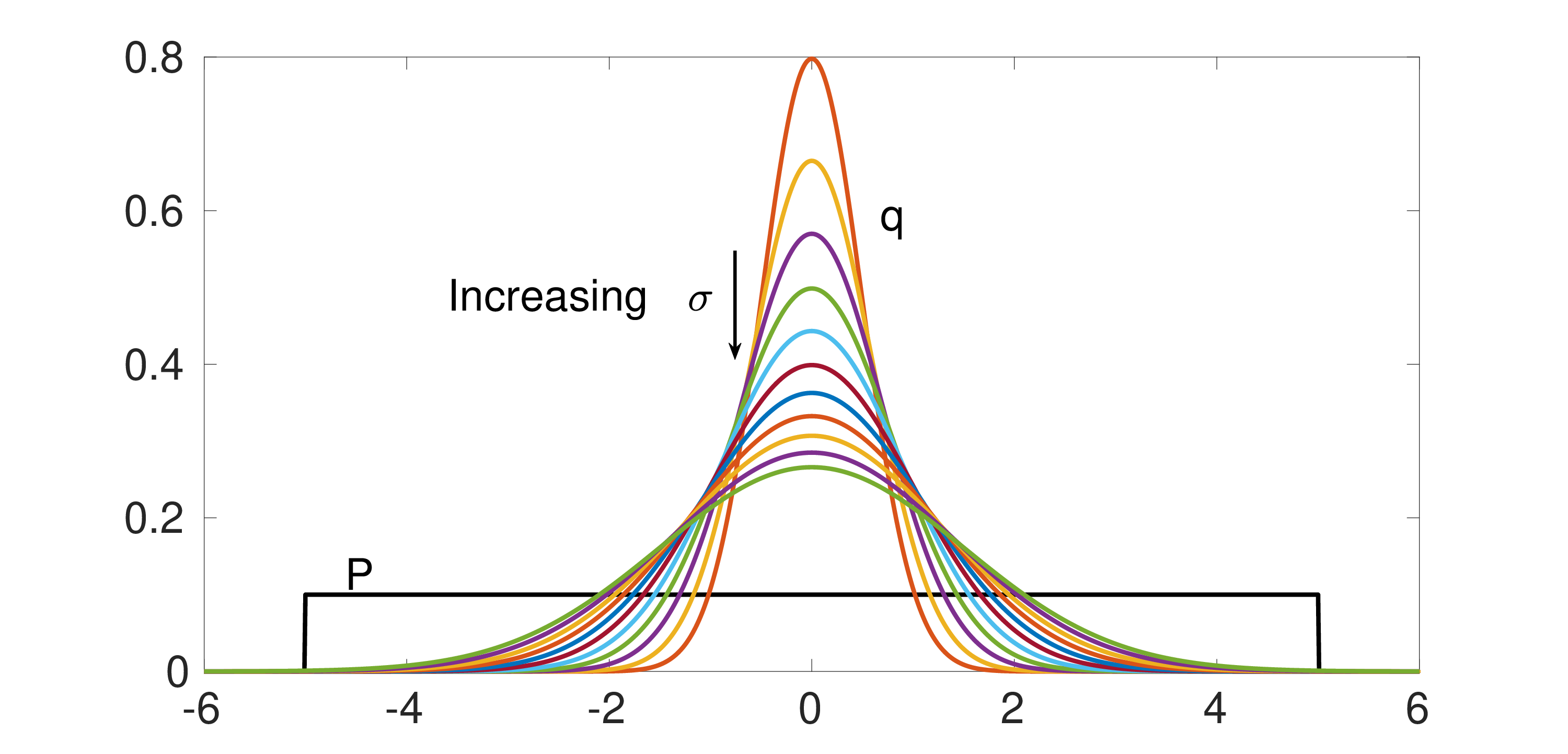}\label{fig:theorem1_dist}}
    \hfil
    \subfigure[][Divergence values for various $\sigma$]{\includegraphics[width=0.9\linewidth]{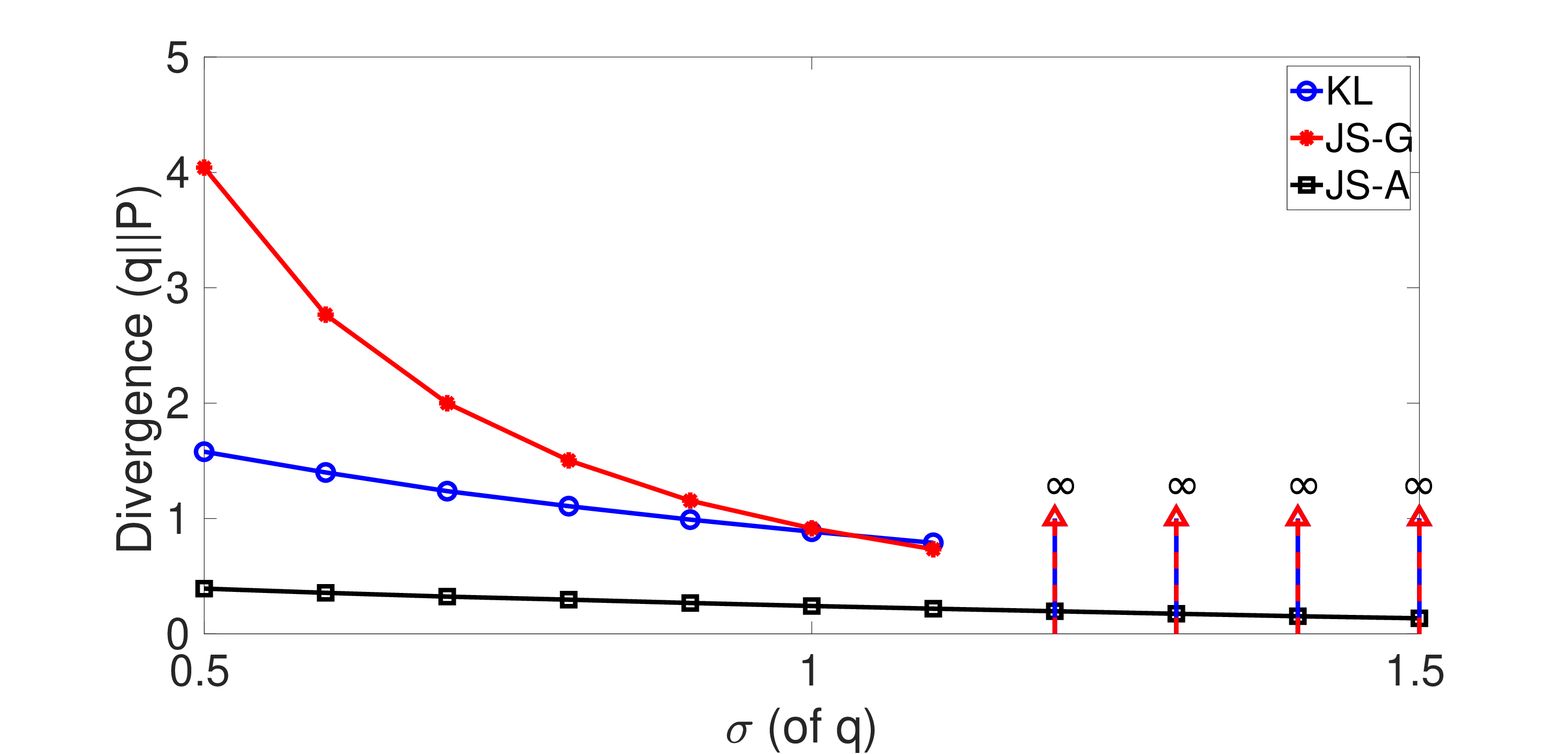}\label{fig:theorem1_div}}
    \caption{Depiction of the unboundedness of the KL and JS-G divergence and the boundedness of the JS-A divergence.}
    \label{fig:theorem1}
\end{figure}

The unboundedness of the KL and the JS-G divergences is depicted through a contrived example in Fig.~\ref{fig:theorem1}. The distribution $q$, where $q = \mathcal{N}(0,\sigma)$, is assumed to be a Gaussian with varying $\sigma$. The distribution $P$, where $P = \mathcal{U}(-5,5)$, is assumed to be a uniform distribution with support (-5,5). The distributions $q$ and $p$ are shown in Fig.~\ref{fig:theorem1_dist}. The KL, the JS-A, and the JS-G divergence of the two distributions $q$ and $P$ are evaluated using 50,000 Monte Carlo samples each. The value of the KL and the JS-G divergence explodes to infinity when the distribution $p$ is zero for a non-zero $q$ due to the effect of $\log (q/P)$. In contrast, the JS-G divergence is always bounded. This can be seen in Fig.~\ref{fig:theorem1_div}.

\begin{figure}
    \centering
    \includegraphics[width = 0.9\linewidth]{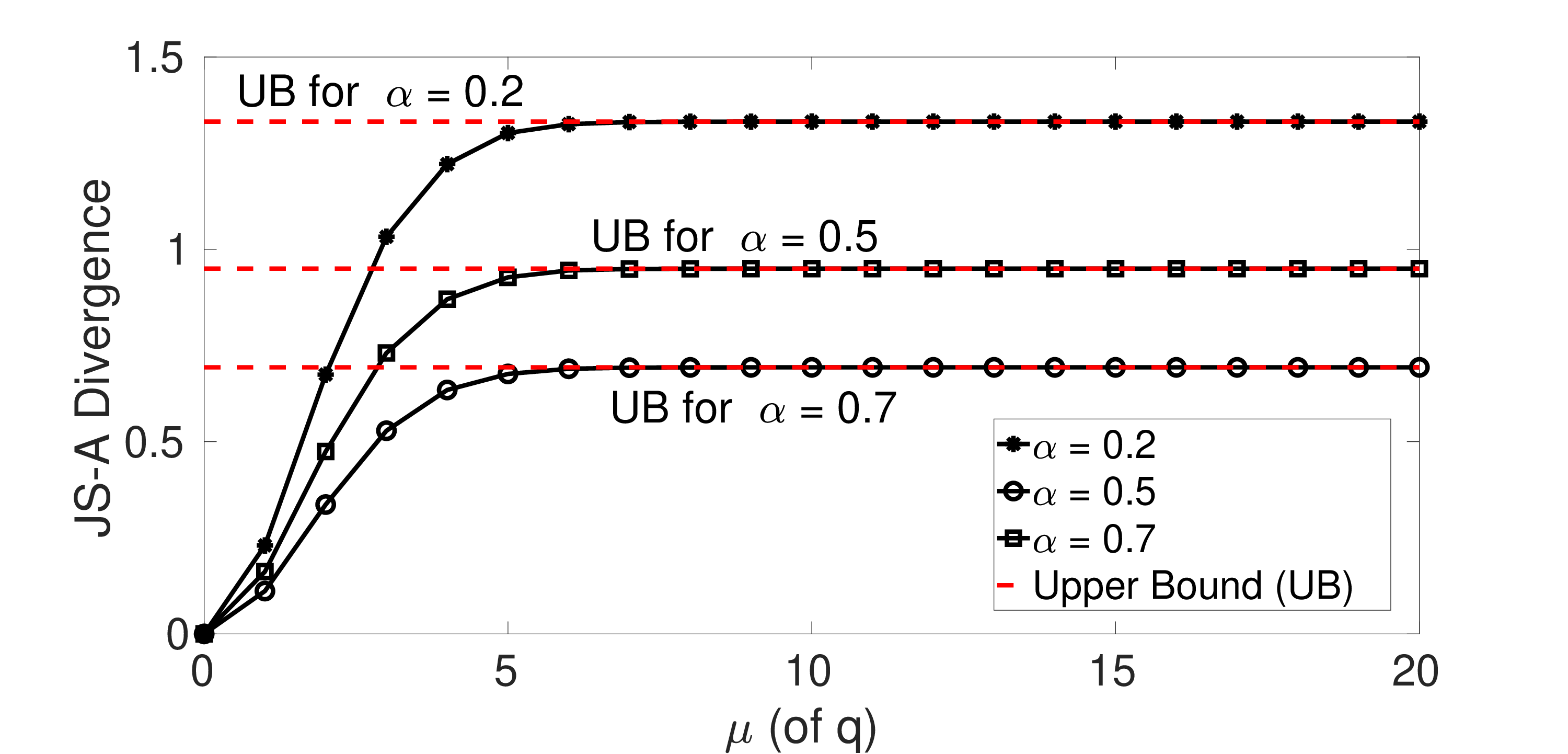}
    \caption{Upper bound of the JS-A divergence for various values of $\alpha$.}
    \label{fig:ub_jsa}
\end{figure}

Numerical examples depicting Theorem-1 are shown in Fig.~\ref{fig:ub_jsa}. Two distribution $q = \mathcal{N}(\mu,1)$ and $P = \mathcal{N}(0,1)$ are considered. The JS-A divergence is evaluated for varying $\mu$. The value of the JS-A divergence increases with increasing $\mu$ until the upper bound is reached and it remains constant henceforth as seen in Fig.~\ref{fig:ub_jsa}. 

\section{Intractability of JS-based VI } \label{sec:intract_vi}
The JS-G loss is given as 
\begin{equation}
\label{eq:elbo_js_app}
\begin{split}
&\mathcal{F}_{JSG}(\mathbb{D},\boldsymbol{\theta}) = \text{JS-G}\left[ q(\mathbf{w}|\boldsymbol{\theta})\:||\: P(\mathbf{w}|\mathbb{D}) \right]\\
    &\;=  (1-\alpha) \text{KL}\left(q\:||\: \text{G}'_\alpha(q,P)\right) + \alpha \text{KL}\left(P\:||\: \text{G}'_\alpha(q,P)\right)
\end{split}
\end{equation}
Where, $G'_\alpha (q,P) = q(\textbf{w}|\boldsymbol{\theta})^\alpha P(\textbf{w}|\mathbb{D})^{(1-\alpha)} $.\\
Rewriting the first term in Eq.~\ref{eq:elbo_js_app} as,
\begin{align}
    T_1 & = (1-\alpha) \text{KL}\left(q\:||\: \text{G}'_\alpha(q,P)\right) \nonumber\\
    &=(1-\alpha)KL[q(\textbf{w}|\boldsymbol{\theta})|| q(\textbf{w}|\boldsymbol{\theta})^\alpha P(\textbf{w}|\mathbb{D})^{(1-\alpha)}] \nonumber\\
        &=(1-\alpha)\int q(\textbf{w}|\boldsymbol{\theta})\log\left[ \frac{q(\textbf{w}|\boldsymbol{\theta})}{q(\textbf{w}|\boldsymbol{\theta})^\alpha P(\textbf{w}|\mathbb{D})^{1-\alpha}}\right]d\textbf{w} \nonumber\\
        &=(1-\alpha)^2 \int q(\textbf{w}|\boldsymbol{\theta})\log\left[ \frac{q(\textbf{w}|\boldsymbol{\theta})}{P(\textbf{w}|\mathbb{D})}\right]d\textbf{w} 
\end{align}
Similarly rewriting the second term in Eq.~\ref{eq:elbo_js_app} as,
\begin{align} \label{eq:2term}
    T_2 &= \alpha \text{KL}\left(P\:||\: \text{G}'_\alpha(q,P)\right) \nonumber\\
        &= \alpha \text{KL}[P(\textbf{w}|\mathbb{D})||q(\textbf{w}|\boldsymbol{\theta})^\alpha P(\textbf{w}|\mathbb{D})^{(1-\alpha)}] \nonumber\\
        &= \alpha \int P(\textbf{w}|\mathbb{D}) \log \left[ \frac{P(\textbf{w}|\mathbb{D})}{q(\textbf{w}|\boldsymbol{\theta})^\alpha P(\textbf{w}|\mathbb{D})^{(1-\alpha)}} \right] d\textbf{w} \nonumber\\
        & = \alpha^2 \int {P(\textbf{w}|\mathbb{D})} \log \left[  \frac{P(\textbf{w}|\mathbb{D})}{q(\textbf{w}|\boldsymbol{\theta})}\right] d\textbf{w}
\end{align}
The term $T_2$ is intractable due to the posterior distribution $P(\textbf{w}|\mathbb{D})$.

\section{Insigths into Monte Carlo estimates} \label{app:mc}
\begin{figure}[h]
    \centering
    \includegraphics[width = 0.99\linewidth]{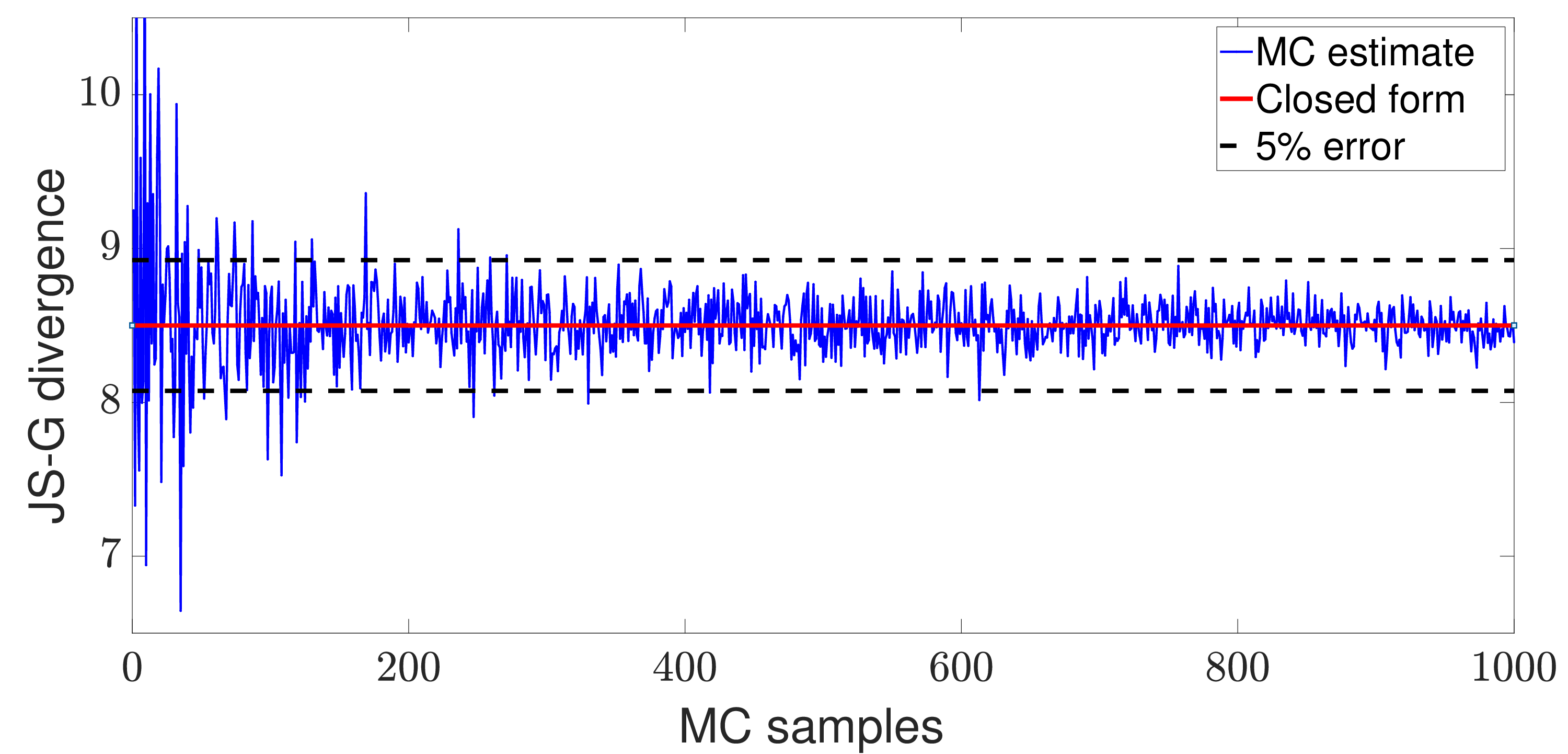}
    \caption{Comparison of MC estimates and the closed form solution of JS-G divergence demonstrating the benefit of closed form solution.}
    \label{fig:mc_est}
\end{figure}

A closed-form solution does not exist for KL and JS divergences for most distributions. In cases where such a closed form for the divergence is not available for a given distribution, we resort to Monte Carlo (MC) methods. However, the estimation of the loss function using MC methods is computationally more expensive than the closed-form evaluation as shown in \ref{fig:mc_est}. In addition, for networks with a large number of parameters, the memory requirement increases significantly with the number of MC samples. Therefore, utilizing the closed-form solution when available can save huge computational efforts and memory.

To estimate the number of MC samples required to achieve a similar level of accuracy of the closed-form expression, JS-G divergence of two Gaussian distributions $\mathcal{N}(5,1)$ and $\mathcal{N}(0,1)$ are evaluated and compared with its closed form counterpart. Fig.~\ref{fig:mc_est} shows the results of the comparison. It is seen that at least 600 samples are required to estimate the JS-G divergence within 5\% error. This implies evaluating the loss function 600 times for a given input and back-propagating the error which requires huge computational efforts. 

\section{Detailed proof of Theorem 2} \label{sec:theorem2_proof}
\textbf{Theorem 2. } 
\textit{For any two arbitrary distributions $P$ and $q$ such that $P \neq q$, $ \widetilde{\mathcal{F}}_{JSG} > {\mathcal{F}}_{KL} $ if and only if $\alpha > \frac{2 \text{ KL} (q||P)}{\text{KL}(q||P) + \text{KL} (P||q)} \in (0,\infty)$}\\

\textbf{Proof:}\\ 
Assuming,
\begin{subequations}
\begin{align*}
     \widetilde{\mathcal{F}}_{JSG} - {\mathcal{F}}_{KL} &> 0\\
\intertext{From Eq.(5) and Eq.(18) from the main text we have, }
     (1-\alpha)^2 \text{KL}(q||P) + \alpha^2 \text{KL} (P||q) - \text{KL}(q||P) &> 0\\
     (\alpha^2 - 2\alpha) \text{KL}(q||P) + \alpha^2 \text{KL} (P||q) &> 0\\
     \intertext{This leads to,}
     \alpha > \frac{2 \text{ KL} (q||P)}{\text{KL}(q||P) + \text{KL} (P||q)} 
\end{align*}
\end{subequations}
This proves that if $ \widetilde{\mathcal{F}}_{JSG} > {\mathcal{F}}_{KL} $  then $\alpha > \frac{2 \text{ KL} (q||P)}{\text{KL}(q||P) + \text{KL} (P||q)} $ .  \\

Conversely, assuming $\alpha > r$, where $r =  \frac{2 \text{ KL} (q||P)}{\text{KL}(q||P) + \text{KL} (P||q)}$, we get

\begin{align*}
    \alpha \left[\text{KL}(q||P) + \text{KL} (P||q)\right] - &2 \text{ KL} (q||P) >\\& r \left[\text{KL}(q||P) + \text{KL} (P||q)\right] - 2 \text{ KL} (q||P) \quad  (\text{since } \alpha>r)
\end{align*}
\begin{align*}
\text{Substituting r on the right hand side,}\\
    \alpha \left[\text{KL}(q||P) + \text{KL} (P||q)\right] - 2 \text{ KL} (q||P) &> 0 \\
    \alpha^2 \left[\text{KL}(q||P) + \text{KL} (P||q)\right] - 2\alpha  \text{ KL} (q||P) &> 0 \quad \quad \quad (\text{since } \alpha>0) \\
    (\alpha^2 - 2\alpha) \text{KL}(q||P) + \alpha^2 \text{KL} (P||q) &> 0\\
    (1-\alpha)^2 \text{KL}(q||P) + \alpha^2 \text{KL} (P||q) - \text{KL}(q||P) &> 0\\
    \widetilde{\mathcal{F}}_{JSG} - {\mathcal{F}}_{KL} &> 0    
\end{align*}
This proves that if  $\alpha > \frac{2 \text{ KL} (q||P)}{\text{KL}(q||P) + \text{KL} (P||q)} $ then $ \widetilde{\mathcal{F}}_{JSG} > {\mathcal{F}}_{KL} $. Hence the theorem is proved.\\

Note that, since KL divergence is always non-negative for any distributions and $P \neq q$, we have $\frac{2 \text{ KL} (q||P)}{\text{KL}(q||P) + \text{KL} (P||q)} > 0$ 

\section{Detailed proof of Theorem 3} \label{sec:theorem3_proof}
\noindent\textbf{Theorem 3.} 

\textit{If $P = \mathcal{N}(\mu_p,\sigma_p^2)\,$ and $\,q = \mathcal{N}(\mu_q,\sigma_q^2)\,$ are Gaussian distributions and $\,P \neq q\,$, then $\frac{2 \text{ KL} (q||P)}{\text{KL}(q||P) + \text{KL} (P||q)}<1$ if and only if $\sigma_p^2 > \sigma_q^2$ }.\\

\textbf{Proof: }\\

Assuming,
\begin{align*} 
     \frac{2 \text{ KL} (q||P)}{\text{KL}(q||P) + \text{KL} (P||q)} <1
\end{align*}
We have,
\begin{align}
    \text{KL} (P||q)&>\text{KL}(q||P) \label{eq:ineqkl_app}
\end{align}
Since $P$ and $q$ be Gaussian distributions with $P = \mathcal{N}(\mu_p,\sigma_p^2)$ and $q = \mathcal{N}(\mu_q,\sigma_q^2)$,  Eq.~\ref{eq:ineqkl_app} can be written as, 
\begin{align*}
    & \ln \frac{\sigma_q^2}{\sigma_p^2} + \frac{\sigma_p^2 + (\mu_q - \mu_p)^2}{\sigma_q^2} -1 >  \ln \frac{\sigma_p^2}{\sigma_q^2} + \frac{\sigma_q^2+(\mu_p - \mu_q)^2}{\sigma_p^2} -1 \\
     &\frac{\sigma_p^2}{\sigma_q^2} + \ln \frac{\sigma_q^2}{\sigma_p^2} + \frac{(\mu_q - \mu_p)^2}{\sigma_q^2}  -  \frac{\sigma_q^2}{\sigma_p^2} - \ln \frac{\sigma_p^2}{\sigma_q^2} - \frac{(\mu_p - \mu_q)^2}{\sigma_p^2}  > 0
\end{align*}
Denoting $\gamma = \frac{\sigma_p^2}{\sigma_q^2} $, we get, 
\begin{align*}
    \gamma - \frac{1}{\gamma} + \ln{\frac{1}{\gamma}} - \ln{\gamma} +  \frac{(\mu_q - \mu_p)^2}{\sigma_q^2} - \frac{(\mu_p - \mu_q)^2}{\gamma \sigma_q^2} &> 0\\
    \gamma - \frac{1}{\gamma} + \ln{\frac{1}{\gamma^2}} +  \frac{(\mu_q - \mu_p)^2}{\sigma_q^2}\left(1 - \frac{1}{\gamma}\right) &> 0\\
    \ln\left[\exp\left({\gamma - \frac{1}{\gamma}}\right)\right] + \ln{\frac{1}{\gamma^2}} +  \frac{(\mu_q - \mu_p)^2}{\sigma_q^2}\left(1 - \frac{1}{\gamma}\right) &> 0
\end{align*}
or,
\begin{align}
   \ln\left[\frac{1}{\gamma^2}\exp\left({\gamma - \frac{1}{\gamma}}\right)\right] +  \frac{(\mu_q - \mu_p)^2}{\sigma_q^2}\left(1 - \frac{1}{\gamma}\right) &> 0 \label{eq:fianl_cond_app}
\end{align}
The second term of  Eq.~\ref{eq:fianl_cond_app} is greater than 0 only when $\gamma > 1$. Consider the first term,
\begin{align*}
    \ln\left[\frac{1}{\gamma^2}\exp\left({\gamma - \frac{1}{\gamma}}\right)\right] > 0 \\
    \intertext{or,}
    \frac{1}{\gamma^2}\exp\left({\gamma - \frac{1}{\gamma}}\right) > 1
\end{align*}
$\frac{1}{\gamma^2}\exp\left({\gamma - \frac{1}{\gamma}}\right) = 1$ for $\gamma = 1$ and it is a monotonically increasing function for $\gamma>1$. This can be seen from its positive slope 
\begin{align*}
    \frac{d }{d \gamma} \left[\frac{1}{\gamma^2}\exp\left({\gamma - \frac{1}{\gamma}}\right)\right] &= \frac{1}{\gamma^2} \exp \left({\gamma - \frac{1}{\gamma}}\right) \times \left(1+\frac{1}{\gamma^2}\right) - \frac{2}{\gamma^3} \exp\left({\gamma - \frac{1}{\gamma}}\right)\\
    &= \frac{1}{\gamma^4}\exp\left({\gamma - \frac{1}{\gamma}}\right) \left[ \gamma^2 + 1 - 2\gamma \right]\\
    &= \frac{1}{\gamma^4}\exp\left({\gamma - \frac{1}{\gamma}}\right)  (\gamma-1)^2\\
    & >0 \quad \quad \quad \quad \quad \text{for } \gamma \neq 1
\end{align*}
Therefore, the first term of  Eq.~\ref{eq:fianl_cond_app} is greater than 0 only when $\gamma > 1$.
Thus, the condition in Eq.~\ref{eq:fianl_cond_app} is satisfied only when $\gamma > 1$, which implies
\begin{equation}
    \sigma_p^2 > \sigma_q^2
\end{equation} 
Thus if $\frac{2 \text{ KL} (q||P)}{\text{KL}(q||P) + \text{KL} (P||q)} <1$ then $\sigma_p^2 > \sigma_q^2\,$. 

Conversely, assuming $\sigma_p^2 > \sigma_q^2$, i.e. $\gamma>1$  consider, 
\begin{align*}
&\ln\left[\frac{1}{\gamma^2}\exp\left({\gamma - \frac{1}{\gamma}}\right)\right] +  \frac{(\mu_q - \mu_p)^2}{\sigma_q^2}\left(1 - \frac{1}{\gamma}\right) > 0
\intertext{Which leads to,}
    &\text{KL} (P||q)>\text{KL}(q||P) \quad \quad \quad (\text{following the steps as above})
\intertext{or,}
    &\text{KL} (P||q) + \text{KL}(q||P) > \text{KL}(q||P) + \text{KL}(q||P)\\
     & \frac{2 \text{KL}(q||P)}{\text{KL} (P||q) + \text{KL}(q||P)} <1
\end{align*}
Thus if $\sigma_p^2 > \sigma_q^2\,$ then $\frac{2 \text{ KL} (q||P)}{\text{KL}(q||P) + \text{KL} (P||q)} <1$. Hence the theorem is proved.\\

\begin{figure}[h]
    \centering
    \includegraphics[width = 0.6\linewidth]{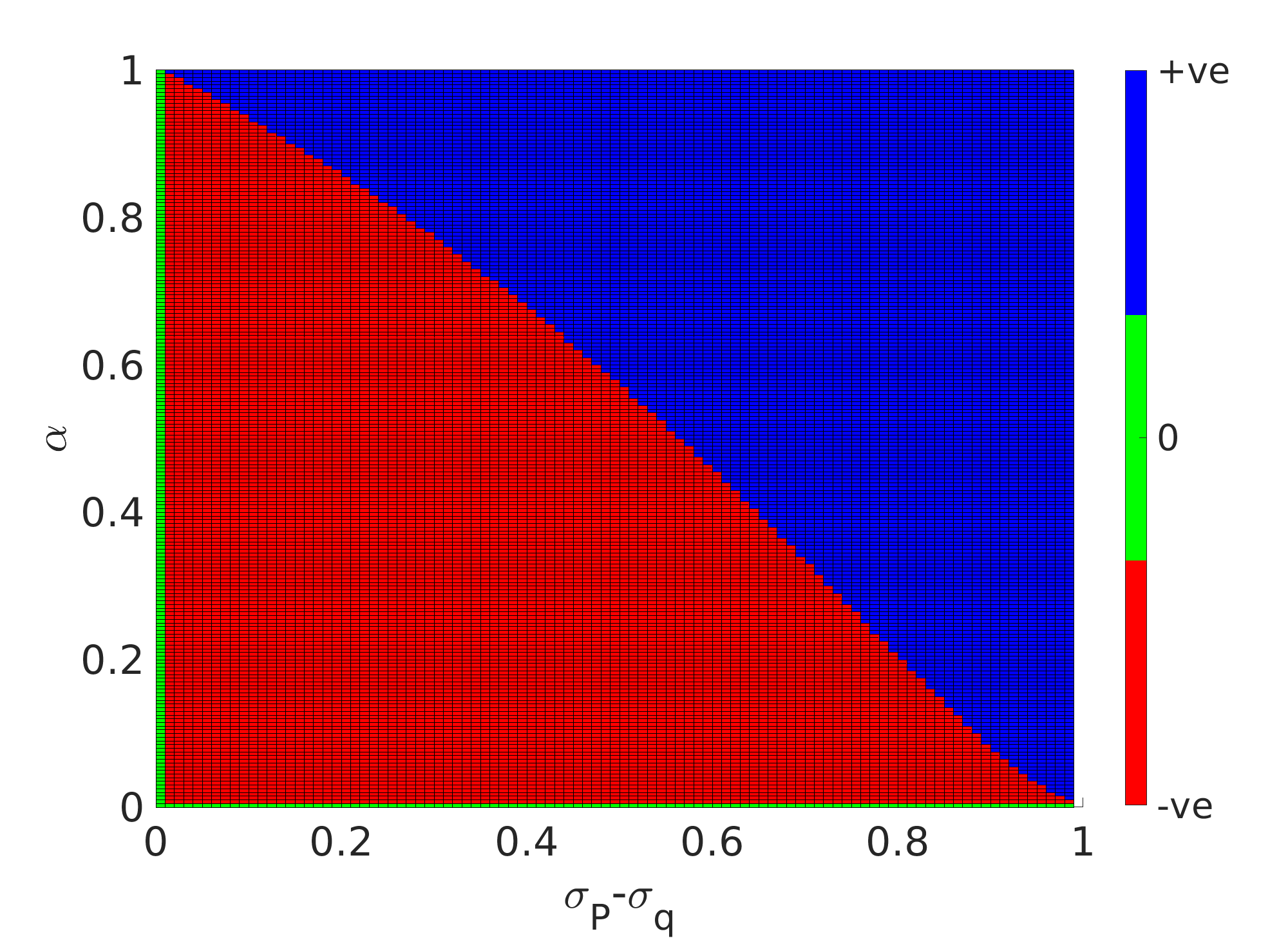}
    \caption{The sign of $\widetilde{\mathcal{F}}_{JSG} - \mathcal{F}_{KL}$ is plotted for various values of $\alpha$ and $\sigma_p - \sigma_q$. }
    \label{fig:theorem3}
\end{figure}

\textbf{Corollary: }From Theorem 2 and 3: $ \widetilde{\mathcal{F}}_{JSG} > {\mathcal{F}}_{KL} $ if $\sigma_p^2 > \sigma_q^2$ and $\forall \, \alpha \in (0,1]$ such that $ \alpha > \frac{2 \text{ KL} (q||P)}{\text{KL}(q||P) + \text{KL} (P||q)}$. Where, $P$ and $q$ are Gaussians and $P \neq q$. \\
Fig.~\ref{fig:theorem3} shows the sign of $ \widetilde{\mathcal{F}}_{JSG} - \mathcal{F}_{KL}$ for various values of $\alpha$ and $\sigma_p - \sigma_q$. Two Gaussian distributions $P = \mathcal{N}(0,1)$ and $q = \mathcal{N}(0,\sigma_q^2)$ are considered for this purpose. It is evident that at least one value of $\alpha$ exists such that $ \widetilde{\mathcal{F}}_{JSG} - \mathcal{F}_{KL} > 0$ when $\sigma_p > \sigma_q$. In addition, for $\alpha = 0$ or $\sigma_p = \sigma_q$, $ \widetilde{\mathcal{F}}_{JSG} = \mathcal{F}_{KL}$.

\section{Data sets} \label{app:dataset}
\subsection{Regression datasets}
\noindent \textbf{Airfoil:} This NASA dataset includes NACA 0012 airfoils of various sizes tested at different wind tunnel speeds and angles of attack. The goal is to predict the scaled sound pressure.\\
\textbf{Aquatic:} This dataset is used to predict quantitative acute aquatic toxicity towards Daphnia Magna from 8 molecular descriptors of 546 chemicals. \\
\textbf{Building: }This dataset includes construction cost, sale prices, project variables, and economic variables corresponding to real estate single-family residential apartments in Tehran, Iran.\\
\textbf{Concrete:} The dataset contains attributes to predict concrete compressive strength.\\
\textbf{Real Estate: } This dataset contains real estate valuation collected from Sindian Dist., New Taipei City, Taiwan.\\
\textbf{Wine: } This dataset is used to predict model wine quality based on physicochemical tests.
\subsection{CIFAR-10}
The CIFAR-10 data set \cite{krizhevsky2009learning} consists of 60,000 images of size $32 \times 32 \times 3$ belonging to 10 mutually exclusive classes. This data set is unbiased, with each of the 10 classes having 6,000 images. \\
Images were resized to $3\times 224 \times 224 $ pixels and normalized using the min-max normalization technique. The training data set was split into 80\% 20\% for training and validation respectively. 
\subsection{Histopathology}
The histopathology data set \cite{janowczyk2016deep,cruz2014automatic} is publicly available under a CC0 license at \cite{mooney_b_2017}. The data set consists of images containing regions of Invasive Ductal Carcinoma. The original data set consisted of 162 whole-mount slide images of Breast Cancer specimens scanned at 40x.
From the original whole slide images, 277,524 patches of size $50\times50\times 3$ pixels were extracted (198,738  negatives and 78,786 positives), labeled by pathologists, and provided as a data set for classification.

The data set consists of a positive (1) and a negative (0) class. 20\% of the entire data set was used as the testing set for our study. The remaining 80\% of the entire data was further split into a training set and a validation set (80\%-20\% split) to perform hyperparameter optimization.  
The images were shuffled and converted from uint8 to float format for normalizing. As a post-processing step, we computed the complement of all the images (training and testing) and then used them as inputs to the neural network. The images were resized to $3 \times 224 \times 224$ pixel-wise normalization and complement were carried out as $p_n = (255-p)/255 $. $p$ is the original pixel value and $p_n$ is the pixel value after normalization and complement.

\section{Results of hyperparameter optimization} \label{app:hyperres}
\begin{table}[htpb]
    \centering
    \caption{CIFAR-10 data set}
    \begin{tabular}{cccccccccc}  \toprule
        \multirow{2}{*}{\textbf{Div}}&\multirow{2}{*}{\textbf{Parameter}}&\multicolumn{5}{c}{\textbf{Noise level} ($\boldsymbol \sigma$)} \\ 
        \cmidrule{3-7}
          & & \textbf{0.1} & \textbf{0.2} &\textbf{0.3}&\textbf{0.4}&\textbf{0.5} \\ \midrule
         KL& LR&$1\mathrm{e}{-4}$& $1\mathrm{e}{-4}$ & $1\mathrm{e}{-4}$& $1\mathrm{e}{-3}$& $1\mathrm{e}{-3}$\\ \midrule
          \multirow{2}*{JS-G}&$\boldsymbol \alpha$ &0.004 &0.1313 &0.2855 &0.3052 &0.2637\\ 
         &LR &$1\mathrm{e}{-4}$& $1\mathrm{e}{-4}$ & $1\mathrm{e}{-4}$& $1\mathrm{e}{-4}$& $1\mathrm{e}{-5}$\\ \midrule
         \multirow{3}{*}{JS-A}&$\boldsymbol \lambda$ & 1000 & 1000 & 1000& 1000& 1000 \\
         &$\boldsymbol \alpha$&0.7584& 0.6324& 0.1381&0.6286 & 0.1588 \\
         & LR &$1\mathrm{e}{-4}$ &$1\mathrm{e}{-4}$& $1\mathrm{e}{-4}$ & $1\mathrm{e}{-3}$& $1\mathrm{e}{-4}$ \\ \bottomrule
         \toprule
         \multicolumn{6}{c}{} \\ 
          \multirow{2}{*}{\textbf{Div}}&\multirow{2}{*}{\textbf{Parameter}}&\multicolumn{4}{c}{\textbf{Noise level} ($\boldsymbol \sigma$)} \\  \cmidrule{3-6}
         & &\textbf{0.6}&\textbf{0.7}&\textbf{0.8}&\textbf{0.9} \\ \midrule
         KL & LR &$1\mathrm{e}{-3}$& $1\mathrm{e}{-3}$ & $1\mathrm{e}{-3}$& $1\mathrm{e}{-3}$ \\ \midrule
         \multirow{2}{*}{JS-G}&$\boldsymbol \alpha$ & 0.2249& 0.3704&  0.3893& 0.7584 \\ 
         &LR&$1\mathrm{e}{-4}$& $1\mathrm{e}{-4}$ & $1\mathrm{e}{-5}$& $1\mathrm{e}{-3}$ \\ \midrule
         \multirow{3}{*}{JS-A}&$\boldsymbol \lambda$ & 100 & 1000& $1\mathrm{e}{4}$& $1\mathrm{e}{5}$ \\
         &$\boldsymbol \alpha$ & 0.4630& 0.1220& 0.2282& 0.5792  \\ 
         & LR &$1\mathrm{e}{-3}$ &$1\mathrm{e}{-4}$& $1\mathrm{e}{-5}$ & $1\mathrm{e}{-5}$  \\
         \bottomrule
    \end{tabular}
    \label{tab:Cifar}
\end{table}
A Tree-structured Parzen Estimator (TPE) algorithm \cite{bergstra2011algorithms} is used for hyperparameter optimization which is a sequential model-based optimization approach. 
 In this approach, models are constructed to approximate the performance of hyperparameters based on historical measurements. New hyperparameters are chosen based on this model to evaluate performance. 
A python library Hyperopt \cite{bergstra2013making} is used to implement this optimization algorithm over a given search space.

Results of the hyperparameter optimization for the two classification data sets are presented in Tables.~\ref{tab:Cifar} and \ref{tab:histo}. In Table.~\ref{tab:Cifar} Div denotes the divergence measure and LR is the learning rate. $\lambda$ is taken as 1 for the JS-G divergence-based loss function throughout this work unless specified otherwise. 


\begin{table}[h]
    \centering
    \caption{Histopathology data set}
    \begin{tabular}{cccc} \toprule
         \textbf{Divergence} &$\boldsymbol{\alpha}$ & $\boldsymbol{\lambda}$ & \textbf{Learning rate}\\ \midrule
         KL& - & - & $1\mathrm{e}{-4}$ \\ \midrule
         JS-G& 0.0838 & 1 & $1\mathrm{e}{-4}$ \\ \midrule
         JS-A& 0.0729 & 100 & $1\mathrm{e}{-4}$ \\ \bottomrule
    \end{tabular}\\
    \label{tab:histo}
\end{table}

\section*{Author Biographies}

\begin{wrapfigure}{l}{0.3\textwidth}
  \begin{center}
   \includegraphics[width=0.9\linewidth,clip,keepaspectratio]{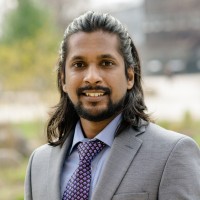}
  \end{center}
\end{wrapfigure}
\noindent \textbf{Ponkrshnan Thiagarajan} is a postdoctoral fellow at Hopkins Extreme Materials Institute, Johns Hopkins University. He received his Ph.D. in the Department of Mechanical Engineering-Engineering Mechanics at Michigan Technological University, Houghton, MI, USA. His research interests include Bayesian neural networks, Bayesian calibration, and Uncertainty quantification of computational models. He received his master's degree in aerospace engineering from the Indian Institute of Technology - Madras, and a bachelor's degree in aeronautical engineering from Anna University, Chennai. \\

\begin{wrapfigure}{l}{0.3\textwidth}
  \begin{center}
   \includegraphics[width=0.9\linewidth,clip,keepaspectratio]{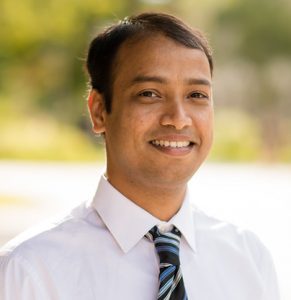}
  \end{center}
\end{wrapfigure}
\noindent \textbf{Susanta Ghosh} is an Assistant Professor, Mechanical Engineering-Engineering Mechanics, a Faculty member of Center for Data Sciences at the Institute of Computing and Cybersystems and a Faculty member of the Center for Applied Mathematics and Statistics at Michigan Technological University, Houghton, MI, USA. He has worked as a post-doctoral research scholar at Duke University, the University of Michigan, and The Technical University of Catalunya. His M.S. and Ph.D. degrees are from the Indian Institute of Science, Bangalore.
\newpage

 \bibliographystyle{elsarticle-num-names} 
 \bibliography{references}

\begin{thebibliography}{40}
\expandafter\ifx\csname natexlab\endcsname\relax\def\natexlab#1{#1}\fi
\providecommand{\url}[1]{\texttt{#1}}
\providecommand{\href}[2]{#2}
\providecommand{\path}[1]{#1}
\providecommand{\DOIprefix}{doi:}
\providecommand{\ArXivprefix}{arXiv:}
\providecommand{\URLprefix}{URL: }
\providecommand{\Pubmedprefix}{pmid:}
\providecommand{\doi}[1]{\href{http://dx.doi.org/#1}{\path{#1}}}
\providecommand{\Pubmed}[1]{\href{pmid:#1}{\path{#1}}}
\providecommand{\bibinfo}[2]{#2}
\ifx\xfnm\relax \def\xfnm[#1]{\unskip,\space#1}\fi
\bibitem[{Samarasinghe(2016)}]{samarasinghe2016neural}
\bibinfo{author}{S.~Samarasinghe}, \bibinfo{title}{Neural networks for applied sciences and engineering: from fundamentals to complex pattern recognition}, \bibinfo{publisher}{Auerbach publications}, \bibinfo{year}{2016}.
\bibitem[{Li et~al.(2021)Li, Liu, Yang, Peng, and Zhou}]{li2021survey}
\bibinfo{author}{Z.~Li}, \bibinfo{author}{F.~Liu}, \bibinfo{author}{W.~Yang}, \bibinfo{author}{S.~Peng}, \bibinfo{author}{J.~Zhou},
\newblock \bibinfo{title}{A survey of convolutional neural networks: analysis, applications, and prospects},
\newblock \bibinfo{journal}{IEEE transactions on neural networks and learning systems}  (\bibinfo{year}{2021}).
\bibitem[{Buda et~al.(2018)Buda, Maki, and Mazurowski}]{buda2018systematic}
\bibinfo{author}{M.~Buda}, \bibinfo{author}{A.~Maki}, \bibinfo{author}{M.~A. Mazurowski},
\newblock \bibinfo{title}{A systematic study of the class imbalance problem in convolutional neural networks},
\newblock \bibinfo{journal}{Neural Networks} \bibinfo{volume}{106} (\bibinfo{year}{2018}) \bibinfo{pages}{249--259}.
\bibitem[{Thiagarajan et~al.(2021)Thiagarajan, Khairnar, and Ghosh}]{thiagarajan2021explanation}
\bibinfo{author}{P.~Thiagarajan}, \bibinfo{author}{P.~Khairnar}, \bibinfo{author}{S.~Ghosh},
\newblock \bibinfo{title}{Explanation and use of uncertainty quantified by bayesian neural network classifiers for breast histopathology images},
\newblock \bibinfo{journal}{IEEE Transactions on Medical Imaging} \bibinfo{volume}{41} (\bibinfo{year}{2021}) \bibinfo{pages}{815--825}.
\bibitem[{Kabir et~al.(2018)Kabir, Khosravi, Hosen, and Nahavandi}]{kabir2018neural}
\bibinfo{author}{H.~D. Kabir}, \bibinfo{author}{A.~Khosravi}, \bibinfo{author}{M.~A. Hosen}, \bibinfo{author}{S.~Nahavandi},
\newblock \bibinfo{title}{Neural network-based uncertainty quantification: A survey of methodologies and applications},
\newblock \bibinfo{journal}{IEEE access} \bibinfo{volume}{6} (\bibinfo{year}{2018}) \bibinfo{pages}{36218--36234}.
\bibitem[{Jospin et~al.(2022)Jospin, Laga, Boussaid, Buntine, and Bennamoun}]{jospin2022hands}
\bibinfo{author}{L.~V. Jospin}, \bibinfo{author}{H.~Laga}, \bibinfo{author}{F.~Boussaid}, \bibinfo{author}{W.~Buntine}, \bibinfo{author}{M.~Bennamoun},
\newblock \bibinfo{title}{Hands-on bayesian neural networks—a tutorial for deep learning users},
\newblock \bibinfo{journal}{IEEE Computational Intelligence Magazine} \bibinfo{volume}{17} (\bibinfo{year}{2022}) \bibinfo{pages}{29--48}.
\bibitem[{Bortolussi et~al.(2024)Bortolussi, Carbone, Laurenti, Patane, Sanguinetti, and Wicker}]{10506195}
\bibinfo{author}{L.~Bortolussi}, \bibinfo{author}{G.~Carbone}, \bibinfo{author}{L.~Laurenti}, \bibinfo{author}{A.~Patane}, \bibinfo{author}{G.~Sanguinetti}, \bibinfo{author}{M.~Wicker},
\newblock \bibinfo{title}{On the robustness of bayesian neural networks to adversarial attacks},
\newblock \bibinfo{journal}{IEEE Transactions on Neural Networks and Learning Systems}  (\bibinfo{year}{2024}) \bibinfo{pages}{1--14}. \DOIprefix\doi{10.1109/TNNLS.2024.3386642}.
\bibitem[{Lemos et~al.(2023)Lemos, Cranmer, Abidi, Hahn, Eickenberg, Massara, Yallup, and Ho}]{lemos2023robust}
\bibinfo{author}{P.~Lemos}, \bibinfo{author}{M.~Cranmer}, \bibinfo{author}{M.~Abidi}, \bibinfo{author}{C.~Hahn}, \bibinfo{author}{M.~Eickenberg}, \bibinfo{author}{E.~Massara}, \bibinfo{author}{D.~Yallup}, \bibinfo{author}{S.~Ho},
\newblock \bibinfo{title}{Robust simulation-based inference in cosmology with bayesian neural networks},
\newblock \bibinfo{journal}{Machine Learning: Science and Technology} \bibinfo{volume}{4} (\bibinfo{year}{2023}) \bibinfo{pages}{01LT01}.
\bibitem[{Dong et~al.(2023)Dong, An, Lu, and Geng}]{dong2023nuclear}
\bibinfo{author}{X.-X. Dong}, \bibinfo{author}{R.~An}, \bibinfo{author}{J.-X. Lu}, \bibinfo{author}{L.-S. Geng},
\newblock \bibinfo{title}{Nuclear charge radii in bayesian neural networks revisited},
\newblock \bibinfo{journal}{Physics Letters B} \bibinfo{volume}{838} (\bibinfo{year}{2023}) \bibinfo{pages}{137726}.
\bibitem[{Pathrudkar et~al.(2023)Pathrudkar, Thiagarajan, Agarwal, Banerjee, and Ghosh}]{pathrudkar2023electronic}
\bibinfo{author}{S.~Pathrudkar}, \bibinfo{author}{P.~Thiagarajan}, \bibinfo{author}{S.~Agarwal}, \bibinfo{author}{A.~S. Banerjee}, \bibinfo{author}{S.~Ghosh},
\newblock \bibinfo{title}{Electronic structure prediction of multi-million atom systems through uncertainty quantification enabled transfer learning},
\newblock \bibinfo{journal}{arXiv preprint arXiv:2308.13096}  (\bibinfo{year}{2023}).
\bibitem[{Luo et~al.(2024)Luo, Zhu, Li, Zhu, Li, Hu, Fan, Chang, Zhuang, and Yang}]{luo2024ultrasonic}
\bibinfo{author}{K.~Luo}, \bibinfo{author}{J.~Zhu}, \bibinfo{author}{Z.~Li}, \bibinfo{author}{H.~Zhu}, \bibinfo{author}{Y.~Li}, \bibinfo{author}{R.~Hu}, \bibinfo{author}{T.~Fan}, \bibinfo{author}{X.~Chang}, \bibinfo{author}{L.~Zhuang}, \bibinfo{author}{Z.~Yang},
\newblock \bibinfo{title}{Ultrasonic lamb wave damage detection of cfrp composites using the bayesian neural network},
\newblock \bibinfo{journal}{Journal of Nondestructive Evaluation} \bibinfo{volume}{43} (\bibinfo{year}{2024}) \bibinfo{pages}{48}.
\bibitem[{Tishby et~al.(1989)Tishby, Levin, and Solla}]{tishby1989consistent}
\bibinfo{author}{N.~Tishby}, \bibinfo{author}{E.~Levin}, \bibinfo{author}{S.~A. Solla},
\newblock \bibinfo{title}{Consistent inference of probabilities in layered networks: Predictions and generalization},
\newblock in: \bibinfo{booktitle}{International Joint Conference on Neural Networks}, volume~\bibinfo{volume}{2}, \bibinfo{organization}{IEEE New York}, \bibinfo{year}{1989}, pp. \bibinfo{pages}{403--409}.
\bibitem[{Denker and LeCun(1990)}]{denker1990transforming}
\bibinfo{author}{J.~Denker}, \bibinfo{author}{Y.~LeCun},
\newblock \bibinfo{title}{Transforming neural-net output levels to probability distributions},
\newblock \bibinfo{journal}{Advances in neural information processing systems} \bibinfo{volume}{3} (\bibinfo{year}{1990}).
\bibitem[{Goan and Fookes(2020)}]{Goan2020}
\bibinfo{author}{E.~Goan}, \bibinfo{author}{C.~Fookes}, \bibinfo{title}{Bayesian Neural Networks: An Introduction and Survey}, \bibinfo{publisher}{Springer International Publishing}, \bibinfo{address}{Cham}, \bibinfo{year}{2020}, pp. \bibinfo{pages}{45--87}.
\bibitem[{Gal(2016)}]{gal2016uncertainty}
\bibinfo{author}{Y.~Gal},
\newblock \bibinfo{title}{Uncertainty in deep learning},
\newblock \bibinfo{journal}{PhD thesis, University of Cambridge}  (\bibinfo{year}{2016}).
\bibitem[{Magris and Iosifidis(2023)}]{magris2023bayesian}
\bibinfo{author}{M.~Magris}, \bibinfo{author}{A.~Iosifidis},
\newblock \bibinfo{title}{Bayesian learning for neural networks: an algorithmic survey},
\newblock \bibinfo{journal}{Artificial Intelligence Review} \bibinfo{volume}{56} (\bibinfo{year}{2023}) \bibinfo{pages}{11773--11823}.
\bibitem[{Hinton and Van~Camp(1993)}]{hinton1993keeping}
\bibinfo{author}{G.~E. Hinton}, \bibinfo{author}{D.~Van~Camp},
\newblock \bibinfo{title}{Keeping the neural networks simple by minimizing the description length of the weights},
\newblock in: \bibinfo{booktitle}{Proceedings of the sixth annual conference on Computational learning theory}, \bibinfo{year}{1993}, pp. \bibinfo{pages}{5--13}.
\bibitem[{Barber and Bishop(1998)}]{barber1998ensemble}
\bibinfo{author}{D.~Barber}, \bibinfo{author}{C.~M. Bishop},
\newblock \bibinfo{title}{Ensemble learning in bayesian neural networks},
\newblock \bibinfo{journal}{Nato ASI Series F Computer and Systems Sciences} \bibinfo{volume}{168} (\bibinfo{year}{1998}) \bibinfo{pages}{215--238}.
\bibitem[{Graves(2011)}]{graves2011practical}
\bibinfo{author}{A.~Graves},
\newblock \bibinfo{title}{Practical variational inference for neural networks},
\newblock \bibinfo{journal}{Advances in neural information processing systems} \bibinfo{volume}{24} (\bibinfo{year}{2011}).
\bibitem[{Hern{\'a}ndez-Lobato and Adams(2015)}]{hernandez2015probabilistic}
\bibinfo{author}{J.~M. Hern{\'a}ndez-Lobato}, \bibinfo{author}{R.~Adams},
\newblock \bibinfo{title}{Probabilistic backpropagation for scalable learning of bayesian neural networks},
\newblock in: \bibinfo{booktitle}{International conference on machine learning}, \bibinfo{organization}{Proceedings of Machine Learning Research}, \bibinfo{year}{2015}, pp. \bibinfo{pages}{1861--1869}.
\bibitem[{Neal(2012)}]{neal2012bayesian}
\bibinfo{author}{R.~M. Neal}, \bibinfo{title}{Bayesian learning for neural networks}, volume \bibinfo{volume}{118}, \bibinfo{publisher}{Springer Science \& Business Media}, \bibinfo{year}{2012}.
\bibitem[{Welling and Teh(2011)}]{welling2011bayesian}
\bibinfo{author}{M.~Welling}, \bibinfo{author}{Y.~W. Teh},
\newblock \bibinfo{title}{Bayesian learning via stochastic gradient langevin dynamics},
\newblock in: \bibinfo{booktitle}{Proceedings of the 28th international conference on machine learning (ICML-11)}, \bibinfo{organization}{Citeseer}, \bibinfo{year}{2011}, pp. \bibinfo{pages}{681--688}.
\bibitem[{Robert et~al.(2018)Robert, Elvira, Tawn, and Wu}]{robert2018accelerating}
\bibinfo{author}{C.~P. Robert}, \bibinfo{author}{V.~Elvira}, \bibinfo{author}{N.~Tawn}, \bibinfo{author}{C.~Wu},
\newblock \bibinfo{title}{Accelerating mcmc algorithms},
\newblock \bibinfo{journal}{Wiley Interdisciplinary Reviews: Computational Statistics} \bibinfo{volume}{10} (\bibinfo{year}{2018}) \bibinfo{pages}{e1435}.
\bibitem[{Blundell et~al.(2015)Blundell, Cornebise, Kavukcuoglu, and Wierstra}]{blundell2015weight}
\bibinfo{author}{C.~Blundell}, \bibinfo{author}{J.~Cornebise}, \bibinfo{author}{K.~Kavukcuoglu}, \bibinfo{author}{D.~Wierstra},
\newblock \bibinfo{title}{Weight uncertainty in neural network},
\newblock in: \bibinfo{booktitle}{International conference on machine learning}, \bibinfo{organization}{Proceedings of Machine Learning Research}, \bibinfo{year}{2015}, pp. \bibinfo{pages}{1613--1622}.
\bibitem[{Hensman et~al.(2014)Hensman, Zwiessele, and Lawrence}]{hensman2014tilted}
\bibinfo{author}{J.~Hensman}, \bibinfo{author}{M.~Zwiessele}, \bibinfo{author}{N.~D. Lawrence},
\newblock \bibinfo{title}{Tilted variational bayes},
\newblock in: \bibinfo{booktitle}{Artificial Intelligence and Statistics}, \bibinfo{organization}{Proceedings of Machine Learning Research}, \bibinfo{year}{2014}, pp. \bibinfo{pages}{356--364}.
\bibitem[{Dieng et~al.(2017)Dieng, Tran, Ranganath, Paisley, and Blei}]{dieng2017variational}
\bibinfo{author}{A.~B. Dieng}, \bibinfo{author}{D.~Tran}, \bibinfo{author}{R.~Ranganath}, \bibinfo{author}{J.~Paisley}, \bibinfo{author}{D.~Blei},
\newblock \bibinfo{title}{Variational inference via $\chi$ upper bound minimization},
\newblock \bibinfo{journal}{Advances in Neural Information Processing Systems} \bibinfo{volume}{30} (\bibinfo{year}{2017}).
\bibitem[{Deasy et~al.(2020)Deasy, Simidjievski, and Li{\`o}}]{deasy2020constraining}
\bibinfo{author}{J.~Deasy}, \bibinfo{author}{N.~Simidjievski}, \bibinfo{author}{P.~Li{\`o}},
\newblock \bibinfo{title}{Constraining variational inference with geometric jensen-shannon divergence},
\newblock \bibinfo{journal}{Advances in Neural Information Processing Systems} \bibinfo{volume}{33} (\bibinfo{year}{2020}) \bibinfo{pages}{10647--10658}.
\bibitem[{Li and Turner(2016)}]{li2016renyi}
\bibinfo{author}{Y.~Li}, \bibinfo{author}{R.~E. Turner},
\newblock \bibinfo{title}{R{\'e}nyi divergence variational inference},
\newblock \bibinfo{journal}{Advances in neural information processing systems} \bibinfo{volume}{29} (\bibinfo{year}{2016}).
\bibitem[{Wan et~al.(2020)Wan, Li, and Hovakimyan}]{wan2020f}
\bibinfo{author}{N.~Wan}, \bibinfo{author}{D.~Li}, \bibinfo{author}{N.~Hovakimyan},
\newblock \bibinfo{title}{F-divergence variational inference},
\newblock \bibinfo{journal}{Advances in neural information processing systems} \bibinfo{volume}{33} (\bibinfo{year}{2020}) \bibinfo{pages}{17370--17379}.
\bibitem[{Nielsen(2019)}]{nielsen2019jensen}
\bibinfo{author}{F.~Nielsen},
\newblock \bibinfo{title}{On the jensen--shannon symmetrization of distances relying on abstract means},
\newblock \bibinfo{journal}{Entropy} \bibinfo{volume}{21} (\bibinfo{year}{2019}) \bibinfo{pages}{485}.
\bibitem[{Higgins et~al.(2017)Higgins, Matthey, Pal, Burgess, Glorot, Botvinick, Mohamed, and Lerchner}]{higgins2016beta}
\bibinfo{author}{I.~Higgins}, \bibinfo{author}{L.~Matthey}, \bibinfo{author}{A.~Pal}, \bibinfo{author}{C.~Burgess}, \bibinfo{author}{X.~Glorot}, \bibinfo{author}{M.~Botvinick}, \bibinfo{author}{S.~Mohamed}, \bibinfo{author}{A.~Lerchner},
\newblock \bibinfo{title}{beta-{VAE}: Learning basic visual concepts with a constrained variational framework},
\newblock in: \bibinfo{booktitle}{International Conference on Learning Representations}, \bibinfo{year}{2017}. \URLprefix \url{https://openreview.net/forum?id=Sy2fzU9gl}.
\bibitem[{Markelle~Kelly(2024)}]{uci}
\bibinfo{author}{K.~N. Markelle~Kelly, Rachel~Longjohn}, \bibinfo{title}{The uci machine learning repository}, \bibinfo{howpublished}{\url{https://archive.ics.uci.edu }}, \bibinfo{year}{Accessed Online; July 14, 2024}.
\bibitem[{Krizhevsky et~al.(2009)Krizhevsky, Hinton et~al.}]{krizhevsky2009learning}
\bibinfo{author}{A.~Krizhevsky}, \bibinfo{author}{G.~Hinton}, et~al.,
\newblock \bibinfo{title}{Learning multiple layers of features from tiny images}  (\bibinfo{year}{2009}).
\bibitem[{Janowczyk and Madabhushi(2016)}]{janowczyk2016deep}
\bibinfo{author}{A.~Janowczyk}, \bibinfo{author}{A.~Madabhushi},
\newblock \bibinfo{title}{Deep learning for digital pathology image analysis: A comprehensive tutorial with selected use cases},
\newblock \bibinfo{journal}{Journal of pathology informatics} \bibinfo{volume}{7} (\bibinfo{year}{2016}) \bibinfo{pages}{29}.
\bibitem[{Cruz-Roa et~al.(2014)Cruz-Roa, Basavanhally, Gonz{\'a}lez, Gilmore, Feldman, Ganesan, Shih, Tomaszewski, and Madabhushi}]{cruz2014automatic}
\bibinfo{author}{A.~Cruz-Roa}, \bibinfo{author}{A.~Basavanhally}, \bibinfo{author}{F.~Gonz{\'a}lez}, \bibinfo{author}{H.~Gilmore}, \bibinfo{author}{M.~Feldman}, \bibinfo{author}{S.~Ganesan}, \bibinfo{author}{N.~Shih}, \bibinfo{author}{J.~Tomaszewski}, \bibinfo{author}{A.~Madabhushi},
\newblock \bibinfo{title}{Automatic detection of invasive ductal carcinoma in whole slide images with convolutional neural networks},
\newblock in: \bibinfo{booktitle}{Medical Imaging 2014: Digital Pathology}, volume \bibinfo{volume}{9041}, \bibinfo{organization}{SPIE}, \bibinfo{year}{2014}, p. \bibinfo{pages}{904103}.
\bibitem[{{Paul Mooney}(2017)}]{mooney_b_2017}
\bibinfo{author}{{Paul Mooney}}, \bibinfo{title}{{Breast Histopathology Images}}, \bibinfo{howpublished}{\url{https://www.kaggle.com/paultimothymooney/breast-histopathology-images}}, \bibinfo{year}{2017}.
\bibitem[{Bergstra et~al.(2011)Bergstra, Bardenet, Bengio, and K{\'e}gl}]{bergstra2011algorithms}
\bibinfo{author}{J.~Bergstra}, \bibinfo{author}{R.~Bardenet}, \bibinfo{author}{Y.~Bengio}, \bibinfo{author}{B.~K{\'e}gl},
\newblock \bibinfo{title}{Algorithms for hyper-parameter optimization},
\newblock in: \bibinfo{booktitle}{25th annual conference on neural information processing systems (NIPS 2011)}, volume~\bibinfo{volume}{24}, \bibinfo{organization}{Neural Information Processing Systems Foundation}, \bibinfo{year}{2011}.
\bibitem[{He et~al.(2016)He, Zhang, Ren, and Sun}]{he2016deep}
\bibinfo{author}{K.~He}, \bibinfo{author}{X.~Zhang}, \bibinfo{author}{S.~Ren}, \bibinfo{author}{J.~Sun},
\newblock \bibinfo{title}{Deep residual learning for image recognition},
\newblock in: \bibinfo{booktitle}{Proceedings of the IEEE conference on computer vision and pattern recognition}, \bibinfo{year}{2016}, pp. \bibinfo{pages}{770--778}.
\bibitem[{Krizhevsky et~al.(2012)Krizhevsky, Sutskever, and Hinton}]{krizhevsky2012imagenet}
\bibinfo{author}{A.~Krizhevsky}, \bibinfo{author}{I.~Sutskever}, \bibinfo{author}{G.~E. Hinton},
\newblock \bibinfo{title}{Imagenet classification with deep convolutional neural networks},
\newblock in: \bibinfo{booktitle}{Advances in neural information processing systems}, \bibinfo{year}{2012}, pp. \bibinfo{pages}{1097--1105}.
\bibitem[{Bergstra et~al.(2013)Bergstra, Yamins, and Cox}]{bergstra2013making}
\bibinfo{author}{J.~Bergstra}, \bibinfo{author}{D.~Yamins}, \bibinfo{author}{D.~Cox},
\newblock \bibinfo{title}{Making a science of model search: Hyperparameter optimization in hundreds of dimensions for vision architectures},
\newblock in: \bibinfo{booktitle}{International conference on machine learning}, \bibinfo{organization}{Proceedings of Machine Learning Research}, \bibinfo{year}{2013}, pp. \bibinfo{pages}{115--123}.

\end{thebibliography}

\end{document}